\documentclass{IEEEtran}

\usepackage{graphicx}
\graphicspath{ {./img/} }
\usepackage{amssymb}
\usepackage[font=small]{caption}
\usepackage{subcaption}
\usepackage[inline]{enumitem}
\usepackage{glossaries}
\usepackage{color}
\usepackage{xcolor}
\usepackage[export]{adjustbox}
\PassOptionsToPackage{hyphens}{url}\usepackage{hyperref}
\newacronym{era5}{ERA5}{ECMWF reanalysis version 5}
\newacronym{esa}{ESA}{European Space Agency}

\newacronym{sar}{SAR}{Synthetic Aperture Radar}
\newacronym{wsss}{WSSS}{Weakly Supervised Semantic Segmentation}
\newacronym{xai}{XAI}{Explainable AI}

\newacronym{cnn}{CNN}{Convolutional Neural Network}
\newacronym{sam}{SAM}{Segment Anything Model}
\newacronym{aer}{AER}{Adversarial Erasing}
\newacronym{crest}{CREST}{Constrained Region Erasing with Soft Targets}
\newacronym{core}{CORE}{Constrained Ordinal Region Expansion}
\newacronym{ce}{CE}{Cross-Entropy}
\newacronym{db}{DB}{Dynamic Bootstrapping}

\newacronym{bus}{BUS}{UCLM Breast Ultrasound lesion Segmentation}
\newglossaryentry{voc}{
  name={VOC},
  text={VOC},
  plural={VOCs},
  first={PASCAL VOC 2012},
  firstplural={PASCAL VOC 2012},
  description={PASCAL VOC 2012}
}

\newacronym{crf}{CRF}{Conditional Random Field}
\newacronym{slp}{SLP}{Sea Level Pressure}

\newacronym{gradcam}{Grad-CAM}{Gradient-weighted Class Activation Mapping}

\newacronym{rgb}{RGB}{red-green-blue}
\newacronym{tn}{TN}{true negative}
\newacronym{tp}{TP}{true positive}
\newacronym{fn}{FN}{false negative}
\newacronym{fp}{FP}{false positive}
\newacronym{ig}{IG}{Integrated Gradients}

\usepackage[noadjust]{cite}

\usepackage{booktabs,makecell}
\usepackage{placeins}
\usepackage{gensymb} 
\usepackage{multirow}
\usepackage{cleveref}

\usepackage{algorithm}
\usepackage{algpseudocode}
\usepackage{amsmath} 

\newcommand{\aer}{AER}
\newcommand{\core}{CORE}
\newcommand{\db}{DB}
\newcommand{\aercore}{\aer{}+\core{}}
\newcommand{\aercoredb}{\aercore{}+\db}

\definecolor{hl-blue}{RGB}{0, 0, 180}
\definecolor{hl-green}{RGB}{0, 128, 0}
\definecolor{hl-red}{RGB}{196, 0, 0}
\hypersetup{
    colorlinks=true,       
    linkcolor=hl-red,      
    citecolor=hl-blue,     
    filecolor=magenta,     
    urlcolor=hl-green         
}

\usepackage[scaled=0.85]{DejaVuSansMono}
\usepackage{listings}
\definecolor{codegreen}{rgb}{0,0.6,0}
\definecolor{codegray}{rgb}{0.5,0.5,0.5}
\definecolor{codepurple}{rgb}{0.58,0,0.82}
\definecolor{backcolour}{rgb}{0.95,0.95,0.95}

\lstdefinestyle{mystyle}{
    backgroundcolor=\color{backcolour},   
    commentstyle=\color{codegreen},
    keywordstyle=\color{magenta},
    numberstyle=\tiny\color{codegray},
    stringstyle=\color{codepurple},
    identifierstyle=\color{black},
    basicstyle=\ttfamily\footnotesize,
    rulecolor=\color{gray},
    frameround=tttt,
    frame=single,
    xleftmargin=14pt,
    xrightmargin=4pt,
    breakatwhitespace=false,         
    breaklines=true,                 
    captionpos=b,                    
    keepspaces=true,                 
    numbers=left,                    
    numbersep=8pt,                  
    showspaces=false,                
    showstringspaces=false,
    showtabs=false,                  
    tabsize=2
}
\begin{document}




\title{Weakly Supervised Polar Low Segmentation in Sentinel-1 SAR Imagery
}

\author{Andrea Federici, 
        Jakob Grahn,
        Giacomo Boracchi,
        Filippo Maria Bianchi$^{*}$
\thanks{*filippo.m.bianchi@uit.no}
\thanks{A. Federici and F. M. Bianchi are with the Dept. of Mathematics and Statistics, UiT the Arctic University of Norway}
\thanks{G. Boracchi is with DEIB, Politecnico di Milano}
\thanks{J. Grahn and F. M. Bianchi are with NORCE, The Norwegian Research Centre AS}%
}

\maketitle

\begin{abstract}
Polar lows are intense maritime cyclones that form rapidly at high latitudes.
Deep learning can detect them in \gls{sar} imagery, but pixel-level segmentation remains an open challenge.
No pixel-level masks are available for training, and a polar low's extent is inherently subjective, with diffuse boundaries that even experts delineate inconsistently.
We propose \gls{crest}, a \gls{wsss} framework that generates masks solely from image-level labels.
Our approach builds on \gls{aer}, which iteratively mines discriminative regions, erases them, and retrains a classifier to reveal complementary cues that become pseudo-labels for segmentation.
However, standard \gls{aer} also collects irrelevant background features, degrading pseudo-label quality.
\gls{crest} addresses this with (i) a \gls{core} module that encodes the spatial-connectedness prior of polar lows, constraining region expansion from a high-confidence seed, and (ii) a \gls{db} loss that treats the mining order as a proxy for label reliability, attenuating supervision from noisier, later-mined regions.
On Sentinel-1 \gls{sar} data, \gls{crest} follows the cyclone structure more closely than standard \gls{aer}, and returns a multi-class rather than binary mask whose classes indicate the reliability assigned to each region.
We further evaluate on BUS-UCLM breast ultrasound and PASCAL VOC person data, whose targets satisfy the same connectedness prior but come with the dense masks the \gls{sar} data lacks.
On both datasets, \gls{crest} performs better than the equivalent \gls{aer} pipeline under identical settings.
\end{abstract}

\begin{IEEEkeywords}
Weakly Supervised Segmentation; Polar Lows; SAR; Adversarial Erasing
\end{IEEEkeywords}

\glsresetall


\section{Introduction}\label{sec:intro}

\begin{figure}[!tbp]
    \centering
    \begin{subfigure}[b]{0.48\columnwidth}
        \centering
        \includegraphics[width=\linewidth]{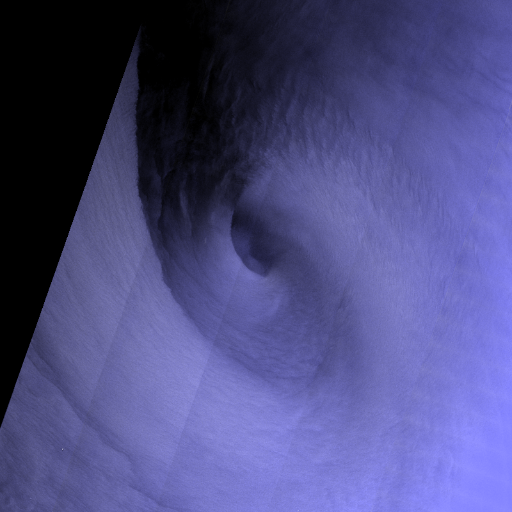}
        \caption{Positive (polar low present)}
        \label{fig:teaser_pos}
    \end{subfigure}\hfill
    \begin{subfigure}[b]{0.48\columnwidth}
        \centering
        \includegraphics[width=\linewidth]{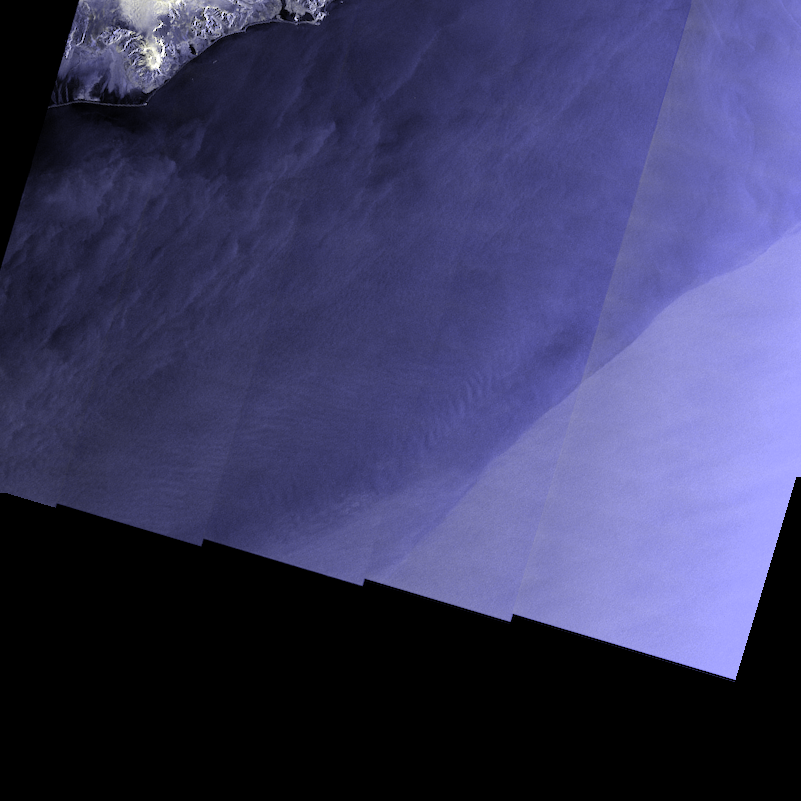}
        \caption{Negative (Background)}
        \label{fig:teaser_neg}
    \end{subfigure}
    \caption{Examples of Sentinel-1 \gls{sar} imagery from our dataset. (a) A positive sample showing a polar low. (b) A negative sample showing a typical maritime background without cyclonic activity. The black regions are ``no-data'' areas in the \gls{sar} images.}
    \label{fig:teaser}
\end{figure}

Polar lows are small but intense maritime cyclones with horizontal scales of approximately $200$--$1{,}000$ km and lifetimes typically shorter than a day~\cite{Rasmussen_Turner_2003}.
They form at high latitudes over open water, often close to sea ice or snow-covered land.
Their formation is driven by atmospheric instability caused by cold air moving over warmer oceans~\cite{montgomery1992polar}.
Promptly detecting and localizing polar lows is critical, as they pose risks to maritime activities~\cite{Rasmussen_Turner_2003} and, when making landfall, can disrupt land and air traffic, damage infrastructure, and trigger avalanches~\cite{Grahn_2022}.
However, polar lows are also difficult to observe because they frequently occur during the polar night, which prevents the use of visible-light sensors.
On the other hand, high-resolution \gls{sar} data from \gls{esa}'s Sentinel-1 do not depend on sunlight and provide day-and-night coverage, making them well suited to our setting~\cite{Grahn_2022} (\Cref{fig:teaser}).
The dataset's positive class comprises maritime mesocyclones, the broader meteorological class to which polar lows belong~\cite{Grahn_2022}. Throughout the paper, ``polar low'' denotes the positive class in this dataset.

In recent years, machine learning has been increasingly adopted to analyze and track cyclones~\cite{alet2026operational} and polar lows~\cite{yang2026polar}.
Deep learning architectures for computer vision, such as \glspl{cnn}, have been successfully employed to detect polar lows in \gls{sar} imagery with high accuracy~\cite{Grahn_2022}.
Determining whether a polar low is present is treated as an image classification task~\cite{Grahn_2022}, while producing a spatial mask can be framed as image segmentation, i.e., pixel-wise classification~\cite{minaee2021image}.
In this work, the target mask represents the \gls{sar}-visible footprint of the cyclone, i.e., the area where its wind pattern on the sea surface is visible in the image.
Such a spatial output is more informative than an image-level prediction.
However, training semantic segmentation networks typically requires dense pixel-level annotations from an expert. 
In contrast, training a classification network only requires image-level labels, which are much faster and cheaper to obtain. 
In our case study, however, the obstacle is not only cost.
The \gls{sar}-visible footprint has diffuse and irregular boundaries that experts might delineate inconsistently because there are no broadly accepted definitions of where a polar low ends.
Those diffuse boundaries also affect the methods that could be used to automate the annotation procedure, since refinement pipelines and promptable foundation models alike rely on strong intensity edges and clear foreground-background contrast (\Cref{sec:rw}).
Our aim is therefore not to compete with such methods on the benchmarks where they excel, but to obtain spatial labels in a setting where they do not apply.

This raises the question of whether the pixel-level masks needed for segmentation can be obtained from image-level labels alone.
Feature attribution methods provide a possible route: they explain a classifier by assigning importance scores to input regions according to their contribution to a prediction~\cite{ancona2018towards_unified,montavon2019lrp}.
These scores are commonly visualized as saliency maps or heatmaps and, for binary classification, can in principle localize the target object~\cite{hohl2024opening,GradCAM2019}.
A prominent research direction therefore thresholds and refines such heatmaps into pixel-level masks, which are used either directly as segmentations or as pseudo-labels for training a segmentation model, using only image-level labels and without manual pixel-level annotations~\cite{ahlswede2022weakly, Forest_2024, wei2025weakly}.
However, such heatmaps are typically coarse, and earlier applications to polar low \gls{sar} imagery primarily captured a few characteristic cues (e.g., the cyclone eye) while overlooking boundaries and secondary structures~\cite{Grahn_2022}.

The limited spatial coverage of a single saliency map motivates an iterative search for less discriminative object regions.
\gls{aer} provides such a mechanism and is an established weakly supervised approach for generating pixel-level pseudo-labels from image-level labels~\cite{wei2018objectregionminingadversarial}.
In its standard formulation, an initial \gls{cnn} classifier is trained using image-level labels. \gls{aer} then iteratively ``erases'' the most discriminative regions identified by saliency maps and retrains the classifier on the modified images.
Each iteration forces the \gls{cnn} to learn new, less salient features, gradually expanding its focus from the most discriminative regions to those with secondary structures. After thresholding and aggregating the saliency maps obtained across iterations, the mined regions can thus be converted into pseudo-labels that provide pixel-level supervision for a segmentation network, which can then operate directly on new images without repeating the iterative erasure process.

\begin{figure}[!tbp]
  \centering
  \setlength{\tabcolsep}{3pt} 
  \begin{tabular}{@{}cc@{}}
    \small \textbf{Original image}                                                                                                      & \small \textbf{Grad-CAM activation} \\
    \includegraphics[width=.46\columnwidth]{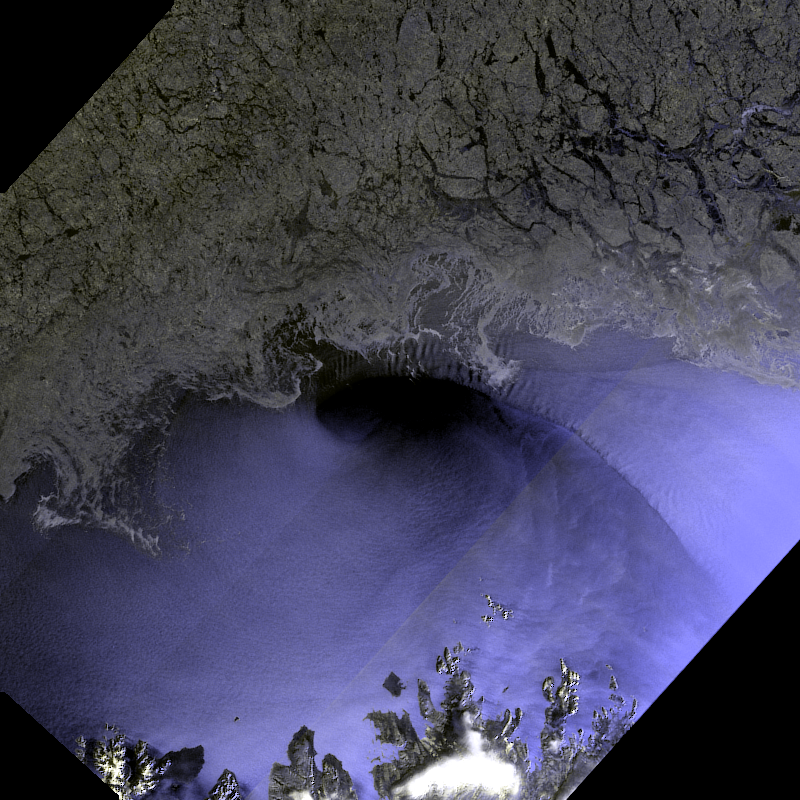} &
    \includegraphics[width=.46\columnwidth]{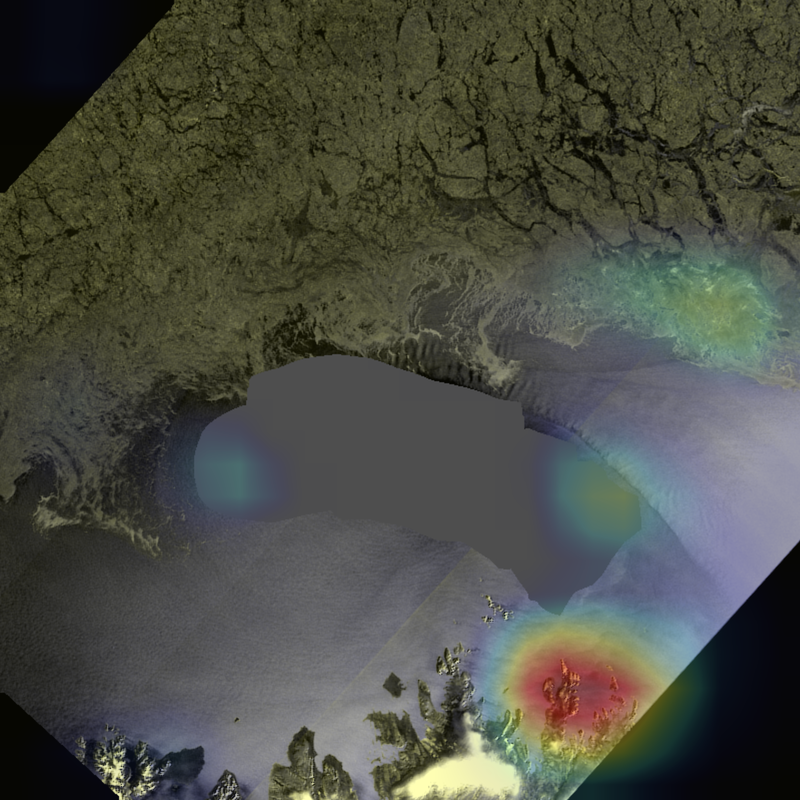}                                                                                   \\[3pt]
    \includegraphics[width=.46\columnwidth]{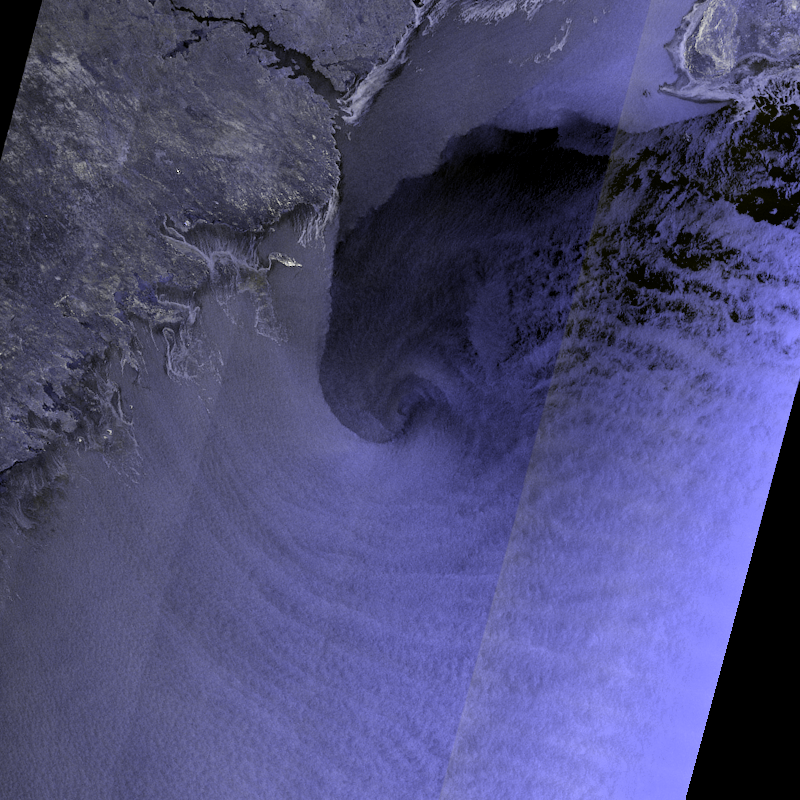} &
    \includegraphics[width=.46\columnwidth]{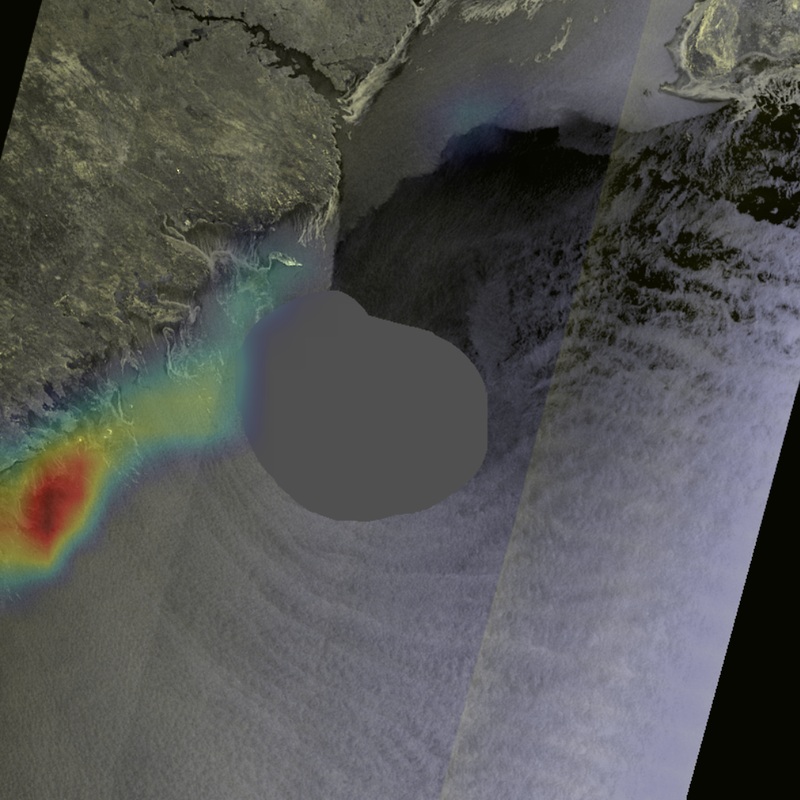}                                                                                   \\
  \end{tabular}
  \caption{Spurious background activations during late \gls{aer} iterations.
    Each row shows a different positive polar low \gls{sar} image.
    Left: original input. Right: Grad-CAM on the image after erasing mined regions.
    With most of the target regions removed, the classifier starts producing activations on the background, injecting noise into the aggregated pseudo-label.}
  \label{fig:ae_spurious_2x2}
\end{figure}

Although vanilla \gls{aer} extends segmentation beyond the most discriminative regions identified by a single application of a saliency method, it suffers from two main limitations.
\begin{itemize}
    \item Classifiers trained in early iterations capture the most salient parts of the cyclone. 
    Later classifiers may instead focus on spurious background features, especially when few cyclone cues remain, as shown in~\Cref{fig:ae_spurious_2x2}. 
    This effect is exacerbated when polar lows have different apparent sizes and are therefore erased at different stages, since smaller cyclones are fully removed after a few iterations while larger ones require more.
    \item Standard \gls{aer} treats all iterations equally, merging all discovered regions into a single binary mask. This mixes reliable early-discovered regions with less reliable ones discovered in later stages.
    Treating reliable and unreliable patterns identified at different stages equally degrades the quality of supervision offered by the pseudo-labels for training a segmentation network.
\end{itemize}

\begin{figure}[!tbp]
    \centering
    \begin{subfigure}[t]{0.32\linewidth}
        \centering
        \includegraphics[width=\textwidth]{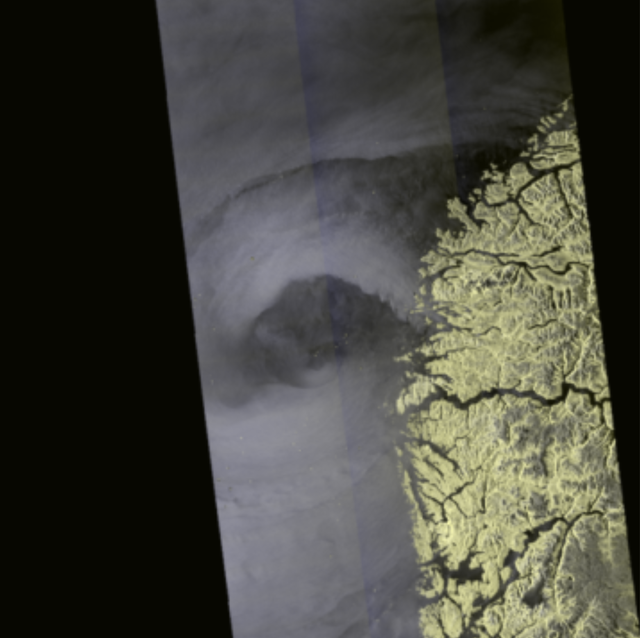}
        \caption{Input Image}
        \label{fig:core_input}
    \end{subfigure}
    \hfill
    \begin{subfigure}[t]{0.32\linewidth}
        \centering
        \includegraphics[width=\textwidth]{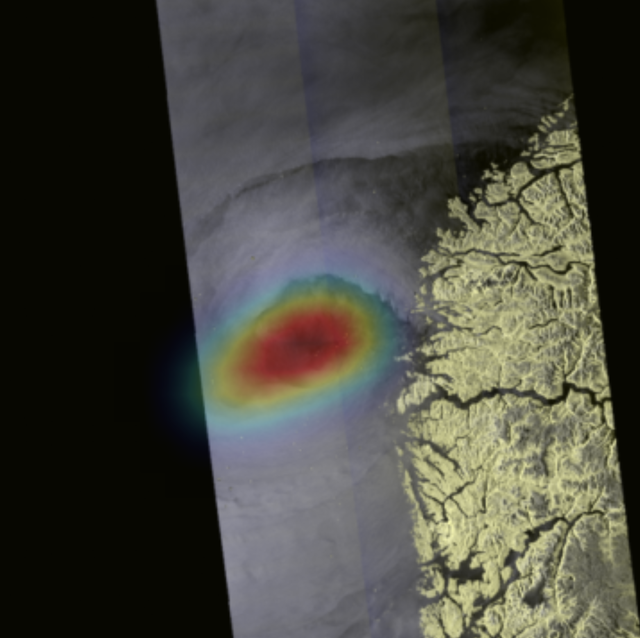}
        \caption{Iter 0}
        \label{fig:core_seed}
    \end{subfigure}
    \hfill
    \begin{subfigure}[t]{0.32\linewidth}
        \centering
        \includegraphics[width=\textwidth]{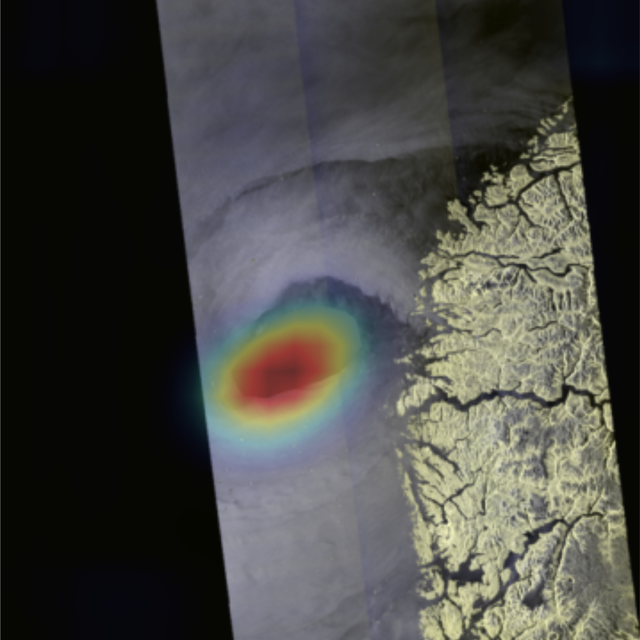}
        \caption{Iter 1}
        \label{fig:core_env}
    \end{subfigure}

    \vspace{0.5em}

    \begin{subfigure}[t]{0.32\linewidth}
        \centering
        \includegraphics[width=\textwidth]{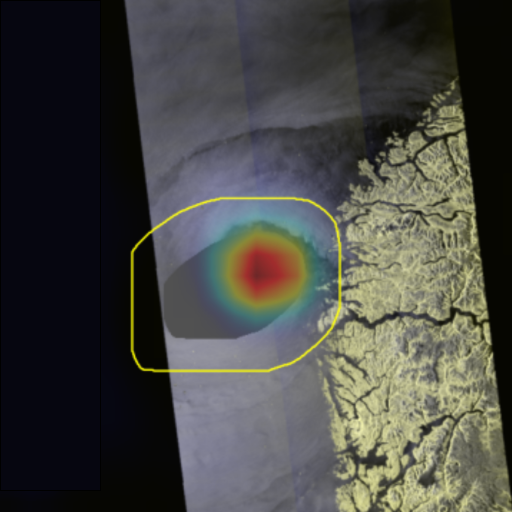}
        \caption{Iter 2}
        \label{fig:core_raw}
    \end{subfigure}
    \hfill
    \begin{subfigure}[t]{0.32\linewidth}
        \centering
        \includegraphics[width=\textwidth]{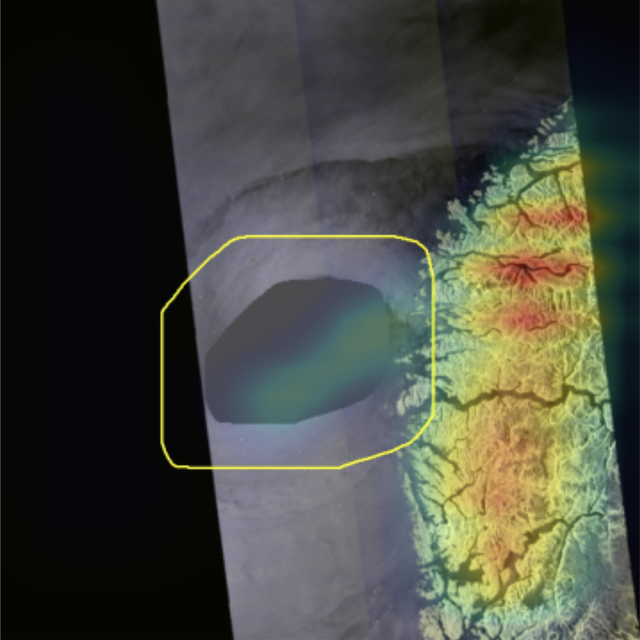}
        \caption{Iter 3}
        \label{fig:core_iter1}
    \end{subfigure}
    \hfill
    \begin{subfigure}[t]{0.32\linewidth}
        \centering
        \includegraphics[width=\textwidth]{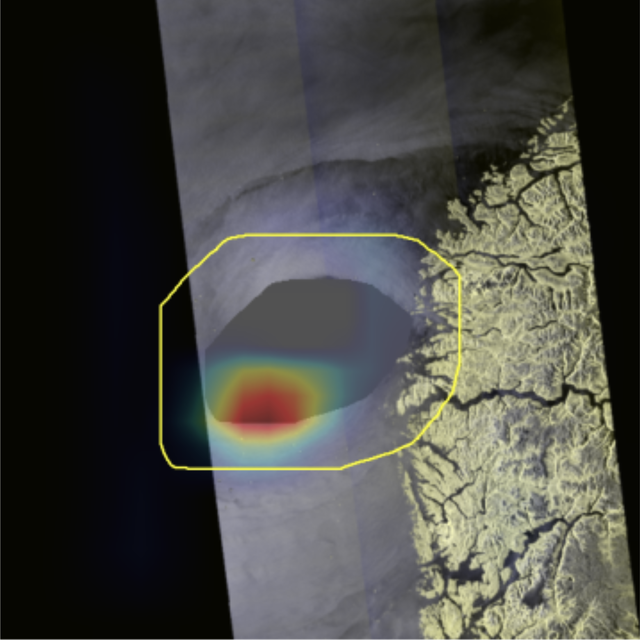}
        \caption{Iter 4}
        \label{fig:core_final}
    \end{subfigure}

    \caption{\textbf{Impact of \glsentryshort{core} on region expansion.} The panels display the Grad-CAM saliency maps at each step of the mining process.
    Once activated, \glsentryshort{core} introduces the yellow local envelope that limits subsequent mask growth.
    In Iteration 3, the classifier produces a strong activation on land outside this envelope; the activation is therefore excluded from the next mask update, as shown in Iteration 4.}
    \label{fig:core_evolution}
\end{figure}

\begin{figure}[!htbp]
    \centering
    \includegraphics[
        width=\columnwidth,
        keepaspectratio
    ]{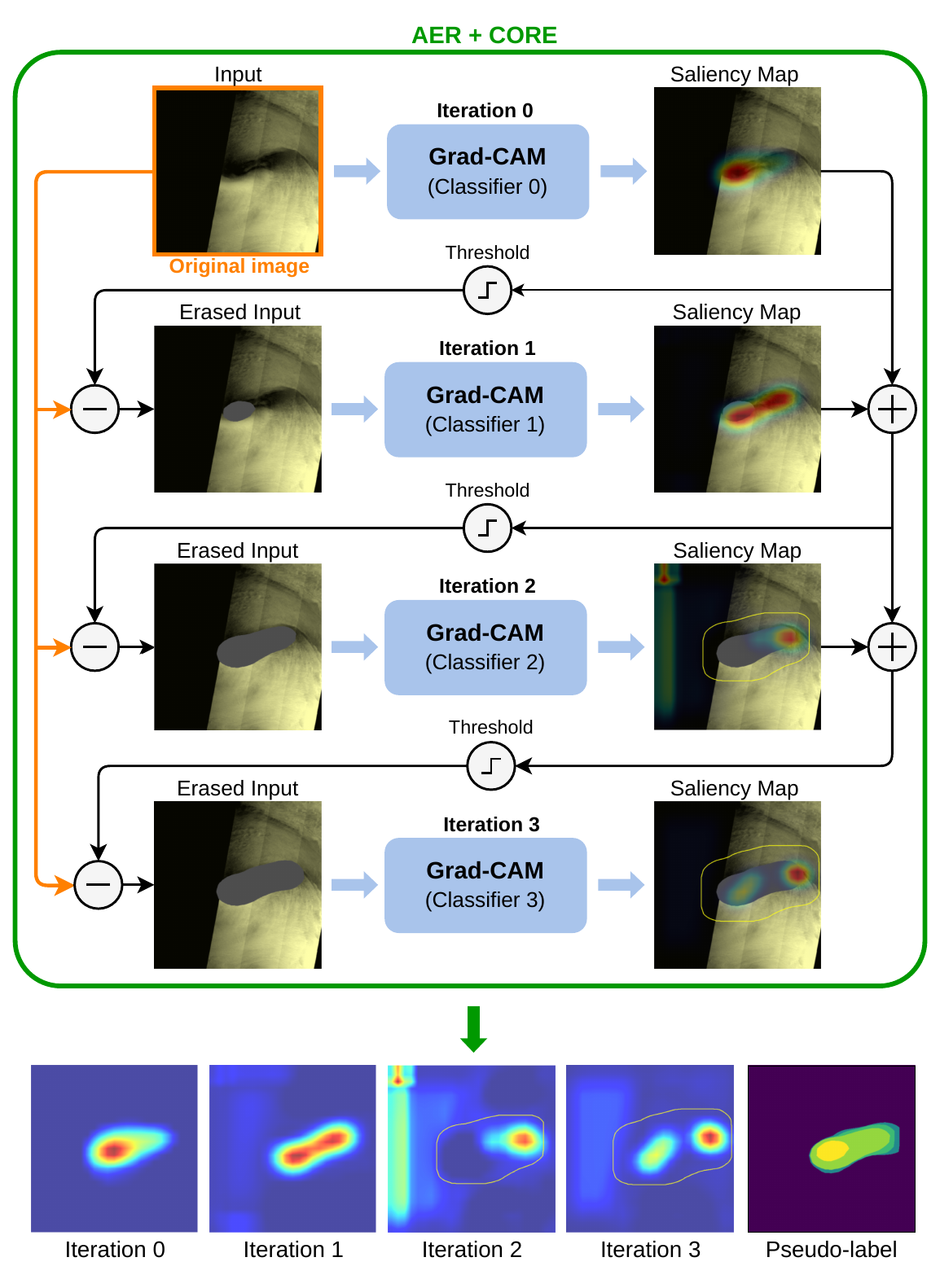}
    \caption{\aercore{} pipeline and pseudo-label construction. Top: the proposed iterative train $\rightarrow$ extract $\rightarrow$ erase loop. 
    We retrain the classifier after each erasing step and recompute Grad-CAM on the updated image. 
    Once activated, \gls{core} restricts the next erasure to a neighborhood (yellow envelope) around the previously mined region. 
    Bottom: Grad-CAM heatmaps from successive iterations and the resulting multi-class pseudo-label, where the class records when each region was mined.}
    \label{fig:adver_process}
\end{figure}

To address \gls{aer}'s limitations, we introduce \gls{crest}, which combines two main contributions.
First, we add a new module, called \gls{core}, designed to regulate the iterative growth of discovered regions through a spatial proximity constraint.
The mechanism implemented by \gls{core} is illustrated in \Cref{fig:core_evolution}.
After activation, \gls{core} excludes newly mined pixels outside a local envelope around the accumulated support, which are the ones that typically correspond to spurious background features.
This design fits the polar low setting, where the \gls{sar}-visible pattern is typically compact, and its peripheral regions can be found by expanding locally around the initial seeds.
Figure~\ref{fig:adver_process} (top) summarizes the whole iterative train-extract-erase loop.

Second, we train the segmentation network with a \gls{db} loss that relaxes supervision where the pseudo-labels are least reliable.
To expose that reliability, we preserve the iteration in which \gls{aer} mines each region as a mining-order tier, moving from binary to multi-class pseudo-labels.
Figure~\ref{fig:adver_process} (bottom) illustrates how heatmaps progressively expand the coverage over iterations to form the final pseudo-label.

For early-mined pixels, which are assumed to be more reliable, the loss supervises the network mostly through the pseudo-label.
For later-mined pixels, it progressively attenuates cross-entropy supervision from the pseudo-label, reducing the influence of potentially noisy regions.

Our main focus is polar low segmentation in Sentinel-1 \gls{sar} imagery. 
Since pixel-level ground-truth masks are unavailable in this domain, we compare the predictions qualitatively against the baselines across test scenes and place the mined tiers in physical context by overlaying the \gls{sar} acquisition with the meteorological fields.
Then, we evaluate \gls{crest} on the \gls{bus} and \gls{voc} datasets.
We select both datasets because their targets satisfy the same spatial-connectedness prior that motivates \gls{core}, namely that each object occupies a coherent region whose peripheral parts are reached by expanding locally from an initial seed.
Ultrasound lesions form compact, connected masses, and in \gls{voc} we restrict the evaluation to the ``person'' class, whose instances are likewise spatially connected, which also preserves a binary foreground-background setting comparable to the polar low task.
Unlike the polar low dataset, these two provide pixel-level masks, which we use only for evaluation and never for training.
They let us measure our pipeline against the \gls{aer} baseline quantitatively, a comparison the polar low data cannot support for lack of dense ground truth.

\section{Background and Related Work}
\label{sec:rw}

\subsection{Deep Learning for Polar Low Detection}
The detection and analysis of polar lows have traditionally relied on manual inspection of satellite imagery~\cite{Rasmussen_Turner_2003} or on threshold-based rules applied to reanalysis data~\cite{kolstad2011global}.
Compared to these traditional procedures, deep learning provides data-driven approaches for automated detection and tracking of polar lows across different data modalities.
Early works, such as Krinitskiy et al.~\cite{krinitskiy2018deep}, showed that \glspl{cnn} can perform binary polar low classification on satellite mosaics.
A recent review notes that deep learning has been used to detect maritime polar mesoscale cyclones in satellite imagery, and that \gls{sar} observations, despite resolving the atmospheric fronts and cyclonic centers that identify these systems, have so far rarely been exploited for polar lows~\cite{moreno2024polar}.
For instance, Zhang et al.~\cite{zhang2024automatic} combined \gls{sar} and radiometer observations to support automated tracking, which illustrates the value of multi-modal inputs.

Visible imagers depend on sunlight, whereas thermal-infrared sensors, passive-microwave radiometers, and microwave scatterometers can provide nighttime observations but generally have coarser spatial resolution than Sentinel-1 \gls{sar}. 
\gls{sar} is independent of solar illumination and provides the high spatial resolution useful for examining mesoscale wind patterns.
In this context, Grahn and Bianchi~\cite{Grahn_2022} introduced the first publicly available Sentinel-1 \gls{sar} dataset for polar low classification and established a baseline using \glspl{cnn}.
Their study indicates that \gls{sar} imagery contains enough information to distinguish polar low patterns from other atmospheric phenomena.

\begin{figure}[!t]
    \centering
    \begin{subfigure}[b]{0.32\columnwidth}
        \centering
        \includegraphics[width=\linewidth, keepaspectratio]{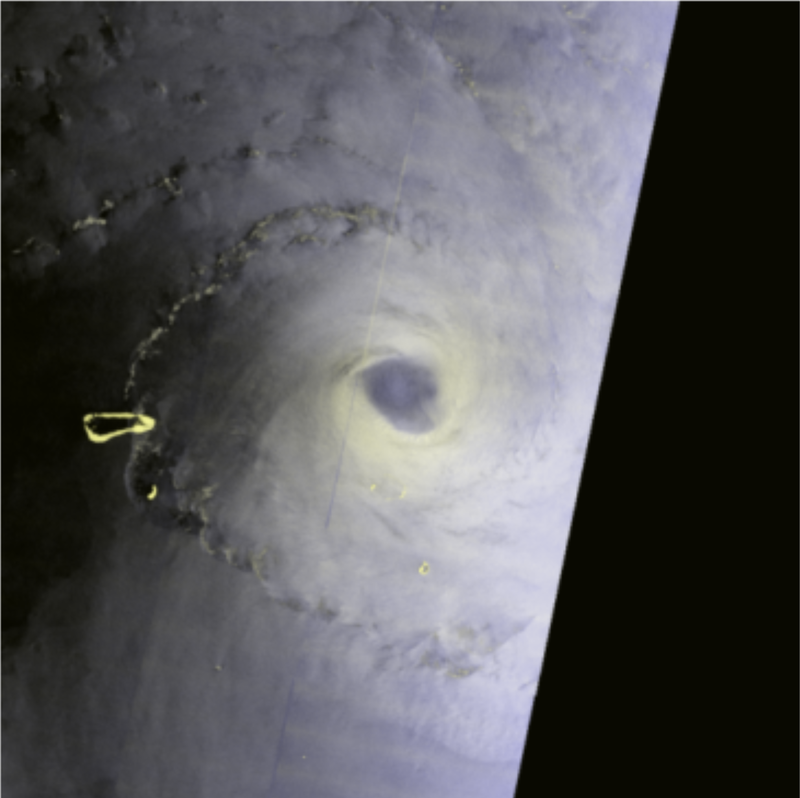}
        \caption{Original SAR image}
    \end{subfigure}
    \hfill
    \begin{subfigure}[b]{0.32\columnwidth}
        \centering
        \includegraphics[width=\linewidth, keepaspectratio]{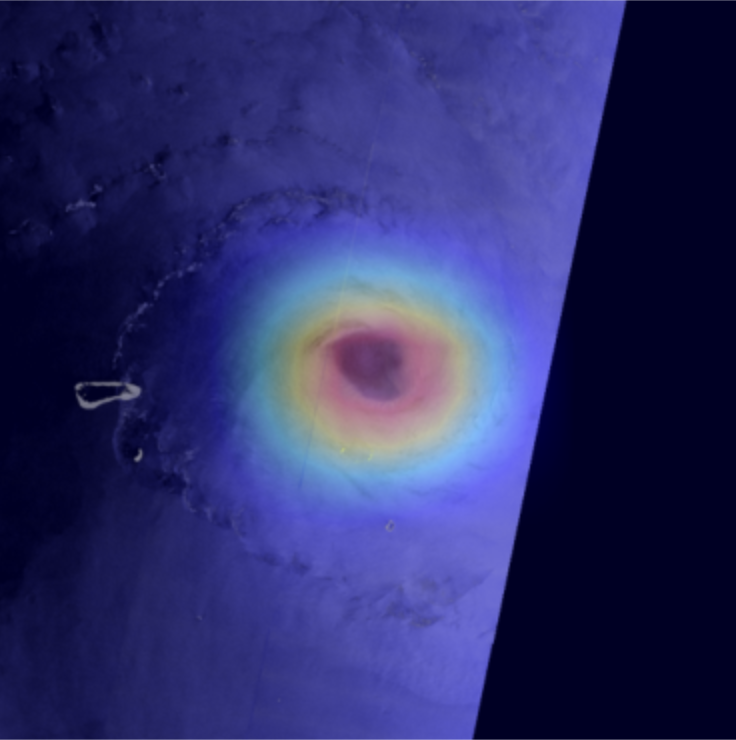}
        \caption{Grad-CAM}
    \end{subfigure}
    \hfill
    \begin{subfigure}[b]{0.32\columnwidth}
        \centering
        \includegraphics[width=\linewidth, keepaspectratio]{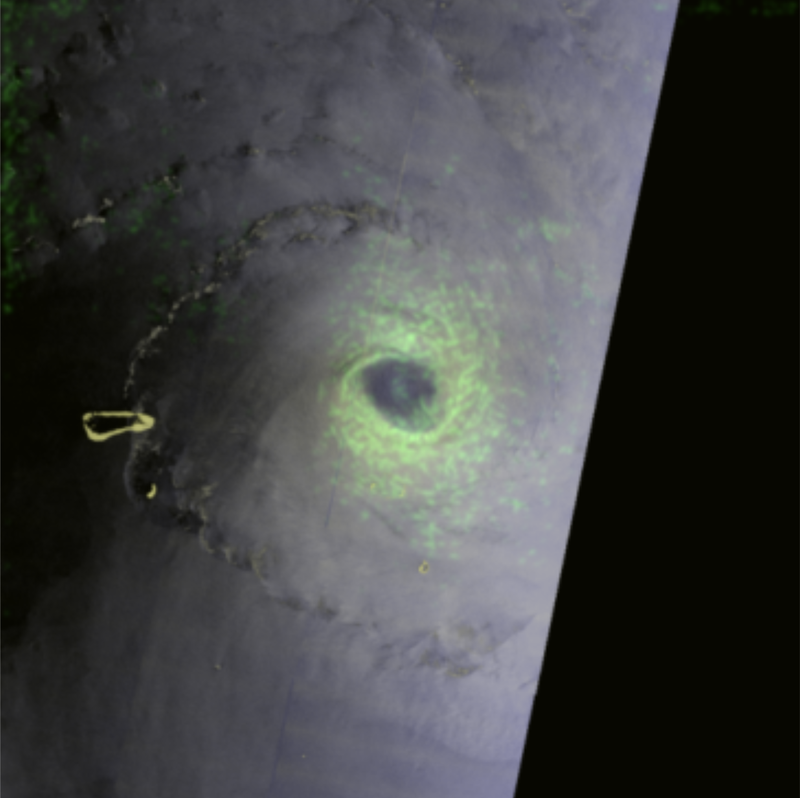}
        \caption{Integrated Gradients}
    \end{subfigure}
    \caption{Grad-CAM and Integrated Gradients on a Sentinel-1 SAR image. In this example, both methods highlight the most discriminative cues but not the full spatial extent of the cyclone.}
    \label{fig:attribution_comparison}
\end{figure}

To interpret the \gls{cnn} classifier decisions, Grahn and Bianchi used \gls{xai} methods, including \gls{gradcam}~\cite{GradCAM2019} and Integrated Gradients~\cite{integratedgradients2017}, to produce heatmaps that indicate which image regions contribute the most to the predicted class (\Cref{fig:attribution_comparison}).
However, a model trained for image-level classification is not required to account for the target pattern across its full spatial extent.
It may instead focus on a few discriminative cues, such as the cyclone eye or strong wind fronts, while neglecting weaker peripheral structures.
Consequently, masks obtained from individual saliency maps are often coarse and incomplete, which limits their usefulness as pixel-level supervision.

\subsection{Weakly Supervised Segmentation and Foundation Models}
The core objective of \gls{wsss} is to derive dense segmentation masks from weak supervision, which typically comes in the form of image-level labels. 
Many established approaches start by thresholding a saliency map and refining it through post-processing.
For example, Forest et al.~\cite{Forest_2024} refine saliency maps using thresholding and morphological closing to fill small gaps and connect nearby activated regions.
Other methods combine seed expansion with a boundary-aware constraint, as in the Seed-Expand-Constrain framework~\cite{kolesnikov2016sec}.

\begin{figure}[!t]
  \centering
  \begin{subfigure}[t]{0.32\columnwidth}
    \includegraphics[width=\linewidth, keepaspectratio]{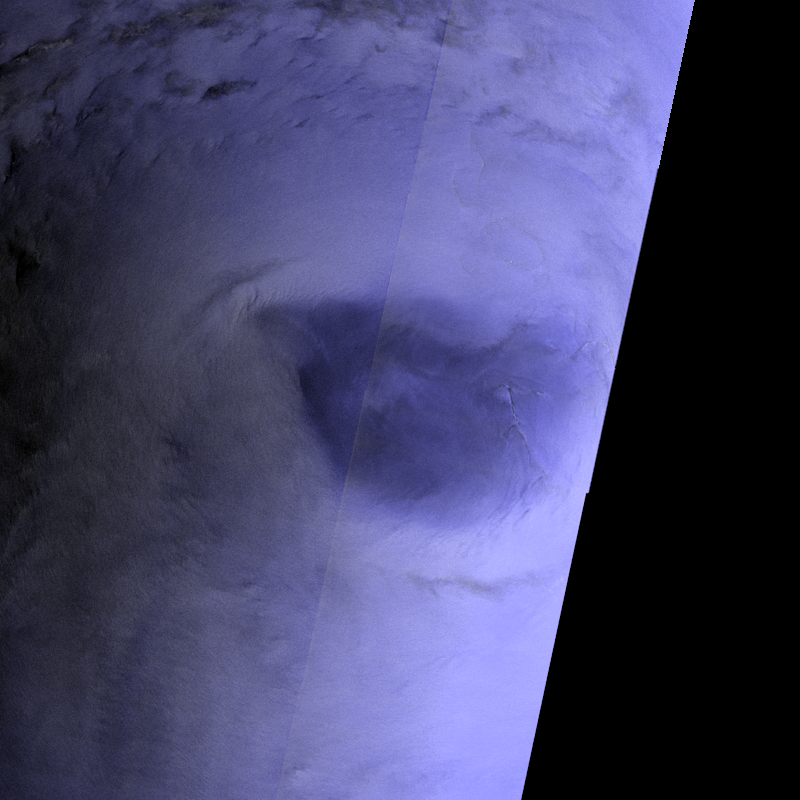}
    \caption{Sentinel-1 SAR input image.}
  \end{subfigure}\hfill
  \begin{subfigure}[t]{0.32\columnwidth}
    \includegraphics[width=\linewidth, keepaspectratio]{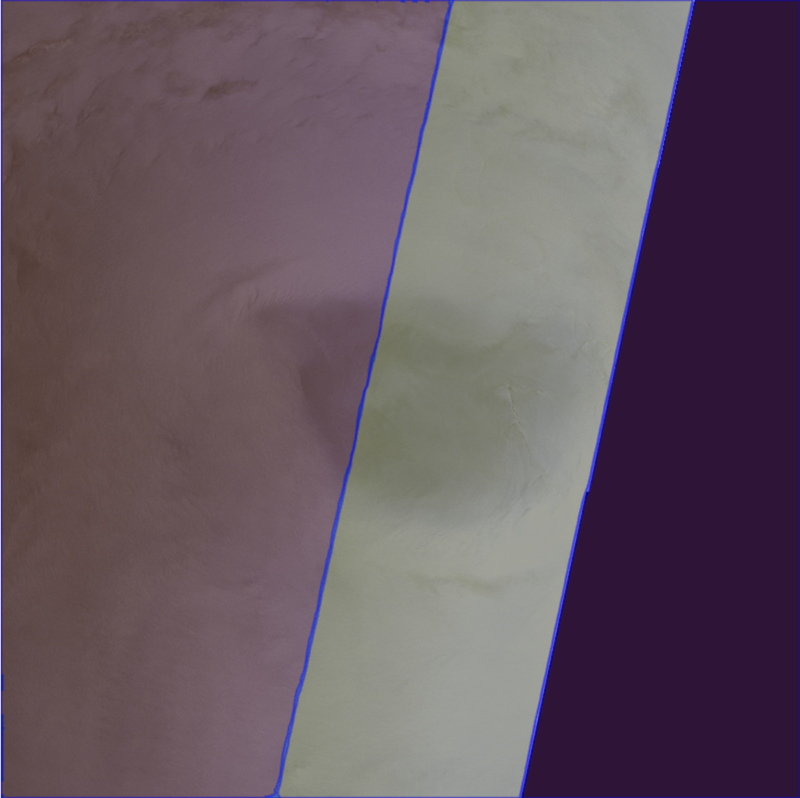}
    \caption{Grid-based prompting highlights satellite imaging artifacts.}
  \end{subfigure}\hfill
  \begin{subfigure}[t]{0.32\columnwidth}
    \includegraphics[width=\linewidth, keepaspectratio]{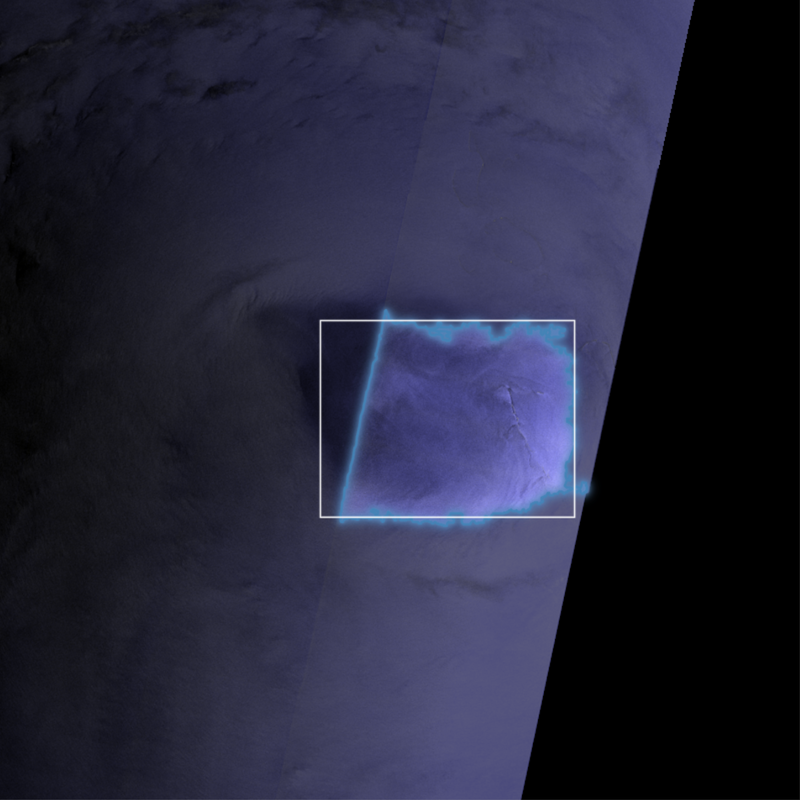}
    \caption{In this example, box-based prompting misses the diffuse cyclone structure.}
  \end{subfigure}
\caption{Zero-shot \glsentryshort{sam} outputs for one Sentinel-1 polar low \gls{sar} image. With grid- and box-based prompts, the model highlights imaging artifacts or misses the diffuse cyclone structure.}
\label{fig:sam}
\end{figure}

Foundation models like the \gls{sam}~\cite{kirillov2023segany, ravi2024sam2} provide interactive segmentation capabilities on natural images, and recent work explores their use in remote sensing~\cite{gelato2026promptable}.
However, both classical refinement methods and promptable segmentation models generally rely on clear boundaries and strong foreground-background contrast. This assumption does not hold for \gls{sar} imagery of atmospheric phenomena such as polar lows, whose diffuse boundaries fade gradually into the background.
In our zero-shot \gls{sam} experiment using grid- and box-based prompts, the model responded to high-contrast imaging artifacts (e.g., satellite scan lines) rather than the diffuse cyclone structure (\Cref{fig:sam}).
Box- and scribble-supervised methods that use boundary cues to expand sparse labels may face similar challenges in this setting~\cite{dai2015boxsup,lin2016scribblesup,tang2018scribbleseg,zhang2025soft}.

\subsection{Adversarial Erasing}
\gls{aer}~\cite{wei2018objectregionminingadversarial} methods are specifically designed to address the incomplete localization of classification-based saliency maps.
In \gls{aer}, a classifier is trained to identify discriminative object regions through saliency maps. 
These regions are then erased from the input images, and the classifier is retrained on the modified dataset. 
This cycle forces the classifier to discover new, less discriminative features of the object that were previously overshadowed by more salient patterns. 
By aggregating the saliency maps generated across several iterations, one can obtain more complete object coverage than with a single saliency map.
Standard \gls{aer} pipelines often refine segmentation outputs with \glspl{crf}~\cite{wei2018objectregionminingadversarial, chen2017deeplab, krahenbuhl2011crf}. 
The original \gls{aer} method also uses Prohibitive Segmentation Learning to weight class contributions~\cite{wei2018objectregionminingadversarial}.
For polar low \gls{sar} imagery, diffuse cyclone boundaries provide weak cues for edge-based refinement, and we therefore omit this stage throughout, as detailed in \Cref{sec:exp}.
In addition, naive aggregation across iterations can inject noise since, when most of the informative regions are erased, the classifier may attend to background texture or imaging artifacts, which produces spurious activations (\Cref{fig:ae_spurious_2x2}).
Thus, there are two key challenges for extending \gls{aer}-based methods to polar low segmentation: limiting spurious region growth across iterations and reducing the influence of later, potentially noisy discoveries.
\section{Problem Formulation}\label{sec:pf}

We consider the following dataset of image-label pairs:
\[\mathcal{D} = \{(\mathbf{X}_i,c_i)\ \mid\ \mathbf{X}_i\in \mathbb{R}^{W\times H\times 3},\ c_i\in \{0,1\},\ i=1,\ldots,\lvert\mathcal{D}\rvert\},\]
where each $\mathbf{X}_i$ is an RGB image of width $W$ and height $H$. 
Although the source \gls{sar} images have only two channels, we use the RGB composites provided with the dataset, which were generated following the preprocessing procedure described in~\cite{Grahn_2022} (see \Cref{app:dataset} for details about the dataset).
The image-level label $c_i$ indicates the presence ($c_i=1$) or absence ($c_i=0$) of a polar low.
In the following, we refer to images containing a polar low as \textit{positive}, and those without as \textit{negative}.
For a positive image, the region to be segmented is the area where the cyclone's sea-surface wind pattern is visible in the \gls{sar} image.

Let $N$ be the maximum number of iterations in the \gls{aer} procedure.
The goal is to use only the image-level labels in $\mathcal{D}$, without any pixel-level annotation, to train a segmentation model
\[f_\theta: \mathbb{R}^{W \times H \times 3} \rightarrow \{0, \ldots, N\}^{W \times H}\]
that maps each input image $\mathbf{X}_i$ to a mining-order map $\mathbf{S}_i = f_\theta(\mathbf{X}_i)$, where class 0 denotes background, and classes from 1 to $N$ record when foreground pixels were mined, with larger values indicating earlier discovery.

\section{Methodology}\label{sec:ps}

\begin{figure}[!t]
    \centering
    \includegraphics[
        width=\columnwidth,
        keepaspectratio
    ]{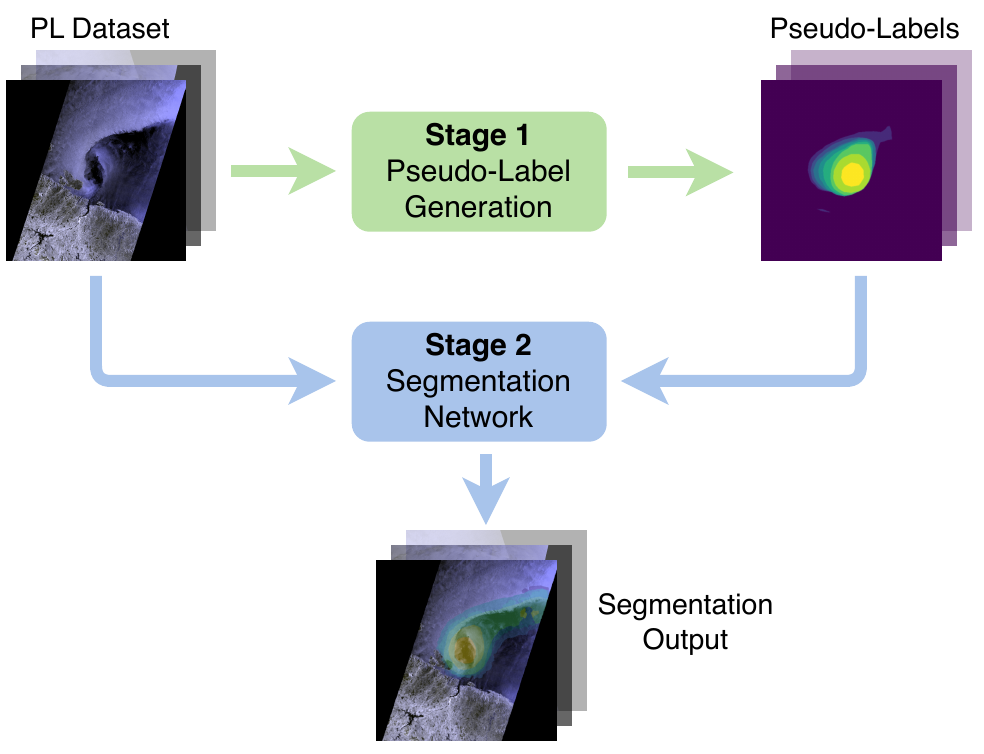}
    \caption{Overview of the proposed two-stage pipeline. Stage 1 generates mining-order pseudo-labels from image-level labels via \gls{aer} with the \gls{core} module. Stage 2 trains a segmentation network using these pseudo-labels and a \acrlong{db} strategy to handle label uncertainty.}
    \label{fig:overview}
\end{figure}

In this section, we present \gls{crest}, our two-stage pipeline for weakly supervised polar low segmentation in \gls{sar} imagery.
A single discriminative classifier responds only to the most prominent parts of the target, such as the cyclone eye, which is why the mining procedure is made iterative. Iteration in turn raises two challenges: the discovered region can expand into unrelated background, and regions found in later iterations provide progressively less reliable supervision.
To address them, \gls{crest} implements a two-stage procedure illustrated in \Cref{fig:overview}.
\textbf{Stage 1} generates pixel-level pseudo-labels from image-level labels using the proposed \aercore{}, an enhanced \gls{aer} procedure in which \gls{core} constrains the iterative mining process.
\textbf{Stage 2} trains a dedicated segmentation network using the generated pseudo-labels and a \gls{db} strategy to handle their inherent noise and uncertainty.
The two stages are detailed below.

\subsection{Stage 1: Pseudo-Label Generation}

This stage transforms weak image-level annotations into dense pixel-level mining-order pseudo-labels.
We perform $N$ mining iterations in a train-extract-erase loop: each iteration trains a classifier and extracts a mask, while an erasing step prepares the data for the next iteration.
The iteration at which each pixel is mined determines its mining-order tier.
Class $0$ denotes background, while foreground tiers $1,\dots,N$ correspond to regions mined at different iterations; higher tier values indicate earlier discoveries, to which we assign greater reliability.
The loop is regularized by our \gls{core} module, which activates at iteration $t_{\mathrm{core}}$ and restricts newly admitted pixels to a local envelope around the previously mined support, preventing expansion into distant background regions.
The procedure is summarized in \Cref{alg:core_pipeline}.

\begin{algorithm}[!t]
    \caption{Adversarial Erasing + CORE}
    \label{alg:core_pipeline}

    \renewcommand{\algorithmicrequire}{\textbf{Input:}}
    \renewcommand{\algorithmicensure}{\textbf{Output:}}

    \begin{algorithmic}[1]
        \Require Dataset $\mathcal{D}=\mathcal{D}^+\cup\mathcal{D}^-$ with positive images $\mathcal{D}^+$ and negative images $\mathcal{D}^-$, iterations $N$, threshold $\tau$, CORE start $t_{\mathrm{core}}$, \gls{core} expansion factor $\kappa$, mean RGB vector $\boldsymbol{\mu}\in\mathbb{R}^3$
        \Ensure Set of pseudo-labels $\mathcal{P} = \{\mathbf{P}_i \mid \mathbf{X}_i \in \mathcal{D}\}$, where $\mathbf{P}_i=\mathbf{0}$ for $\mathbf{X}_i\in\mathcal{D}^-$
        \item[\textbf{Notation:}] $\oplus$ denotes morphological dilation, which expands a binary mask according to the structuring element $B_r$. $\odot$ denotes element-wise multiplication. Masks are broadcast over RGB channels.

        \State Initialize $\mathbf{P}_i \leftarrow \mathbf{0}^{W \times H}$ for all $\mathbf{X}_i \in \mathcal{D}$ \label{alg:init_p}
        \State Initialize $\mathbf{X}_{i,0}\leftarrow\mathbf{X}_i$ and $\mathcal{D}_0\leftarrow\{(\mathbf{X}_{i,0},c_i)\}_{i=1}^{|\mathcal{D}|}$

        \For{$t = 0$ \textbf{to} $N-1$}
            \State \textbf{1. Train:} Train classifier $\texttt{Clf}_t$ on current dataset $\mathcal{D}_t$ \label{alg:train}

            \State \textbf{2. Extract:} \label{alg:extract_start}
            \For{each positive image $\mathbf{X}_{i,t} \in \mathcal{D}^+_t$}
                \State $\mathbf{G}_{i,t} \leftarrow \text{Grad-CAM}(\texttt{Clf}_t,\mathbf{X}_{i,t}; c=1)$ \Comment{Positive-class heatmap}

                \If{$t = 0$} \label{alg:init_start}
                    \State $\mathbf{A}_{i,0} \leftarrow \mathbf{G}_{i,0}$ \Comment{Init. accumulated heatmap}
                    \State $\mathbf{M}_{i,0} \leftarrow \mathbf{1}\{\mathbf{A}_{i,0} \ge \tau\}$ \Comment{Threshold}
                    \State $\mathbf{D}_{i,0} \leftarrow \mathbf{M}_{i,0}$ \Comment{All initial pixels are new} \label{alg:init_end}
                \Else
                    \State $\mathbf{A}_{i,t} \leftarrow \operatorname{clip}_{[0,1]}\left(\mathbf{A}_{i,t-1} + \mathbf{G}_{i,t}\right)$ \Comment{Accumulate} \label{alg:refine_start}

                    \If{$t < t_{\mathrm{core}}$} \Comment{Before CORE activation}
                        \State $\mathbf{M}_{i,t} \leftarrow \mathbf{1}\{\mathbf{A}_{i,t} \ge \tau\}$ \Comment{Unconstrained}
                    \Else
                        \State $r \leftarrow \max(1, \operatorname{round}(\kappa \cdot \sqrt{|\mathbf{M}_{i,t-1}|}))$ \Comment{Radius from mask area}
                        \State $B_r \leftarrow \text{Square element of side } 2r+1$
                        \State $\mathbf{R}_{i,t} \leftarrow \mathbf{M}_{i,t-1} \oplus B_r$ \Comment{CORE envelope}
                        \State $\mathbf{M}_{i,t} \leftarrow \mathbf{1}\{\mathbf{A}_{i,t} \ge \tau\} \cap \mathbf{R}_{i,t}$ \Comment{Constrain} \label{alg:core_logic}
                    \EndIf
                    \State $\mathbf{D}_{i,t} \leftarrow \mathbf{M}_{i,t} \setminus \mathbf{M}_{i,t-1}$ \Comment{New pixels} \label{alg:refine_end}
                \EndIf

                \State $\mathbf{P}_i[\mathbf{D}_{i,t}] \leftarrow N - t$ \Comment{Assign mining-order tier} \label{alg:assign_p}
            \EndFor \label{alg:extract_end}

            \If{$t<N-1$} \Comment{Prepare the next mining iteration}
                \State \textbf{3. Erase:} \label{alg:erase_start}
                \For{each image $\mathbf{X}_{i,t} \in \mathcal{D}_t$}
                    \If{$c_i=1$}
                        \State $\mathbf{M}_{i, \text{erase}} \leftarrow \mathbf{M}_{i,t}$ \Comment{Use image's own mask} \label{alg:positive_erase}
                    \Else
                        \State $\mathbf{M}_{i,\text{erase}} \leftarrow \text{Random select}(\{\mathbf{M}_{j,t} \mid  \mathbf{X}_j \in \mathcal{D}^+ \})$ \label{alg:negative_erase}
                    \EndIf
                    \State $\mathbf{X}_{i,t+1} \leftarrow (\boldsymbol{1} - \mathbf{M}_{i,\mathrm{erase}})\odot \mathbf{X}_{i,t} + \boldsymbol{\mu}\odot \mathbf{M}_{i,\mathrm{erase}}$ \label{alg:erase_formula}
                \EndFor \label{alg:erase_end}
                \State $\mathcal{D}_{t+1}\leftarrow\{(\mathbf{X}_{i,t+1},c_i)\}_{i=1}^{|\mathcal{D}|}$
            \EndIf
        \EndFor

        \State \Return $\mathcal{P}$
    \end{algorithmic}
\end{algorithm}

\textbf{Train} (Line~\ref{alg:train}). 
At each iteration $t$, we train the classifier $\texttt{Clf}_t$ on the current version of the dataset $\mathcal{D}_t$. 
Note that $\mathcal{D}_t$ changes at each iteration because the regions discovered in previous steps are erased from the images, forcing the model to focus on the remaining, less discriminative parts of the target patterns.

\textbf{Extract} (Lines~\ref{alg:extract_start}--\ref{alg:extract_end}). 
Let $\mathcal{D}^+$ be the set of positive images, i.e., those containing a polar low in our case.
For each positive image $\mathbf{X}_i \in \mathcal{D}_t^+$, we compute a Grad-CAM heatmap $\mathbf{G}_{i,t} \in [0,1]^{W\times H}$ that identifies the most discriminative regions.
For clarity, we distinguish two cases in the mask construction: initialization at $t=0$, which creates the initial seed mask, and subsequent expansion steps at $t\ge 1$, which update the accumulated heatmap with each newly computed saliency map.

\noindent\hspace*{1em}\emph{Initialization ($t=0$, Lines~\ref{alg:init_start}--\ref{alg:init_end}).}
For each image $\mathbf{X}_i$, we initialize the accumulated heatmap $\mathbf{A}_{i,0}=\mathbf{G}_{i,0}$ and threshold it to obtain the initial binary mask $\mathbf{M}_{i,0}=\mathbf{1}\{\mathbf{A}_{i,0}\ge\tau\}$. The threshold $\tau$ controls the precision--coverage trade-off of the mined regions: higher values retain only the strongest saliency responses, whereas lower values include larger but potentially noisier regions.
Pixels in $\mathbf{M}_{i,0}$ receive the highest tier $N$ because the first classifier typically highlights the most discriminative target cues (e.g., the cyclone eye), to which the mining-order scheme assigns the highest reliability:
\[
    \mathbf{P}_i[\mathbf{M}_{i,0}] \leftarrow N.
\]
Here, $\mathbf{P}_i$ is the pseudo-label associated with image $\mathbf{X}_i$.

\noindent\hspace*{1em}\emph{Refinement ($t\ge 1$, Lines~\ref{alg:refine_start}--\ref{alg:refine_end}).}
From the second iteration onward, we update the accumulated heatmap by adding the new saliency map and clipping the result to the normalized heatmap range $[0,1]$:
\[
    \mathbf{A}_{i,t}=\operatorname{clip}_{[0,1]}\left(\mathbf{A}_{i,t-1}+\mathbf{G}_{i,t}\right).
\]
We then derive the current mask $\mathbf{M}_{i,t}$ by applying the same threshold $\tau$ as before. 
When $t \ge t_{\mathrm{core}}$, we enable the \gls{core} constraint to limit each update to local expansion:
\[
    \mathbf{M}_{i,t} =
    \begin{cases}
        \mathbf{1}\{\mathbf{A}_{i,t}\ge\tau\},            & t<t_{\mathrm{core}},    \\[4pt]
        \mathbf{1}\{\mathbf{A}_{i,t}\ge\tau\}\ \cap\ \mathbf{R}_{i,t}, & t\ge t_{\mathrm{core}},
    \end{cases}
\]
where $\mathbf{R}_{i,t}$ is the \gls{core} allowed region, generated by morphologically dilating the previous mask $\mathbf{M}_{i,t-1}$ (see \Cref{fig:core_evolution}). 
The dilation radius is set as $r=\max(1,\operatorname{round}(\kappa\sqrt{|\mathbf{M}_{i,t-1}|}))$, where $|\mathbf{M}_{i,t-1}|$ is the area of the previous mask and $\kappa$ is an expansion factor controlling how far the mask is allowed to grow at each iteration.
After activation, each update is restricted to pixels near the previously mined support.
Because the accumulated heatmap is non-decreasing and dilation retains the previous mask, the mask sequence is also non-decreasing: $\mathbf{M}_{i,t-1}\subseteq \mathbf{M}_{i,t}$.
Thus, \gls{core} constrains new additions while retaining earlier pixels; an empty mask yields an empty envelope in subsequent iterations.
Newly discovered pixels $\mathbf{D}_{i,t} = \mathbf{M}_{i,t} \setminus \mathbf{M}_{i,t-1}$ are assigned a value in the pseudo-label based on the iteration $t$ at which they were discovered:
\[
    \mathbf{P}_i[\mathbf{D}_{i,t}] \leftarrow N-t.
\]
This assignment encodes the assumption that regions discovered later provide less reliable evidence of the target pattern and should therefore receive a lower tier value. 
Pixels that remain undiscovered after $N$ iterations form the pseudo-background class ($0$); in positive images, this class may also contain target pixels not recovered by the mining process.

\textbf{Erase} (Lines~\ref{alg:erase_start}--\ref{alg:erase_end}). Except after the final mining iteration, we prepare the next dataset $\mathcal{D}_{t+1}$ by removing the discovered features from the current training images~(Line~\ref{alg:erase_formula}). 
For each positive image $\mathbf{X}_i \in \mathcal{D}^+$ (i.e., containing a cyclone), we replace the pixels in $\mathbf{M}_{i,t}$ with the mean RGB vector $\boldsymbol{\mu}$ computed over the dataset $\mathcal{D}$~(Line~\ref{alg:positive_erase}).
The vector $\boldsymbol{\mu}$ is computed once from the unmodified training images.
We use mean replacement rather than zero-filling following the original adversarial erasing implementation~\cite{wei2018objectregionminingadversarial}. 
Here, $\mathbf{M}_{i,\mathrm{erase}} \in \{0,1\}^{W\times H}$ is broadcast across the three RGB channels, and $\boldsymbol{\mu}\in\mathbb{R}^3$ is broadcast over spatial positions.
To expose both classes to erased inputs, for every negative image $\mathbf{X}_i\in\mathcal{D}^{-}$ we sample a positive-image mask $\mathbf{M}_{j,t}$ and erase the corresponding region~(Line~\ref{alg:negative_erase}).
Donor masks are resampled at each iteration, so negative images see varied erased regions across the mining sequence.
This ensures that the presence of an erased region cannot be exploited by the next classifier $\texttt{Clf}_{t+1}$ to distinguish the classes. 
The corresponding pseudo-label for every negative image remains all background, so Stage~2 still sees a background-only mask for those samples.

After $N$ iterations, the output is a set of mining-order pseudo-labels $\mathcal{P} = \{\mathbf{P}_i\}$ in which higher values indicate earlier discoveries and receive higher assigned reliability.
Unlike a flat binary mask, this representation preserves the mining order, which Stage~2 uses in two ways: as the reliability signal by which it modulates supervision, and as the label space it predicts over, so that the ordering survives into the segmentation output rather than being collapsed at training time.

\subsection{Stage 2: Segmentation Network Training}

Stage~2 trains a shared image-to-mask model on the original images $\mathbf{X}_i$ paired with the Stage~1 pseudo-labels $\mathbf{P}_i$; the erased images are used only during Stage~1.
Unlike the instance-specific attribution maps, the trained segmentation model can be applied directly to unseen images and can learn spatial patterns shared across the training set rather than reproducing every attribution artifact.
This provides a mechanism for reducing noise and fragmentation in the Stage~1 pseudo-labels.

Although the pseudo-label tiers are ordered by mining iteration, the ``distance'' between them is not metric (i.e., class 3 is not ``three times'' as reliable as class 1). 
Therefore, we formulate the training task as a pixel-wise classification problem using a cross-entropy loss rather than a regression-based approach. Treating them as categorical classes allows us to assign independent reliability weights to each tier, providing better control over the supervision signal.

\subsubsection{Dynamic Bootstrapping (DB)}
To modulate the influence of the pseudo-label tiers, we employ a \gls{db} strategy~\cite{reed2015training_noisy_labels}. 
Early in training, the cross-entropy term uses the pseudo-labels directly. 
As training progresses, it attenuates the supervision from less reliable tiers, reducing the tendency to overfit their errors.

Formally, let $\mathcal{S}=\{0,\dots,N\}$ be the set of mining-order tiers in the pseudo-labels. 
For a given pixel, let $s\in \mathcal{S}$ be its assigned tier, encoded as a one-hot vector $\mathbf{y} \in \{0,1\}^{N+1}$:
\begin{equation}
    y_k =
    \begin{cases}
        1, & \text{if } k = s, \\
        0, & \text{otherwise.}
    \end{cases}
\end{equation}

Let $\mathbf{p}$ be the softmax distribution predicted for that pixel, so that $\mathbf{p}\in\Delta^N=\{\mathbf{q}\in[0,1]^{N+1}:\sum_{k=0}^{N}q_k=1\}$, and let $\rho_s \in (0, 1]$ denote the reliability assigned to tier $s$.
The foreground reliabilities follow the mining order established in Stage~1, with lower values assigned to regions discovered in later iterations, while the background tier ($s=0$) is assigned high reliability.
Attenuation should begin only after the network has learned an initial representation from the pseudo-labels; otherwise, early and poorly calibrated predictions could prematurely weaken their supervision.
We therefore use the clipped linear schedule as a function of the current training epoch $e$, with $e_{\mathrm{end}}>e_{\mathrm{start}}$,
\begin{equation}
    \gamma(e)
    =
    \operatorname{clip}_{[0,1]}
    \left(
        \frac{e-e_{\mathrm{start}}}
             {e_{\mathrm{end}}-e_{\mathrm{start}}}
    \right),
\end{equation}
which is $0$ up to epoch $e_{\mathrm{start}}$, increases linearly during the interval, and remains $1$ from epoch $e_{\mathrm{end}}$ onward.
This progressively activates the tier-dependent attenuation in a manner similar to curriculum learning~\cite{bengio2009curriculum}.
Combining the training schedule with the tier reliability gives the mixing coefficient

\begin{equation}
    \lambda_s(e) = \gamma(e)(1-\rho_s),
\end{equation}
and the soft target $\tilde{\mathbf{y}}$ used in the cross-entropy loss is defined as the convex mixture
\begin{equation}
    \tilde{\mathbf{y}} =
    \bigl(1-\lambda_s(e)\bigr)\mathbf{y}
    + \lambda_s(e)\operatorname{sg}(\mathbf{p}),
\end{equation}
where $\mathbf{p}$ is the current prediction from the same forward pass and $\operatorname{sg}(\cdot)$ denotes the stop-gradient operator: its argument is used in the forward pass but treated as constant during backpropagation.
If $\mathbf{z}$ denotes the logits for the pixel, the resulting cross-entropy gradient is
\begin{equation}
    \label{eq:db_gradient_scaling}
    \frac{\partial \mathcal{L}_{\mathrm{CE}}}{\partial \mathbf{z}}
    =
    \mathbf{p}-\tilde{\mathbf{y}}
    =
    \bigl(1-\lambda_s(e)\bigr)(\mathbf{p}-\mathbf{y}).
\end{equation}
Because the prediction in the target is detached, the per-pixel cross-entropy gradient is exactly the hard-label gradient rescaled by $1-\lambda_s(e)$: at each pixel the soft target changes the gradient's magnitude, not its direction. Across an image, this acts as a tier-dependent reweighting of the per-pixel hard-label gradients rather than a single global rescaling.
This equivalence is specific to cross-entropy; the Dice component remains anchored to the hard pseudo-labels.

The high reliability of the background tier keeps $\lambda_0(e)$ small and therefore preserves a strong correction when the model assigns foreground probability to a pixel labeled as background.
This suppresses false-positive leakage, while $\rho_0$ controls the balance between precision and foreground coverage.

\subsubsection{Loss Function}
We minimize a composite loss function that combines Soft \gls{ce} and Dice~\cite{milletari2016vnet_dice} losses:

\begin{equation}
    \mathcal{L} = \alpha\, \mathcal{L}_{\text{CE}}(\mathbf{p}, \tilde{\mathbf{y}}) + (1-\alpha)\, \mathcal{L}_{\text{Dice}}(\mathbf{p}, \mathbf{y}),
\end{equation}

where $\alpha \in [0, 1]$ balances the relative contribution of the two loss terms.
For a training image with pixel set $\Omega$, the Soft Cross-Entropy loss is defined as:

\begin{equation}
    \mathcal{L}_{\text{CE}}(\mathbf{p}, \tilde{\mathbf{y}}) = - \frac{1}{|\Omega|} \sum_{u \in \Omega} \sum_{k=0}^{N} \tilde{y}_{u,k} \log p_{u,k}.
\end{equation}

The Dice loss is computed against the hard one-hot pseudo-labels. 
We first define
\begin{equation}
    \mathcal{K}
    =
    \left\{
        k\in\{0,\dots,N\}
        \,\middle|\,
        \sum_{u\in\Omega}y_{u,k}>0
    \right\},
\end{equation}
the set of tiers present in that image.
For each $k\in\mathcal{K}$, we compute

\begin{equation}
    \ell^{(k)}_{\text{Dice}}
    = 1 -
    \frac{2 \sum_{u \in \Omega} p_{u,k}\, y_{u,k}}
    {\sum_{u \in \Omega} p_{u,k} + \sum_{u \in \Omega} y_{u,k}}.
\end{equation}

We omit absent tiers and average the remaining class losses using reliability weights normalized over $\mathcal{K}$:

\begin{equation}
    \mathcal{L}_{\text{Dice}}(\mathbf{p}, \mathbf{y})
    =
    \sum_{k\in\mathcal{K}} w_k \ell^{(k)}_{\text{Dice}},
    \qquad
    w_k = \frac{\rho_k}{\sum_{j\in\mathcal{K}}\rho_j}.
\end{equation}
Both image-level loss terms are computed per image, and the resulting composite losses are averaged over the mini-batch.
When the background tier is present, its reliability gives the background-overlap term a weight comparable to the earliest-mined foreground tiers, so foreground probability assigned to background-labeled pixels still incurs a region-level penalty.

The soft \gls{ce} term is the pixel-wise component through which tier-dependent attenuation is applied.
Cross-entropy is computed independently at each pixel, so fragmented activations have limited effect on the objective when most pixels are classified correctly.
The Dice term directly optimizes class-level overlap, which promotes spatially coherent masks and counterbalances the dominance of background pixels.

We deliberately compute Dice against the hard pseudo-labels so that the overlap term remains anchored to the regions mined in Stage~1, while the class weights $w_k$ reduce the influence of less reliable tiers.
A simple two-class example illustrates why directly replacing the hard target with a soft target in the reliability-weighted Dice term can be undesirable.

Consider one pixel with a fixed soft target $\mathbf{q}=(0.6,0.4)$, a prediction $\mathbf{p}=(x,1-x)$, and class weights $(w_1,w_2)=(0.8,0.2)$.
Both $\mathbf{q}$ and $\mathbf{p}$ lie on the probability simplex.
The corresponding reliability-weighted soft Dice loss is
\begin{equation}
    \begin{aligned}
    \mathcal{L}_{\mathrm{Dice}}(x,\mathbf{q})
    ={}&
    0.8\left(1-\frac{1.2x}{x+0.6}\right) \\
    &+
    0.2\left(1-\frac{0.8(1-x)}{1.4-x}\right).
    \end{aligned}
\end{equation}
Its derivative is
\begin{equation}
    \frac{\mathrm{d}\mathcal{L}_{\mathrm{Dice}}}{\mathrm{d}x}
    =
    -\frac{0.576}{(x+0.6)^2}
    +\frac{0.064}{(1.4-x)^2}.
\end{equation}
Setting this derivative to zero gives $(x+0.6)/(1.4-x)=3$ and therefore $x=0.9$.
The derivative is negative for $x<0.9$ and positive for $x>0.9$, so the loss is minimized at $\mathbf{p}=(0.9,0.1)$ rather than at the soft target $\mathbf{q}=(0.6,0.4)$.
Thus, in the presence of unequal reliability weights, a soft Dice target can shift the optimum toward the more heavily weighted class instead of encouraging the prediction to reproduce the target distribution.
Consequently, tier-dependent attenuation acts most directly through \gls{ce}; the hard-target Dice term continues to provide region-level supervision even when the corresponding \gls{ce} gradient is attenuated.

\subsubsection{Effect of Dynamic Bootstrapping}
\begin{figure*}[!t]
    \centering
    \includegraphics[width=\textwidth, keepaspectratio]{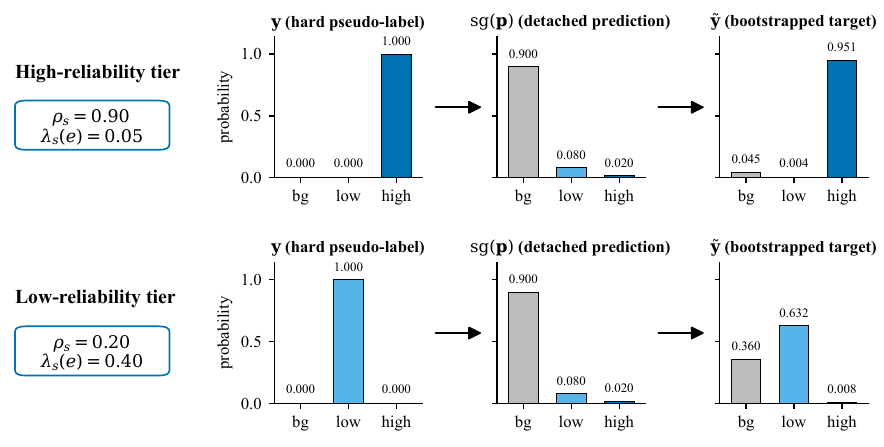}
    \caption{Effect of dynamic bootstrapping under the same background-biased model prediction $\mathbf{p}=[0.9,0.08,0.02]$ at $\gamma(e)=0.5$. The bootstrapped target mixes the hard pseudo-label $\mathbf{y}$ with the detached prediction $\operatorname{sg}(\mathbf{p})$. For the high-reliability case, $\rho_s=0.9$ and $\lambda_s(e)=0.05$, corresponding to CE gradient $\times\,0.95$ ($5\%$ attenuation); for the low-reliability case, $\rho_s=0.2$ and $\lambda_s(e)=0.40$, corresponding to CE gradient $\times\,0.60$ ($40\%$ attenuation).}
    \label{fig:dynamic_bootstrapping_effect}
\end{figure*}

\Cref{fig:dynamic_bootstrapping_effect} illustrates the tier-dependent effect at an epoch where $\gamma(e)=0.5$.
For clarity, it considers three classes (background plus low- and high-reliability foreground) and uses the same model prediction $\mathbf{p}=[0.9,0.08,0.02]$ in both cases, which isolates the effect of the tier reliability.

\paragraph{High-reliability tier}
Consider a pixel labeled as high-reliability foreground, for which $\rho_s=0.9$, $\mathbf{y}=[0,0,1]$, and $\lambda_s(e)=0.05$.
Although the model prediction contradicts the pseudo-label, the bootstrapped target remains close to the hard label:
\begin{equation}
    \begin{aligned}
    \tilde{\mathbf{y}}
    &=0.95[0,0,1]+0.05[0.9,0.08,0.02] \\
    &=[0.045,0.004,0.951].
    \end{aligned}
\end{equation}
The hard-target cross-entropy is
\begin{equation}
    -\log(0.02) \approx 3.91,
\end{equation}
whereas the bootstrapped target gives
\begin{equation}
    \begin{aligned}
    -\bigl(&0.045\log(0.9)+0.004\log(0.08)\\
           &+0.951\log(0.02)\bigr)
    \approx 3.74.
    \end{aligned}
\end{equation}
For this prediction, the forward loss value is therefore reduced by approximately $4.5\%$.
More importantly for optimization, \Cref{eq:db_gradient_scaling} shows that the cross-entropy gradient is multiplied by $1-\lambda_s(e)=0.95$, corresponding to a $5\%$ attenuation.

\paragraph{Low-reliability tier}
Now consider a pixel labeled as low-reliability foreground, for which $\rho_s=0.2$, $\mathbf{y}=[0,1,0]$, and $\lambda_s(e)=0.40$.
For the same model prediction, the target becomes
\begin{equation}
    \begin{aligned}
    \tilde{\mathbf{y}}
    &=0.60[0,1,0]+0.40[0.9,0.08,0.02] \\
    &=[0.360,0.632,0.008].
    \end{aligned}
\end{equation}
The hard-target cross-entropy is
\begin{equation}
    -\log(0.08) \approx 2.53,
\end{equation}
whereas the bootstrapped target gives
\begin{equation}
    \begin{aligned}
    -\bigl(&0.360\log(0.9)+0.632\log(0.08)\\
           &+0.008\log(0.02)\bigr)
    \approx 1.67.
    \end{aligned}
\end{equation}
Here, the forward loss value is reduced by approximately $34\%$, while the exact gradient multiplier is $1-\lambda_s(e)=0.60$, corresponding to a $40\%$ attenuation.

The reductions in the forward loss values depend on the particular prediction $\mathbf{p}$ and should not be interpreted as the amount by which the optimizer's update is reduced.
The gradient attenuation is instead determined exactly by $1-\lambda_s(e)$.
Thus, \gls{db} retains most of the cross-entropy supervision from reliable tiers while relaxing it more strongly for tiers that are more likely to contain mining errors.

\section{Experiments}\label{sec:exp}

In this section, we evaluate the proposed \gls{crest} pipeline.
We first outline our evaluation strategy, which addresses the lack of pixel-level ground truth for the primary \gls{sar} images.
Then, we describe how we perform the comparative analysis against baselines and discuss the results.
We conclude by presenting a progressive ablation study of the proposed modules.
We defer to Appendix~\ref{app:dataset} the description of the datasets, including those unrelated to our case study, that are used for quantitative benchmarking, while in Appendix~\ref{app:exp_settings} we detail the experimental setup.

\subsection{Evaluation Strategy}

Unless otherwise stated, all results reported in this section are computed on the masks predicted by the trained Stage~2 segmentation network, not on the raw pseudo-labels produced by Stage~1. 
The pseudo-labels are an intermediate supervision signal: they guide optimization of the segmentation model but remain noisy and image-specific by construction. 
Evaluating the trained segmentation network is therefore the relevant endpoint because it is the deployable component of the pipeline.

Since the \gls{sar} dataset has no pixel-level ground truth, we assess the predicted masks qualitatively, comparing them against those produced by the baselines on the visible cyclone structure across test scenes.
The gallery in Appendix~\ref{app:pl_results} and a consistency check under geometric transformations in Appendix~\ref{app:robustness} show additional results.
In addition, we overlay an acquisition with auxiliary \gls{slp} and 10\,m wind fields from \gls{era5}~\cite{hersbach2020era5} to place the mined tiers in physical context.

To complement this qualitative analysis with a quantitative evaluation, we also apply the proposed pipeline to two public datasets that provide pixel-level ground truth: \textbf{\gls{bus}}~\cite{BUS_citation}, a dataset of breast ultrasound images for tumor segmentation, and \textbf{\gls{voc}}~\cite{Everingham_2025_VOC2012}, a standard computer vision benchmark. 
In \gls{voc}, we select the \texttt{person} class to form a binary semantic segmentation task; an image may contain multiple people as well as objects from other classes.
We select these two datasets because their targets are compatible with the same spatial-connectedness prior of \gls{core}, namely that each object forms a coherent region whose peripheral parts can be reached by expanding locally from the most discriminative seed.
Each tumor is a compact, connected mass, and person instances are likewise spatially connected.
Having multiple instances within the same image does not conflict with this prior: thresholding the accumulated heatmap yields one seed per instance, and because the \gls{core} envelope is a dilation of the entire current mask, each seed expands within its own local neighborhood.
The last rows of \Cref{fig:bus_qual,fig:voc_qual} show this for two detached lesions and for multiple people.
These two datasets therefore serve two purposes.
First, they make the results easier to judge, since the boundaries of tumors and persons are less ambiguous than those of a cyclone.
Second, they carry the pixel-level ground truth that the \gls{sar} dataset does not, which allows \gls{crest} and standard \aer{} to be compared numerically under identical settings (\Cref{sec:comparison}).
This comparison is not a claim of benchmark performance: all methods run under a deliberately minimal shared protocol with no post-processing, so the absolute values reported here sit well below what specialized state-of-the-art methods achieve on either dataset.
What these tables are for is the difference between \gls{crest} and standard \aer{}, measured where the predicted masks can actually be scored, which is what the polar low data do not allow.


\subsection{Method Comparison}\label{sec:comparison}

We benchmark our proposed pipeline against two baselines to assess the contribution of each component.

\paragraph{\gls{gradcam}} This baseline represents the simplest of the weakly supervised approaches. 
We generate a single \gls{gradcam} attribution map from the classifier and threshold it (top 30\%) to create a binary pseudo-label, which is used to supervise the segmentation network. This method relies entirely on the initial discriminative regions found by the classifier without performing any iterative discovery of new regions.

\paragraph{Standard \aer{}} This is the vanilla adversarial erasing pipeline. It employs an iterative mining strategy: after training the classifier, the most discriminative regions are erased, and the classifier is retrained on the modified images to discover new, complementary features. Once the mining procedure is complete, all regions discovered across iterations are merged into a single binary mask, which serves as the pseudo-label for the segmentation network.

\paragraph{\gls{crest} (ours)} Our full method builds upon standard \aer{} by integrating two main components: the \core{} module constrains new region growth during mining, while the \gls{db} loss uses the preserved mining order as a proxy for reliability and attenuates cross-entropy supervision from later-mined regions.

\vspace{0.5em}
\noindent To ensure a controlled comparison, all methods use the same Xception backbone, \gls{gradcam} implementation, SegFormer architecture, and training schedule.
The standalone \gls{gradcam} reference uses a top-30\% threshold, whereas the iterative methods use the dataset-specific absolute thresholds reported in Appendix~\ref{app:exp_settings}.
Consequently, the standalone \gls{gradcam} baseline is not identical to iteration~0 of standard \gls{aer}; the latter is reported in Appendix~\ref{app:detailed_results} as the single-pass reference under the iterative-method protocol.

The original \gls{aer} formulation builds both its classification and segmentation networks on DeepLab-CRF-LargeFOV, applies a \gls{crf} to post-process the segmentation output at test time, and in an optional further round uses \gls{crf}-refined training-set predictions as supervision~\cite{wei2018objectregionminingadversarial, chen2017deeplab, krahenbuhl2011crf}.
However, we decided not to apply \gls{crf} because it relies on strong intensity edges, which the diffuse cyclone boundaries in \gls{sar} imagery do not provide.
We, therefore, omitted this stage in the polar low setting and on \gls{bus} and \gls{voc} as well, holding all methods to identical conditions.

\subsection{Results}

\begin{figure}[t]
    \centering
    \setlength{\tabcolsep}{1.5pt} 
    \resizebox{\columnwidth}{!}{%
        \begin{tabular}{cccc}
            {\scriptsize\textbf{Input}}                                                   &
            {\scriptsize\textbf{\gls{gradcam}}}                                           &
            {\scriptsize\textbf{Standard \aer{}}}                                       &
            {\scriptsize\textbf{\gls{crest}}}                                                \\
            \includegraphics[width=.24\linewidth]{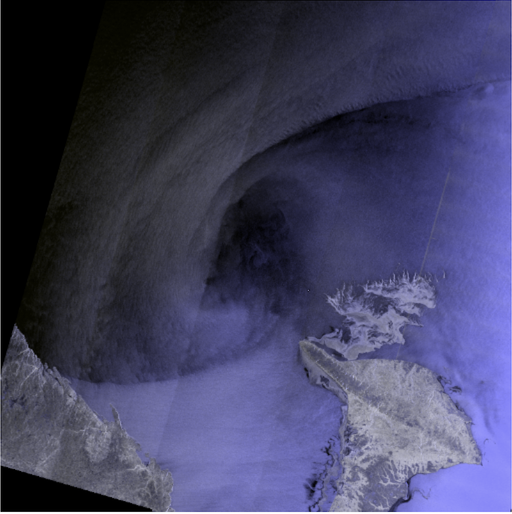}   &
            \includegraphics[width=.24\linewidth]{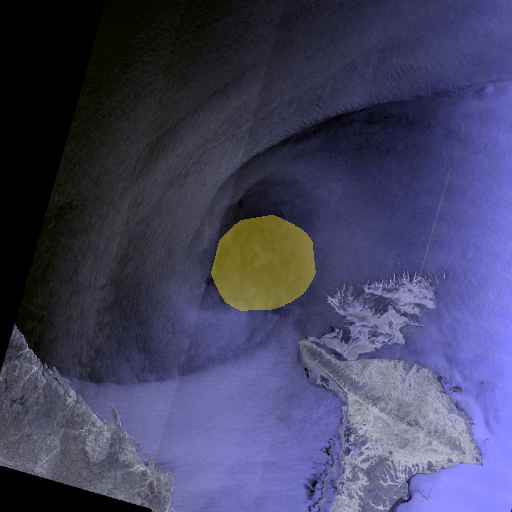} &
            \includegraphics[width=.24\linewidth]{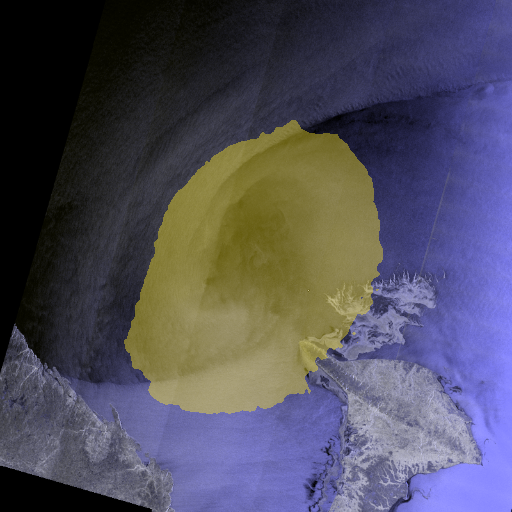}     &
            \includegraphics[width=.24\linewidth]{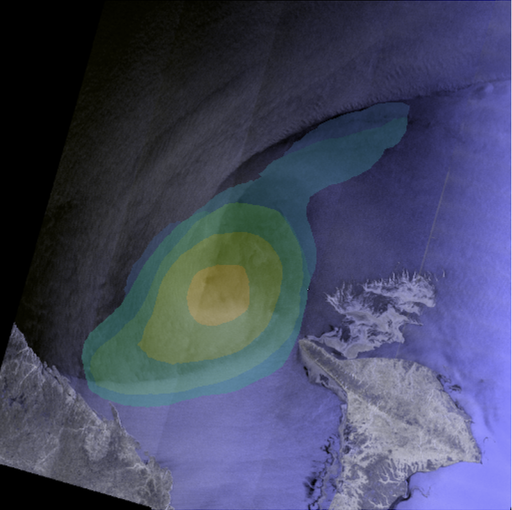} \\
            \includegraphics[width=.24\linewidth]{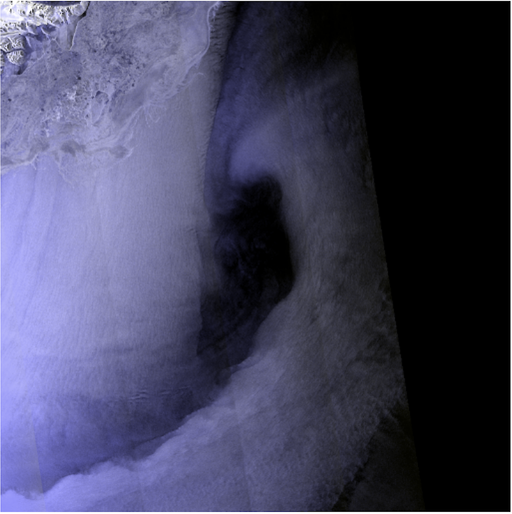}   &
            \includegraphics[width=.24\linewidth]{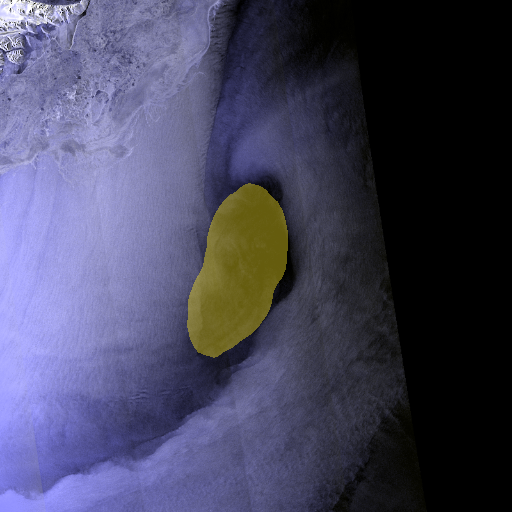} &
            \includegraphics[width=.24\linewidth]{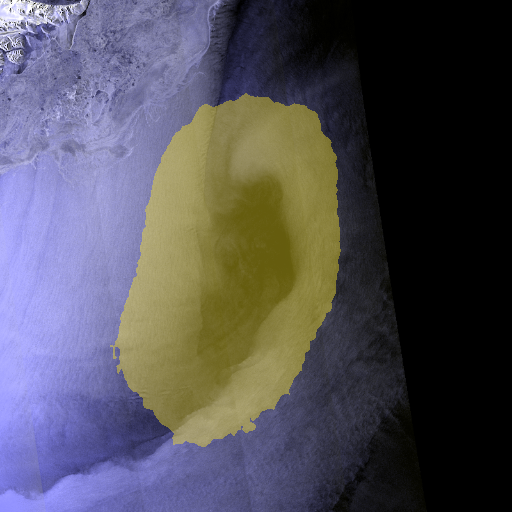}     &
            \includegraphics[width=.24\linewidth]{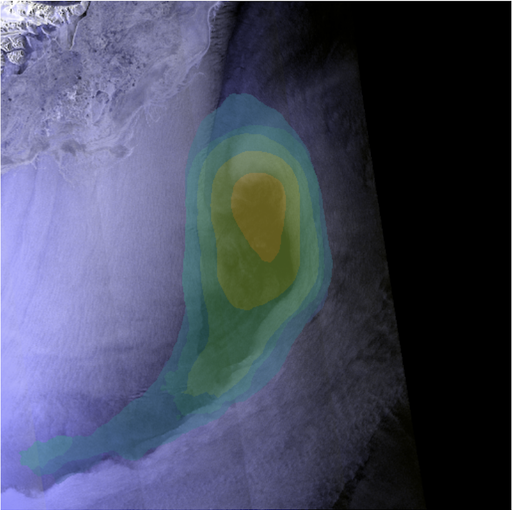} \\
            \includegraphics[width=.24\linewidth]{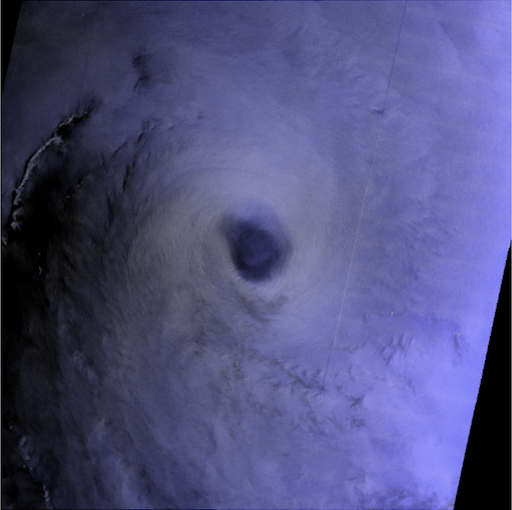}   &
            \includegraphics[width=.24\linewidth]{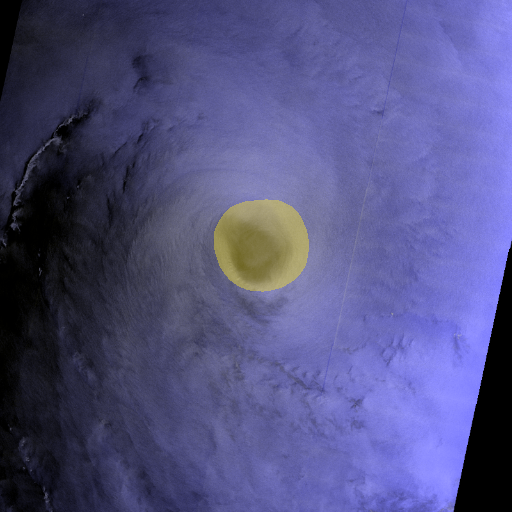} &
            \includegraphics[width=.24\linewidth]{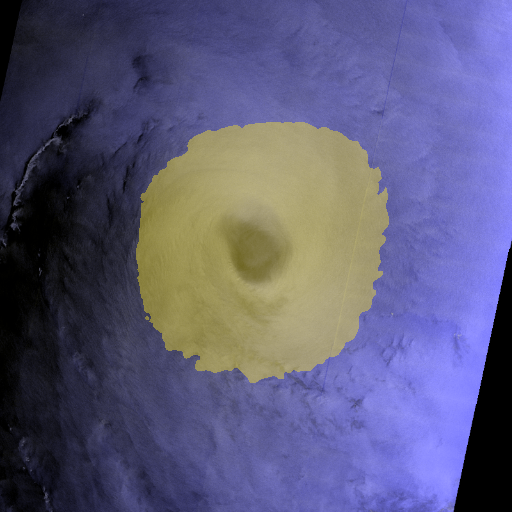}     &
            \includegraphics[width=.24\linewidth]{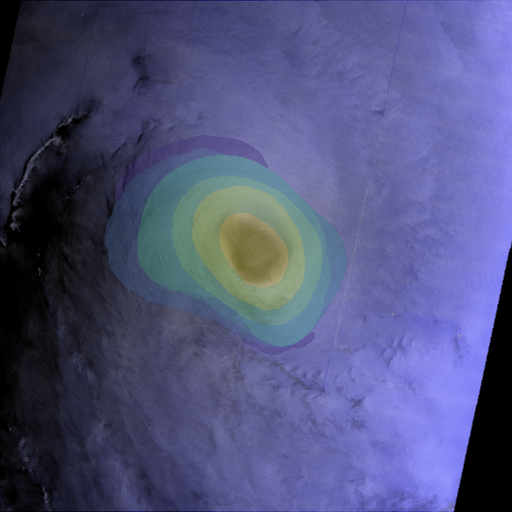} \\
            \includegraphics[width=.24\linewidth]{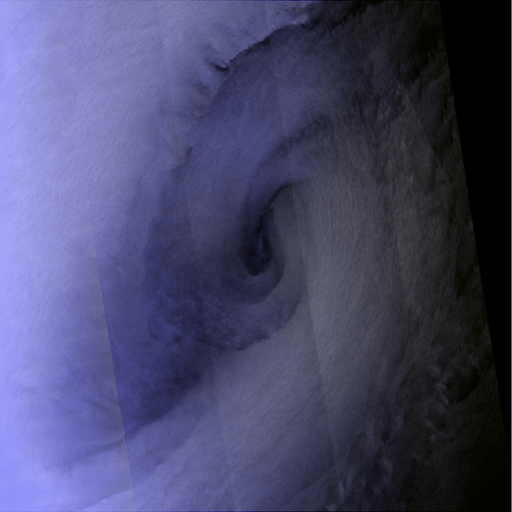}   &
            \includegraphics[width=.24\linewidth]{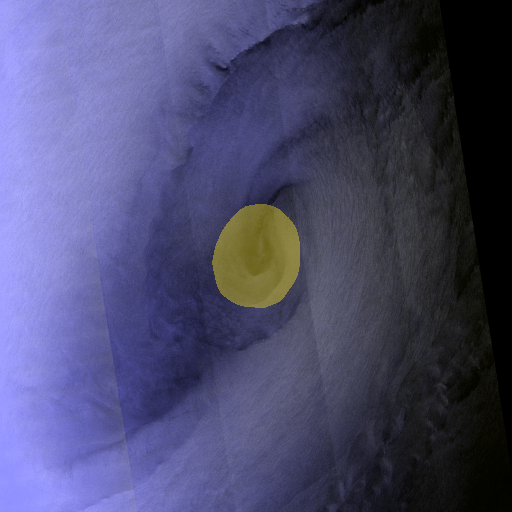} &
            \includegraphics[width=.24\linewidth]{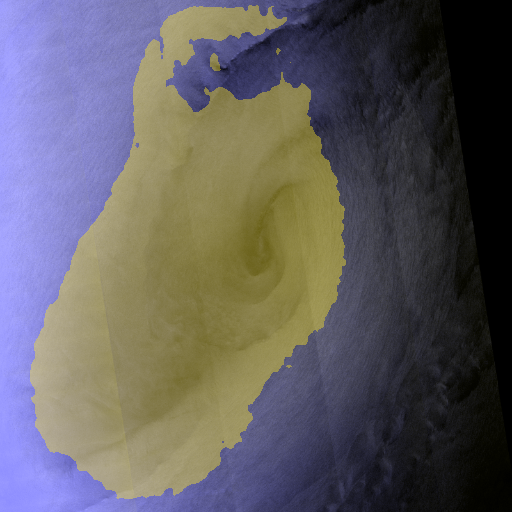}     &
            \includegraphics[width=.24\linewidth]{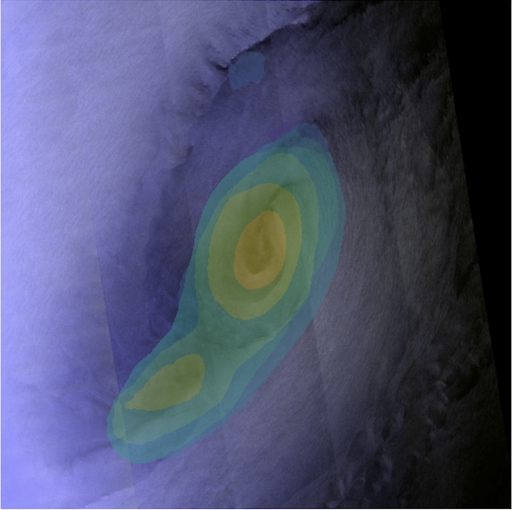} \\
        \end{tabular}}
    \caption{Visual comparison of segmentation masks for maritime mesocyclones in \gls{sar} images.
    First column: input images. 
    Second: segmentation network trained with single-pass \gls{gradcam} masks, which concentrate mainly around the cyclone eye in these examples. 
    Third: standard \aer{} extends the prediction to more peripheral regions, but can also incorporate background. 
    Fourth: \gls{crest} predictions, with colors encoding mining order.}
    \label{fig:pl_comparison}
\end{figure}

\begin{figure}[t]
    \centering
    \includegraphics[
        width=\columnwidth,
        keepaspectratio
    ]{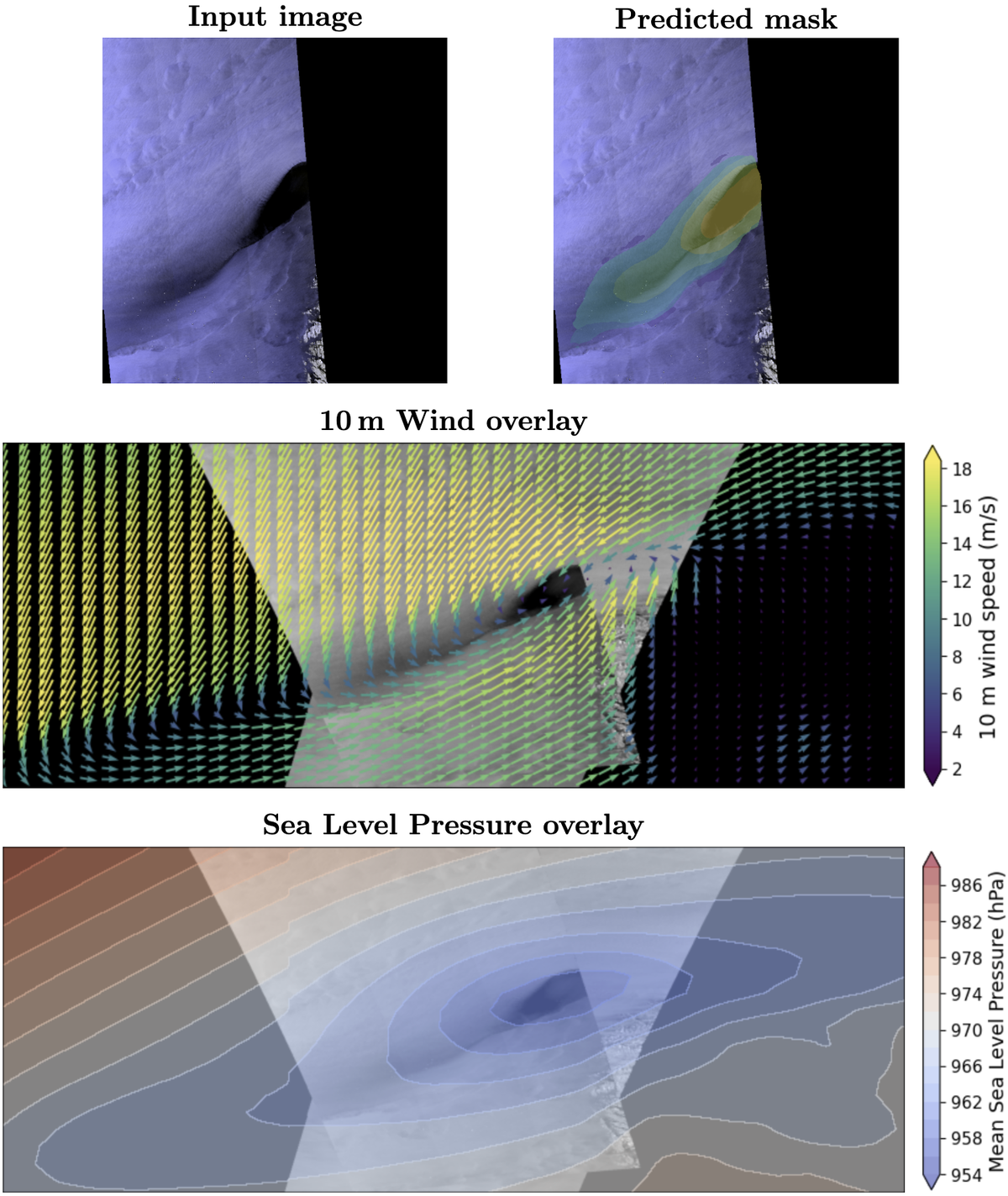}
    \caption{Illustrative case: a single \gls{sar} acquisition overlaid with auxiliary meteorological fields.} Top row: \gls{sar} and predicted mask. Middle: \gls{sar} with 10\,m wind. Bottom: \gls{sar} with \gls{slp}. The meteorological fields were manually aligned for visualization and are not used as model inputs, pixel-level supervision, or ground truth for segmentation metrics. Time and region: 2019-12-08 17:28~UTC; 59.58--64.66\degree N, 4.39\degree W--9.45\degree E.
    \label{fig:slp}
\end{figure}

\paragraph{\gls{sar} cyclones}

We begin with the primary application: segmentation of polar lows that constitute the dataset-defined positive class.
Since quantitative ground truth is unavailable, we rely on visual inspection across test scenes.
For the \gls{sar} results, the iterative methods retain mining iterations up to $t=4$, so \gls{crest} masks can show up to five foreground tiers, and fewer when the network does not predict all of them in a given image.

\Cref{fig:pl_comparison} illustrates the limitations of the \gls{gradcam} and \gls{aer} baselines.
The \gls{gradcam} approach (second column) focuses mainly on the cyclone eye in these examples and does not capture the full visible extent of the system.
Standard \gls{aer} (third column) mines larger regions but can also drift into the background and include unrelated wind-driven sea-surface patterns or other background features.
In these examples, \gls{crest} (fourth column) produces spatially coherent masks that follow spiral- and comma-like structures while avoiding the broad background expansion seen for standard \gls{aer}.
The same behavior holds over the gallery of Appendix~\ref{app:pl_results} and under the geometric transformations of Appendix~\ref{app:robustness}.
The tiers also carry information that the binary output of standard \gls{aer} discards: warm colors mark the regions mined first and cool colors those mined last, so a single prediction records both where the model places the cyclone and in what order it recovered its parts.

\Cref{fig:slp} places one prediction in meteorological context.
\gls{slp} and 10\,m wind are diagnostic variables for mesocyclones, since a mature system is associated with a compact sea-level pressure minimum and with strong, organized near-surface winds that wrap around the vortex and its frontal bands.
In this scene, the early-mined regions (yellow/orange) lie near the low-pressure center while later-mined tiers (purple/blue) extend into the surrounding circulation, so the mining order runs from the cyclone core outward.
These variables are not provided to the segmentation model as inputs or pixel-level supervision, and the overlay is not used to compute segmentation scores.

The \gls{era5} fields are available at hourly steps, whereas each \gls{sar} image is acquired at an arbitrary time within the hour, so the cyclone moves between the nearest reanalysis step and the acquisition, and the closest fields had to be aligned with the image by hand.

\begin{table*}[t]
    \centering
    \caption{Best test performance on \gls{bus} and \gls{voc}. For the iterative methods, we report the mining iteration that achieved the best Macro IoU; Tables~\ref{tab:bus_full_results} and~\ref{tab:voc_full_results} report all iterations. Best results per metric are highlighted in bold.}
    \label{tab:test_results}
    \begin{tabular*}{\textwidth}{@{\extracolsep{\fill}}lrrrrrrrrrrrr@{}}
        \toprule
        \multirow{3}{*}{Method} & \multicolumn{6}{c}{\makecell{\gls{bus}}} & \multicolumn{6}{c}{\makecell{\gls{voc}}} \\
        \cmidrule(lr){2-7} \cmidrule(lr){8-13}
        & \multicolumn{2}{c}{\makecell{Macro}} & \multicolumn{4}{c}{\makecell{Foreground}} & \multicolumn{2}{c}{\makecell{Macro}} & \multicolumn{4}{c}{\makecell{Foreground}} \\
        \cmidrule(lr){2-3} \cmidrule(lr){4-7} \cmidrule(lr){8-9} \cmidrule(lr){10-13}
        & IoU $\uparrow$ & Dice $\uparrow$ & IoU $\uparrow$ & Dice $\uparrow$ & Precision $\uparrow$ & Recall $\uparrow$ & IoU $\uparrow$ & Dice $\uparrow$ & IoU $\uparrow$ & Dice $\uparrow$ & Precision $\uparrow$ & Recall $\uparrow$ \\
        \midrule
        \gls{gradcam}            & 57.2 & 64.2 & 17.7 & 30.1 & \textbf{62.4} & 19.8  & 51.8 & 54.8 & 5.7 & 10.7 & 39.8 & 6.2 \\
        Standard \aer{}            & 60.2 & 68.6 & 24.3 & 39.1 & 45.0 & 34.6  & 65.4 & 74.4 & 33.5 & 50.2 & 38.8 & 71.3 \\
        \textbf{\gls{crest}}       & \textbf{64.8} & \textbf{73.9} & \textbf{32.7} & \textbf{49.3} & 58.7 & \textbf{42.5}  & \textbf{67.4} & \textbf{76.6} & \textbf{37.5} & \textbf{54.6} & \textbf{41.1} & \textbf{81.0} \\
        \bottomrule
    \end{tabular*}
\end{table*}

\begin{figure}[t]
    \centering
    \setlength{\tabcolsep}{1.5pt}
    \begin{tabular}{ccc}
        {\scriptsize\textbf{Input}}                                                 &
        {\scriptsize\textbf{Prediction}}                                            &
        {\scriptsize\textbf{Ground truth}}                                            \\
        \includegraphics[width=0.32\linewidth]{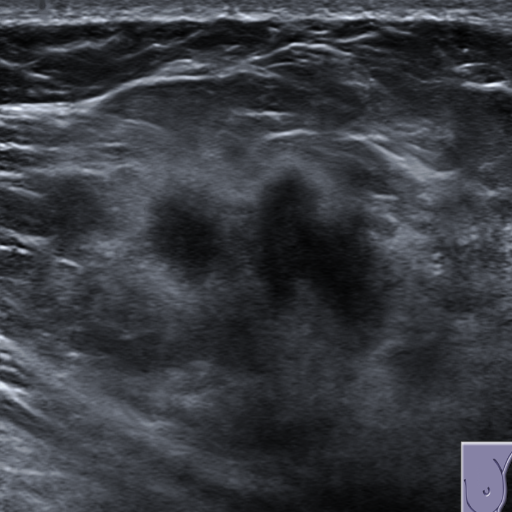}  &
        \includegraphics[width=0.32\linewidth]{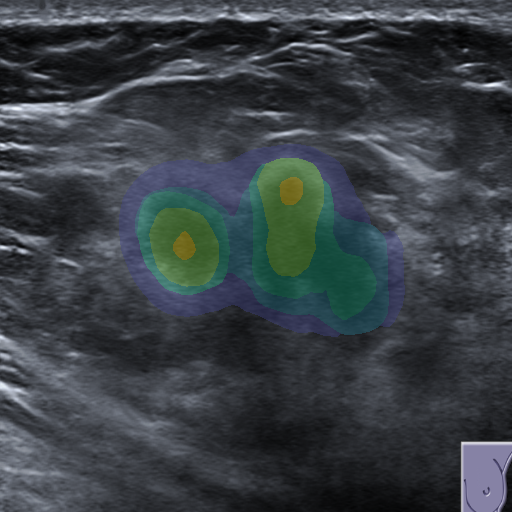} &
        \includegraphics[width=0.32\linewidth]{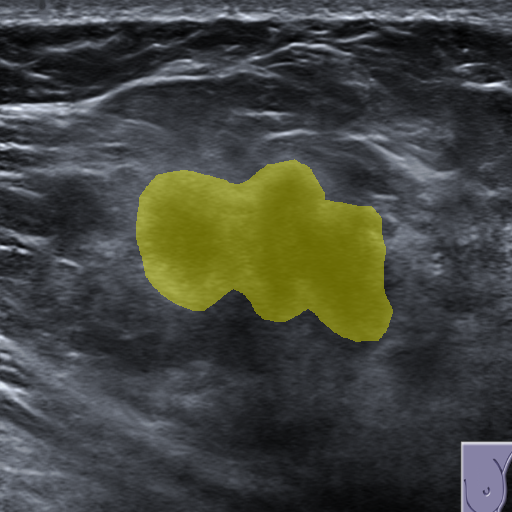}     \\
        \includegraphics[width=0.32\linewidth]{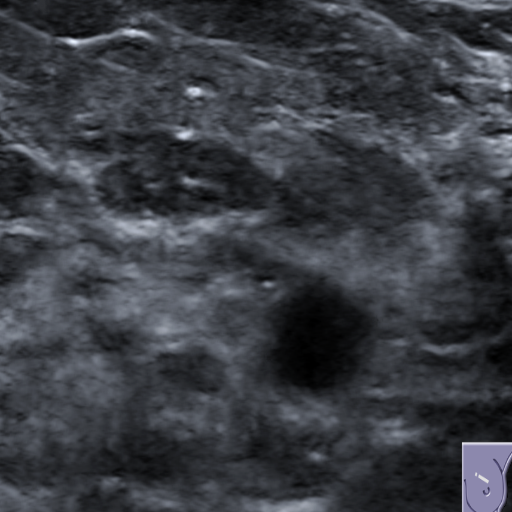}  &
        \includegraphics[width=0.32\linewidth]{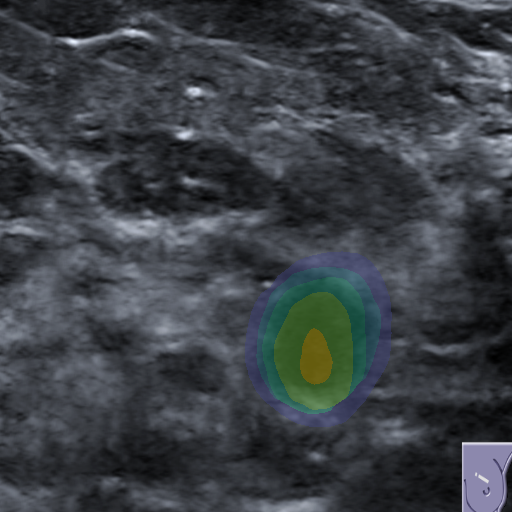} &
        \includegraphics[width=0.32\linewidth]{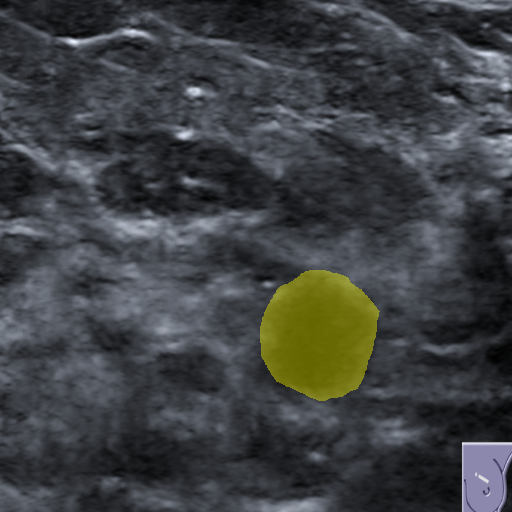}     \\
        \includegraphics[width=0.32\linewidth]{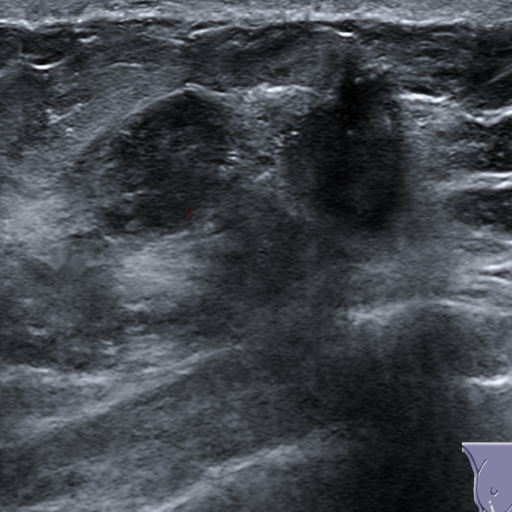}  &
        \includegraphics[width=0.32\linewidth]{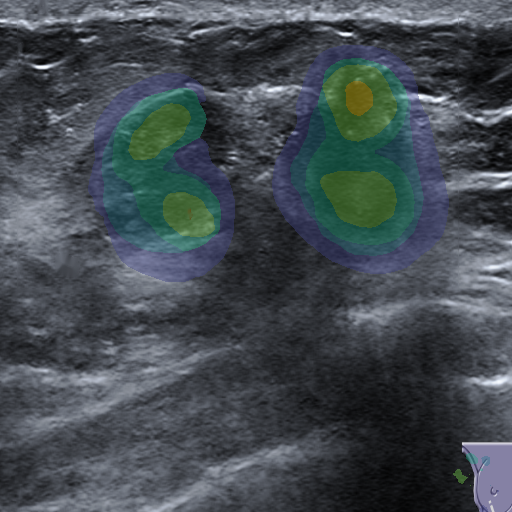} &
        \includegraphics[width=0.32\linewidth]{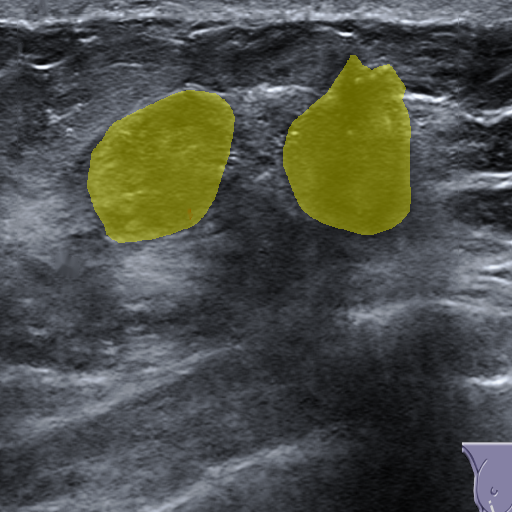}     \\
    \end{tabular}
    \caption{\gls{bus} qualitative examples. 
    The predicted multi-class masks preserve the mining order from \gls{aer}: warmer colors (yellow) indicate earlier-mined, more discriminative regions, while cooler colors (blue) indicate regions discovered later. 
    In the displayed examples, the earliest tiers fall on the lesion core and later tiers extend toward its margin, and the ground-truth masks are shown for visual comparison.}
    \label{fig:bus_qual}
\end{figure}

\begin{figure}[!htbp]
    \centering
    \setlength{\tabcolsep}{2pt} 
    \begin{tabular}{@{}cccc@{}}
        \scriptsize\textbf{Input}                                                                          & \scriptsize\textbf{Multi-class} & \scriptsize\textbf{Binarized} & \scriptsize\textbf{Ground truth} \\
        \includegraphics[width=.24\linewidth]{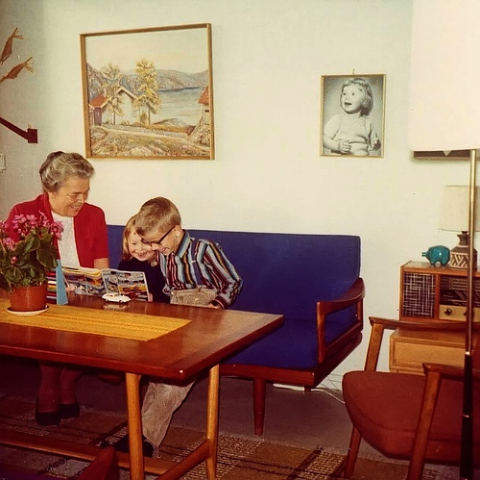}         &
        \includegraphics[width=.24\linewidth]{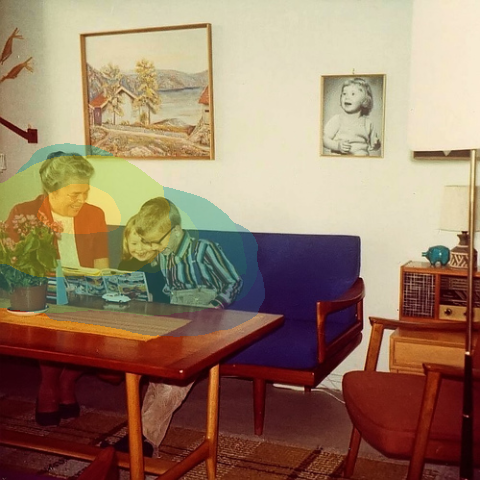} &
        \includegraphics[width=.24\linewidth]{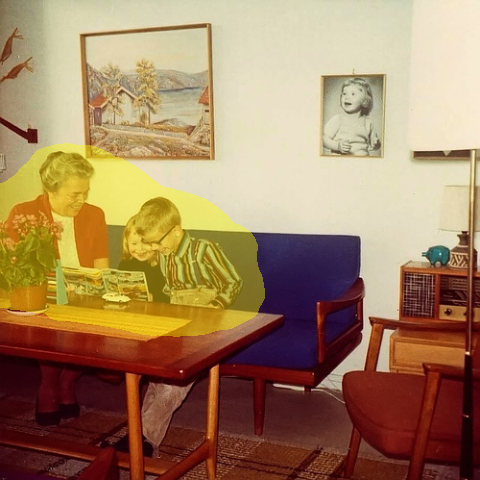}       &
        \includegraphics[width=.24\linewidth]{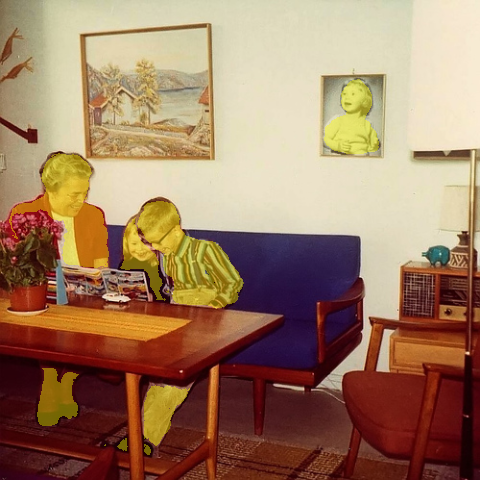}                                                                                                            \\
        \includegraphics[width=.24\linewidth]{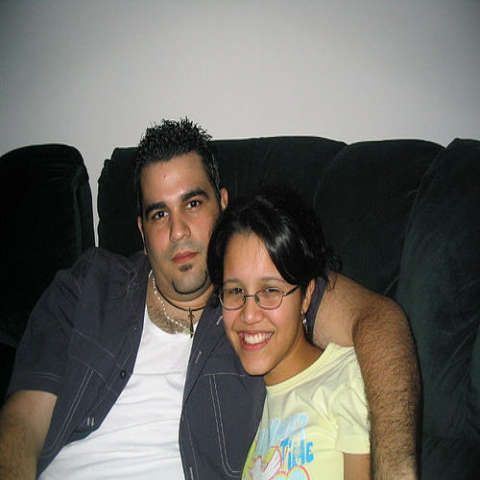}         &
        \includegraphics[width=.24\linewidth]{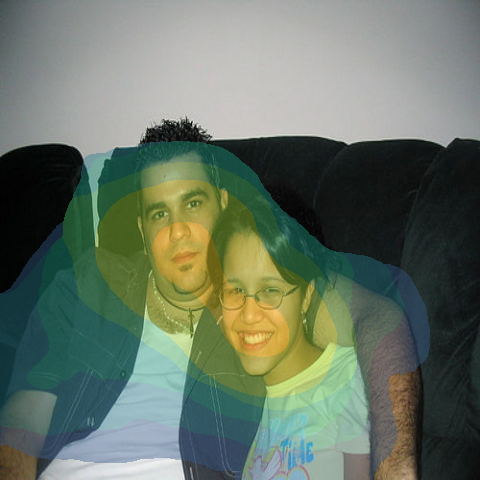} &
        \includegraphics[width=.24\linewidth]{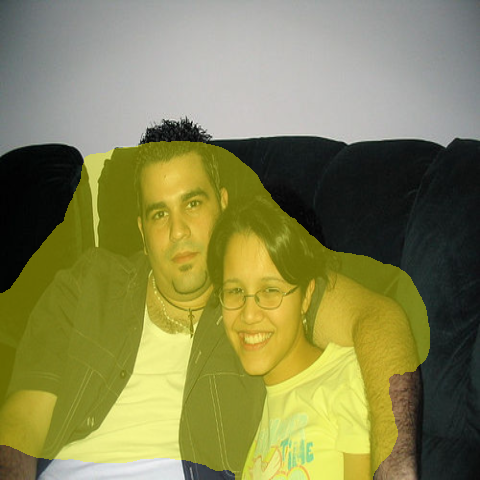}       &
        \includegraphics[width=.24\linewidth]{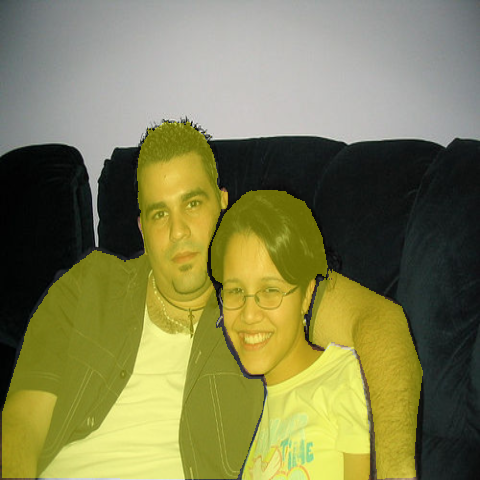}                                                                                                            \\
        \includegraphics[width=.24\linewidth]{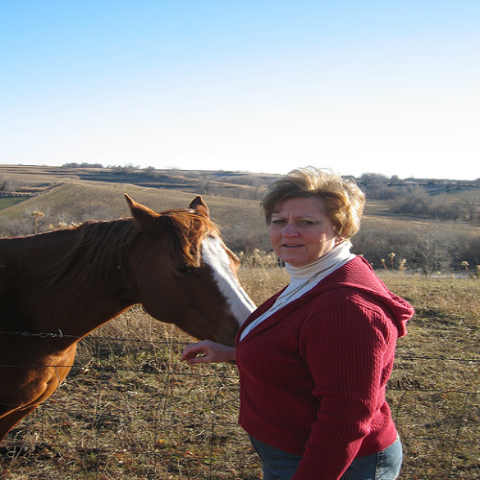}         &
        \includegraphics[width=.24\linewidth]{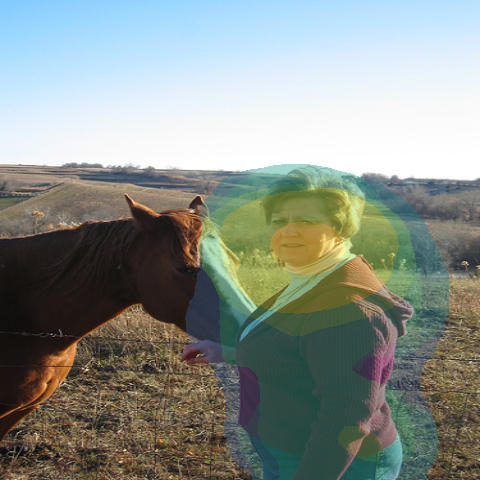} &
        \includegraphics[width=.24\linewidth]{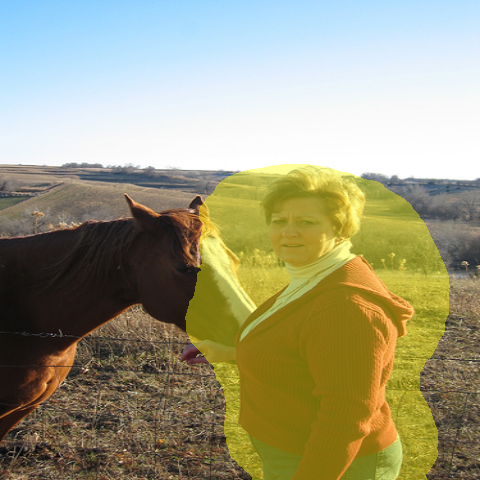}       &
        \includegraphics[width=.24\linewidth]{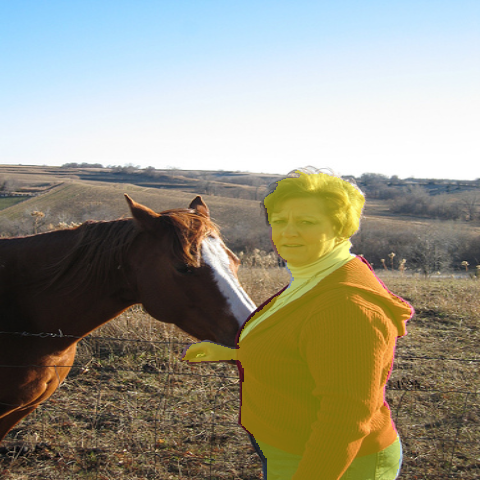}                                                                                                            \\
        \includegraphics[width=.24\linewidth]{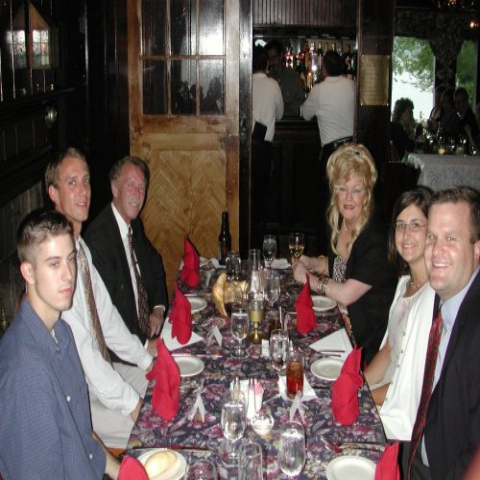}         &
        \includegraphics[width=.24\linewidth]{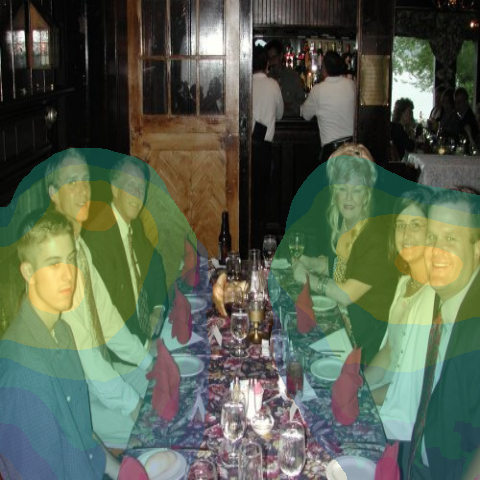} &
        \includegraphics[width=.24\linewidth]{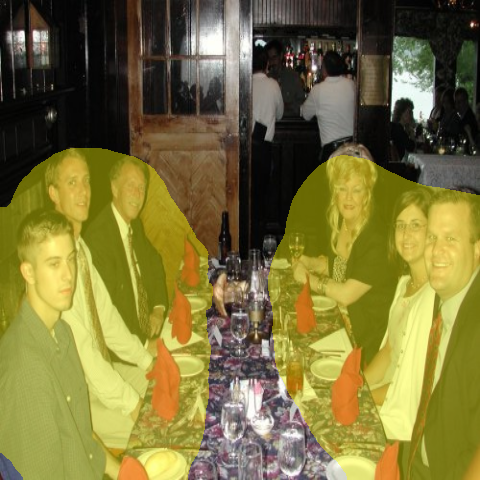}       &
        \includegraphics[width=.24\linewidth]{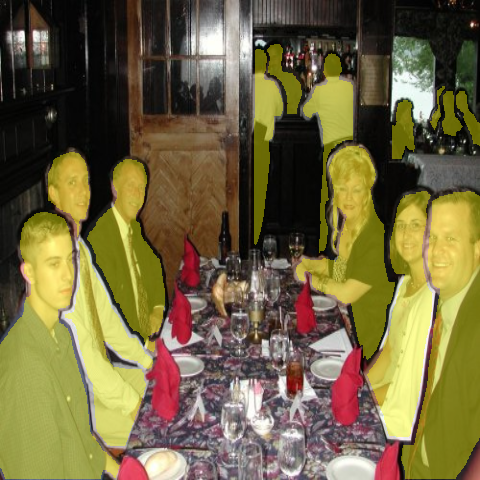}                                                                                                            \\
    \end{tabular}
    \caption{\gls{voc} qualitative examples for the \texttt{person} class. Colors in the multi-class output indicate mining order: warmer colors are earlier discoveries and cooler colors are later discoveries. In these examples, early discoveries tend to emphasize discriminative features such as the face and upper body, while later discoveries extend to other body regions. The binarized output is used for evaluation against the ground-truth mask.}
    \label{fig:voc_qual}
\end{figure}

\paragraph{\gls{bus} \& \gls{voc}}

Following the qualitative cyclone analysis, we now turn to the \gls{bus} and \gls{voc} datasets, whose first role is to make the mining order legible.
The visual comparisons in Figs.~\ref{fig:bus_qual} and~\ref{fig:voc_qual} illustrate how the color-coded mining tiers relate to different parts of the target objects in selected examples.
In all three datasets, the first tier lands on the most diagnostic part of the target, and later tiers spread outward from it.
On \gls{voc}, the face and upper body come first, then the rest of the person; on \gls{bus}, the lesion core, then its margin; on \gls{sar}, the cyclone eye, then the surrounding bands (\Cref{fig:pl_comparison,fig:slp}).
\gls{voc} shows this most clearly, since the ordering can be judged without domain expertise, whereas on \gls{sar} imagery the same judgment requires a meteorologist.
That the same core-to-periphery progression appears on all three targets indicates that the mining order follows how strongly a region signals the target, rather than anything specific to cyclones.
The ablations below supply a quantitative counterpart: weighting supervision by that order improves the resulting masks, which would not happen if the order carried no information about reliability.
These qualitative examples are shown at $t=4$ on \gls{bus} and $t=5$ on \gls{voc}, independently of the operating points used for the quantitative results below.

Since these datasets provide ground-truth masks, they permit a quantitative assessment under the protocols described in Appendix~\ref{app:dataset}.
At inference, the predicted tier is $\hat{s}_u=\arg\max_k p_{u,k}$ at each pixel $u$. 
For evaluation, pixels with $\hat{s}_u>0$ are merged into a single foreground class.
For the iterative methods, we vary how many mined tiers are retained in the final mask, and Tables~\ref{tab:test_results} and~\ref{tab:ablation_components} report, for each method, its best test result over the evaluated iterations.
For traceability, the retained iteration is the one maximizing Macro IoU over the reported sweep, with ties broken by foreground IoU: on \gls{bus}, $t=2$ for standard \aer{}, $t=4$ for \aercore{}, and $t=3$ for \gls{crest}; on \gls{voc}, $t=4$ for standard \aer{} and \aercore{}, and $t=6$ for \gls{crest}, which ties $t=5$ on Macro IoU.
Tables~\ref{tab:bus_full_results} and~\ref{tab:voc_full_results} report the full per-iteration sweeps.
True positives, false positives, and false negatives are accumulated over all evaluated test pixels before each class score is computed.
Foreground IoU and Dice are the primary metrics; Macro IoU and Macro Dice are secondary summaries computed as the unweighted mean over background and foreground.
On \gls{bus}, \gls{crest} reaches a Macro IoU of $64.8\%$, compared with $60.2\%$ for standard \gls{aer} and $57.2\%$ for \gls{gradcam}.
For the foreground lesion class, \gls{crest} obtains an IoU of $32.7\%$ against $24.3\%$ for standard \gls{aer}, a gap of $8.4$ points.
On \gls{voc} the two methods are closer, with \gls{crest} ahead by $4.0$ points of foreground IoU on the \texttt{person} class.

Since these two peaks occur at different iterations, we also compare both methods at the iteration where standard \gls{aer} peaks ($t=2$ on \gls{bus}, $t=4$ on \gls{voc}).
Even at the baseline's own best iteration, \gls{crest} leads by $2.6$ and $1.7$ points of Macro IoU and by $4.4$ and $3.3$ points of foreground IoU (Tables~\ref{tab:bus_full_results} and~\ref{tab:voc_full_results}), margins that exceed the run-to-run spread measured below.
The gaps then widen over the later iterations, to $9.9$ and $13.9$ points on \gls{bus} at $t=4$ and to $3.8$ and $6.5$ points on \gls{voc} at $t=6$.
Only the earliest \gls{voc} iterations run the other way: at $t=1$ and $t=2$, where little beyond the most discriminative core has been mined, standard \gls{aer} is marginally ahead in foreground IoU, by $1.5$ and $0.2$ points.
This is consistent with the role of the \gls{db} loss, which pays off later, once noisier iterations begin to contribute supervision.

Taken together, these numbers show that the constrained, tier-aware pipeline improves on standard \gls{aer} under identical conditions, so the multi-class output and the region constraint do not compromise mask quality.

Refinement-based and promptable methods rely on strong intensity edges, which do not appear in the diffuse cyclone boundaries (\Cref{fig:sam}).
To our knowledge, standard \gls{aer} is therefore one of the most suitable approaches currently applicable to polar lows in \gls{sar} imagery.
\gls{bus} and \gls{voc} compare standard \gls{aer} with \gls{crest}, so they let us measure the improvement that we can only see qualitatively on the \gls{sar} data (\Cref{fig:pl_comparison}).

\subsection{Ablations \& Analysis}

\begin{table*}[!ht]
    \centering
    \caption{Progressive ablation results on the \gls{bus} and \gls{voc} test sets, where each row shows the effect of adding components to the baseline \aer{} pipeline. For each method, we report the mining iteration that achieved the best Macro IoU.}
    \label{tab:ablation_components}
    \begin{tabular*}{\textwidth}{@{\extracolsep{\fill}}lcccccccc@{}}
        \toprule
        Method &
        \multicolumn{4}{c}{\gls{bus}} &
        \multicolumn{4}{c}{\gls{voc}} \\
        \cmidrule(lr){2-5}\cmidrule(lr){6-9}
        & Macro IoU $\uparrow$ & Macro Dice $\uparrow$ & IoU$_{c{=}1}$ $\uparrow$ & Dice$_{c{=}1}$ $\uparrow$
        & Macro IoU $\uparrow$ & Macro Dice $\uparrow$ & IoU$_{c{=}1}$ $\uparrow$ & Dice$_{c{=}1}$ $\uparrow$ \\
        \midrule
        Standard \aer{} & 60.2 & 68.6 & 24.3 & 39.1 & 65.4 & 74.4 & 33.5 & 50.2 \\
        \aercore{} & 61.4 & 70.0 & 26.4 & 41.8 & 66.5 & 75.6 & 35.6 & 52.5 \\
        \textbf{\aercoredb{}} & \textbf{64.8} & \textbf{73.9} & \textbf{32.7} & \textbf{49.3} & \textbf{67.4} & \textbf{76.6} & \textbf{37.5} & \textbf{54.6} \\
        \bottomrule
    \end{tabular*}
\end{table*}

We perform a progressive ablation study to assess the cumulative contributions of the spatial constraint provided by \gls{core} and the reliability-aware supervision from \gls{db}.
In the tables and plots, we use the notation \aercoredb{} to denote the full \gls{crest} configuration.
To judge which differences are meaningful, we first quantify how much results vary between runs.
Before \gls{core} activates, the standard \aer{} and \aercore{} rows of Tables~\ref{tab:bus_full_results} and~\ref{tab:voc_full_results} run the same procedure under the same seed (Appendix~\ref{app:exp_settings}) and differ only through nondeterministic execution: Macro IoU by up to $1.2$ points and foreground IoU by up to $2.3$ points, which we take as the run-to-run spread.

\paragraph{Incremental module impact}

Table~\ref{tab:ablation_components} quantifies the gain from each module.
Starting from the standard \aer{} baseline, the addition of the \gls{core} module (\aercore{}) improves Macro IoU on both datasets ($+1.2$ points on \gls{bus} and $+1.1$ points on \gls{voc}), and foreground IoU by $+2.1$ points on both.
The contribution of \gls{core} is not a higher peak but the prevention of the mask drifting into the background at later iterations.
The largest gain comes from adding the \gls{db} loss on top of \gls{core}, which increases foreground IoU by $6.3$ points on \gls{bus} and $1.9$ points on \gls{voc}, with the \gls{voc} gain concentrated in the final iterations.

\paragraph{Iteration-wise stability}

\begin{figure}[!ht]
    \centering
    \includegraphics[width=\linewidth]{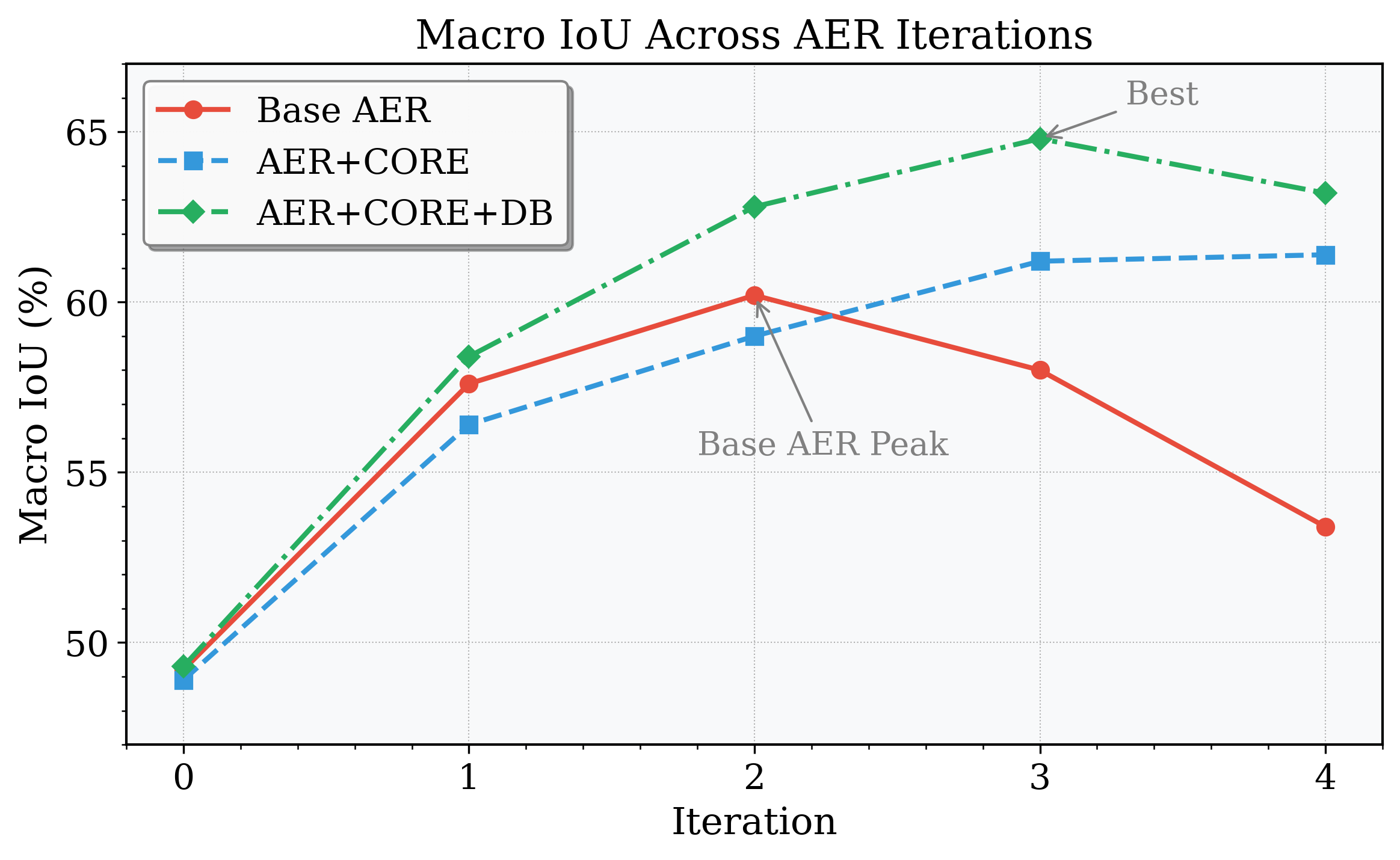}
    \caption{Macro IoU over retained adversarial-erasing iterations on the \gls{bus} test set. 
    Standard \aer{} degrades after iteration 2, whereas the \aercore{} and full \gls{crest} (\aercoredb{}) trajectories vary less over the later iterations.}
    \label{fig:iteration_stability}
\end{figure}

One weakness of standard \gls{aer} is its sensitivity to the number of retained mining iterations. 
With few iterations, the pseudo-label may contain only the most discriminative core, leading to under-segmentation.
With many iterations, the classifier may increasingly rely on background textures or acquisition artifacts after informative regions have been erased; recall can increase while precision decreases.
This behavior is clearly visible in the detailed \gls{bus} results in Table~\ref{tab:bus_full_results}: standard \gls{aer} moves from $70.6\%$ precision / $2.0\%$ recall at iteration 0 to $19.4\%$ precision / $53.3\%$ recall at iteration 4, illustrating the shift from severe under-segmentation to over-expansion.
To illustrate this sensitivity, \Cref{fig:iteration_stability} plots final segmentation performance on \gls{bus}, indexed by the last retained mining iteration.
The curve for vanilla \gls{aer} (orange) reveals a sharp performance degradation after iteration 2, a pattern consistent with progressively noisier pseudo-label expansion.
The \gls{core} trajectory (blue) is flatter over the evaluated later iterations.
At the last evaluated iteration on \gls{bus} ($t=4$, by which point \gls{core} has been active for three iterations), \aercore{} reaches $61.4\%$ Macro IoU and $26.4\%$ foreground IoU against $53.4\%$ and $16.6\%$ for standard \aer{} (Table~\ref{tab:bus_full_results}), gaps of $8.0$ and $9.8$ points that lie far outside the spread estimated at the start of this subsection.
The corresponding gaps on \gls{voc} at $t=6$ are smaller, $1.9$ and $3.2$ points, so the effect is most pronounced on the dataset whose baseline degrades most sharply.
The full \gls{crest} trajectory (green; \aercoredb{}) also varies less than standard \gls{aer} over the later iterations.
This behavior is consistent with the lower weights assigned to late-mined pixels and indicates reduced, rather than eliminated, sensitivity to the retained-iteration endpoint.

\paragraph{Precision--recall trade-off}

\begin{figure}[!ht]
    \centering
    \includegraphics[width=\linewidth]{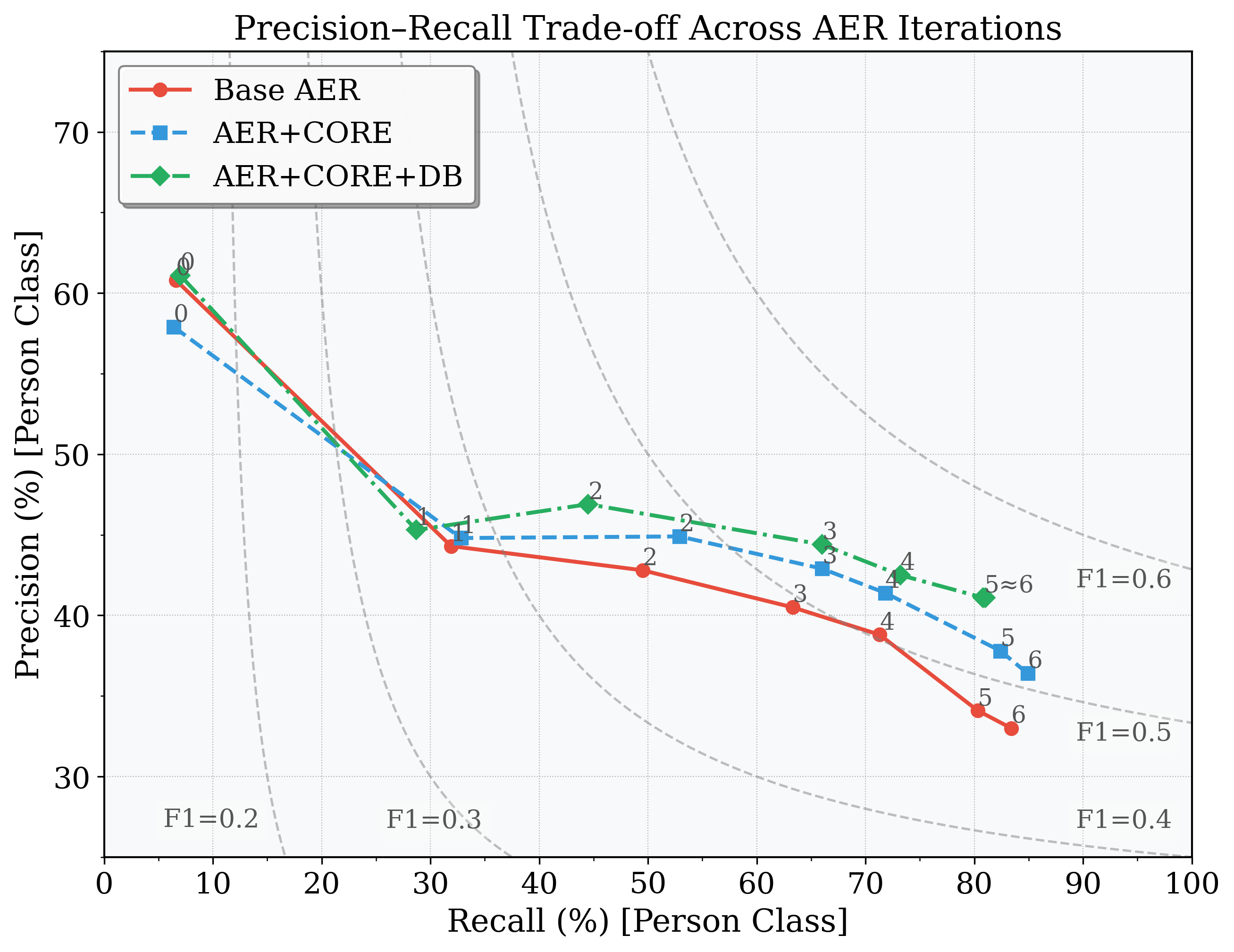}
    \caption{Discrete precision--recall trajectories on the \gls{voc} test set as additional mining iterations are retained.
    Numbers indicate the iteration index.
    In the legend, \aercoredb{} is the full \gls{crest} configuration.}
    \label{fig:pr_curve}
\end{figure}

\Cref{fig:pr_curve} visualizes the discrete trade-off between expanding coverage and retaining precision across iterations on \gls{voc}.
For standard \aer{} (orange), recall increases from $6.6\%$ to $83.4\%$, while precision decreases from $60.8\%$ to $33.0\%$, which is consistent with progressively broader foreground predictions.
The \gls{core} and full \gls{crest} trajectories retain more precision over the later iterations.
At the final iteration, \gls{crest} achieves $41.1\%$ precision at $81.0\%$ recall, compared with $33.0\%$ precision at $83.4\%$ recall for standard \aer{}.
The advantage is not confined to the endpoint: at $t=4$, \gls{crest} is ahead on both axes, with $42.5\%$ precision at $73.2\%$ recall against $38.8\%$ at $71.3\%$ for standard \aer{}.

\begin{table*}[!ht]
    \centering
    \caption{Ablation results using Grad-CAM++ attribution on the \gls{bus} and \gls{voc} test sets. For each method, we report the mining iteration that achieved the best Macro IoU (standard \aer{} at $t=2$ and \gls{crest} at $t=3$ on \gls{bus}; both at $t=6$ on \gls{voc}).}
    \label{tab:ablation_attribution}
    \begin{tabular}{lcccc}
        \toprule
        Method                        &
        \multicolumn{2}{c}{\gls{bus}} &
        \multicolumn{2}{c}{\gls{voc}}                                                                                   \\
        \cmidrule(lr){2-3}\cmidrule(lr){4-5}
                                      & Macro IoU $\uparrow$ & IoU$_{c{=}1}$ $\uparrow$
                                      & Macro IoU $\uparrow$ & IoU$_{c{=}1}$ $\uparrow$                                 \\
        \midrule
        Standard \aer{}               & 57.5                 & 18.6                     & 65.7          & 34.4          \\
        \textbf{\gls{crest}}         & \textbf{59.6}        & \textbf{22.7}            & \textbf{68.4} & \textbf{38.8} \\
        \bottomrule
    \end{tabular}
\end{table*}

\paragraph{Sensitivity to the attribution method}

We conclude with a sensitivity check in which the attribution mechanism is changed from \gls{gradcam} to Grad-CAM++~\cite{chattopadhay2018grad_cam_plus}, without changing the thresholds or $\kappa$. 
\Cref{tab:ablation_attribution} reports higher values for \gls{crest} than for standard \gls{aer} on both datasets.
On \gls{bus}, foreground IoU is lower for both methods than with \gls{gradcam} ($-5.7$ and $-10.0$ points), possibly because the fixed thresholds and $\kappa$ are less well suited to the activation profile of Grad-CAM++, whereas on \gls{voc} both methods improve slightly ($+0.9$ and $+1.3$ points).
Under this substitution, the gap in foreground IoU between \gls{crest} and standard \gls{aer} on \gls{bus} is $4.1$ points.
This suggests that the observed advantage over standard \gls{aer} is retained when the attribution method is changed.

\section{Conclusions}\label{sec:conclusions}

In this work, we presented \gls{crest}, a novel weakly supervised framework for training a neural network for image segmentation.
We designed the proposed method to bridge the gap between binary classification labels and pixel-level understanding of polar lows in \gls{sar} imagery.
While standard \gls{aer} techniques can iteratively discover object features, their mining process can drift into the background and blend irrelevant regions with meaningful target features.

We addressed this limitation with two main contributions. 
First, the \gls{core} module restricts new discoveries to a local envelope around the accumulated support. 
Second, the tier-aware \gls{db} loss uses the preserved mining order as a proxy for reliability and progressively attenuates cross-entropy supervision from later-mined regions.

In the qualitative examples, \gls{crest} produces visually coherent masks that follow the visible cyclone structure more closely than standard \gls{aer}, and whose tiers order the mask from the cyclone core outward, with no human delineation entering the pipeline at any stage.
On the \gls{bus} medical dataset and the \gls{voc} benchmark, which unlike the \gls{sar} data provide dense masks, the same core-to-periphery ordering appears on targets whose extent is easier to judge than a cyclone's.
At each method's best iteration, \gls{crest} improves on standard \gls{aer} by $8.4$ points of foreground IoU on \gls{bus} and $4.0$ points on \gls{voc}; compared at the iteration where the baseline itself peaks, the margins are $4.4$ and $3.3$ points, and they widen over the later iterations.
Since refinement-based methods do not apply to diffuse cyclone boundaries, standard \gls{aer} is, to our knowledge, the best approach currently applicable to polar lows, so these margins measure the improvement that the \gls{sar} data can only show qualitatively.
These experiments validate the mechanism under dense ground truth rather than compete on these benchmarks: no \gls{crf} or other post-processing is applied to any method, so the absolute values remain below what specialized state-of-the-art segmentation methods achieve on either dataset.

The current formulation of \gls{core} assumes locally coherent growth around the already discovered support.
Although this constraint could be relaxed in future work, the results show that \gls{crest} can extract spatial structure from image-level labels in different settings.
From a meteorological perspective, this is an important practical contribution because dense cyclone masks are scarce and difficult to standardize: their boundaries are diffuse, evolve rapidly, and often require expert interpretation. 
Our framework provides a scalable route for generating candidate spatial labels from existing detection archives.

The masks could support future studies of cyclone size, shape, compactness, asymmetry, and internal structure, as well as their relationship with surrounding environmental fields.
More broadly, the same weakly supervised paradigm is relevant to other meteorological events for which pixel-level masks are scarce and difficult to obtain because the phenomena have fuzzy boundaries and require expert interpretation. 
In this context, \gls{crest} can help generate spatial labels in scientific domains where they are currently missing.

\section{Acknowledgment}
This work is supported by the Research Council of Norway through \textit{RELAY:~Relational Deep Learning for Energy Analytics} (project no. 345017).
The authors wish to thank NVIDIA Corporation for donating the GPUs used in this project.

The auxiliary meteorological fields were downloaded from the Copernicus Climate Change Service (2023) Climate Data Store.
The results contain modified Copernicus Climate Change Service information 2019. Neither the European Commission nor ECMWF is responsible for any use that may be made of the Copernicus information or data it contains.
\Cref{fig:slp} contains modified Copernicus Sentinel data 2019 processed in Copernicus Browser.
\bibliographystyle{IEEEtran}
\bibliography{references.bib}
\clearpage
\appendices

\section{Additional Qualitative Results on Polar Lows}
\label{app:pl_results}

\begin{figure}[H]
  \centering
  \setlength{\tabcolsep}{2pt} 
  \begin{tabular}{@{}c c @{\hspace{15pt}} c c@{}}
    \textbf{Input}                                                                                                        & \textbf{Output} & \textbf{Input} & \textbf{Output} \\
    \includegraphics[width=.22\linewidth]{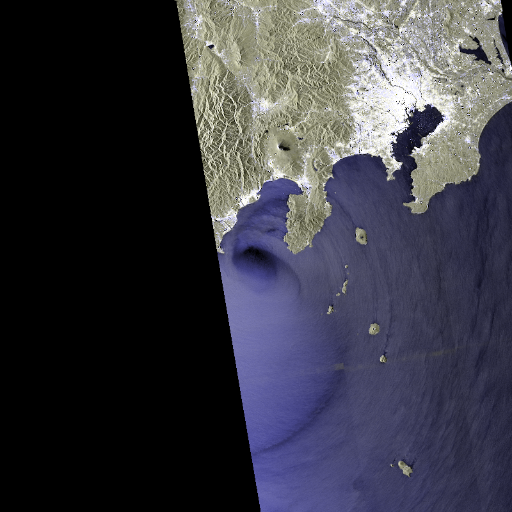}   &
    \includegraphics[width=.22\linewidth]{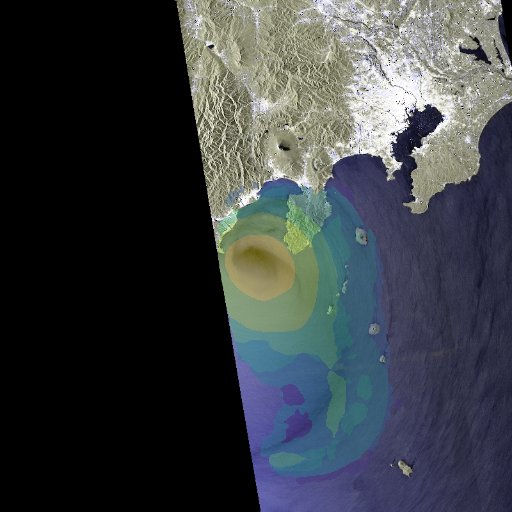} &
    \includegraphics[width=.22\linewidth]{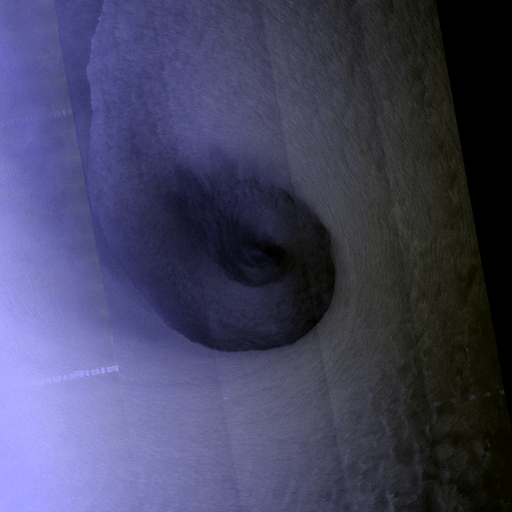}   &
    \includegraphics[width=.22\linewidth]{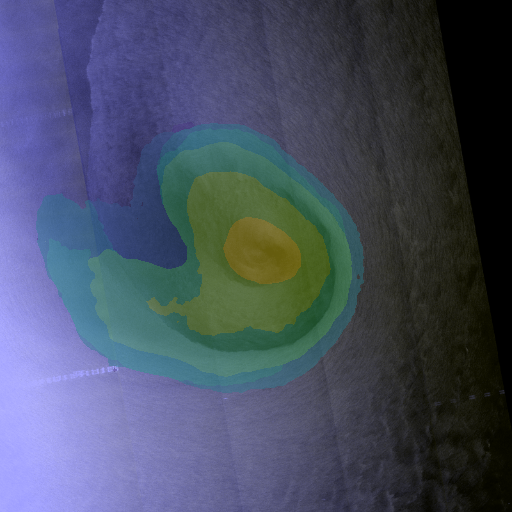}                                                      \\
    \includegraphics[width=.22\linewidth]{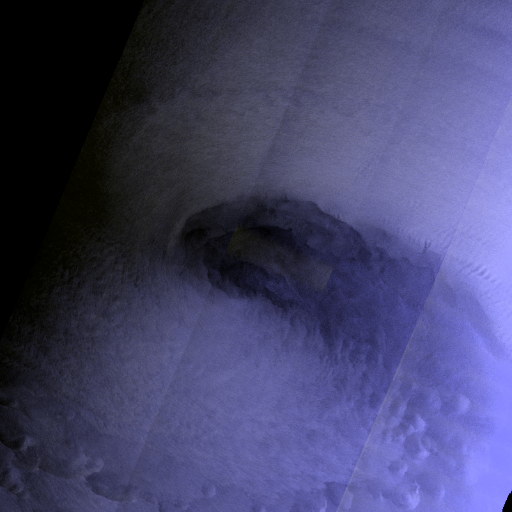}   &
    \includegraphics[width=.22\linewidth]{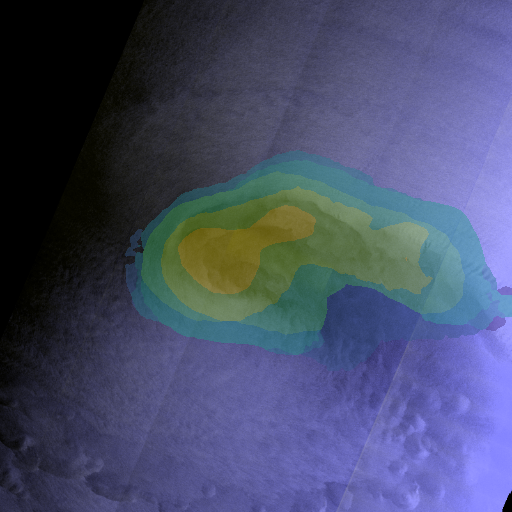} &
    \includegraphics[width=.22\linewidth]{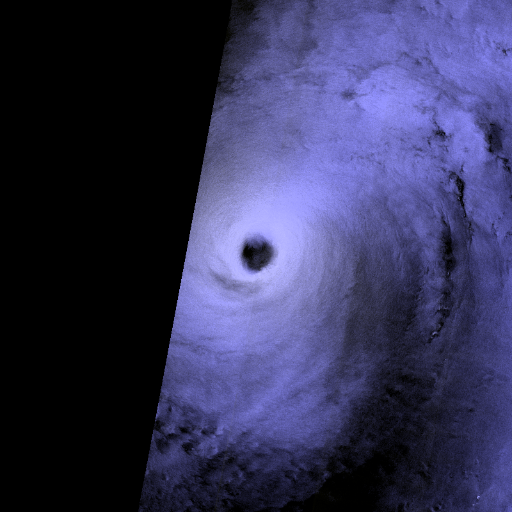}   &
    \includegraphics[width=.22\linewidth]{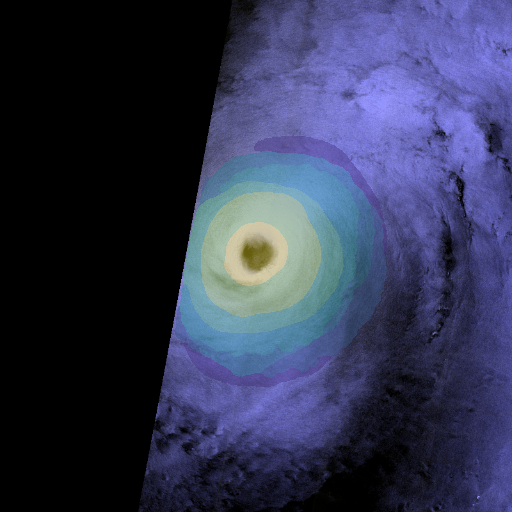}                                                      \\
    \includegraphics[width=.22\linewidth]{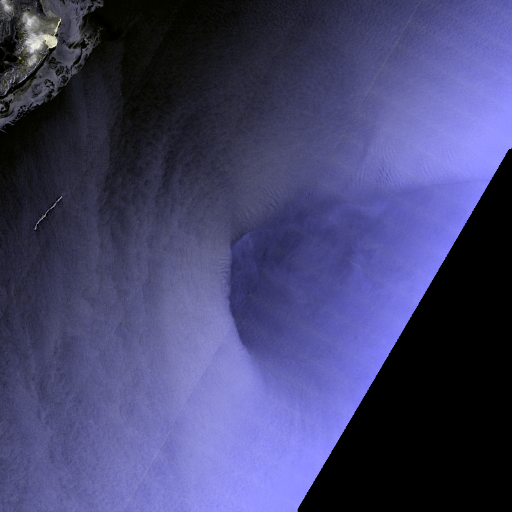}   &
    \includegraphics[width=.22\linewidth]{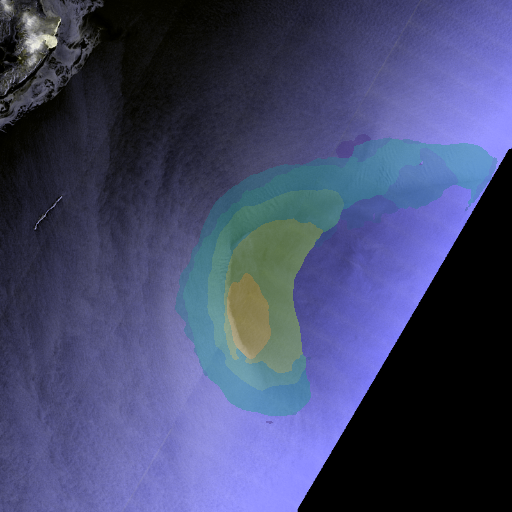} &
    \includegraphics[width=.22\linewidth]{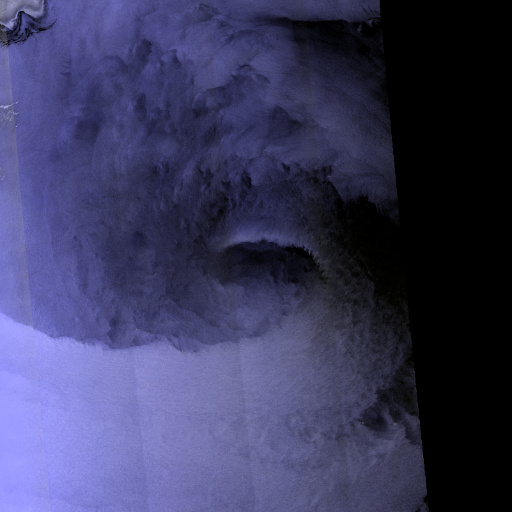}   &
    \includegraphics[width=.22\linewidth]{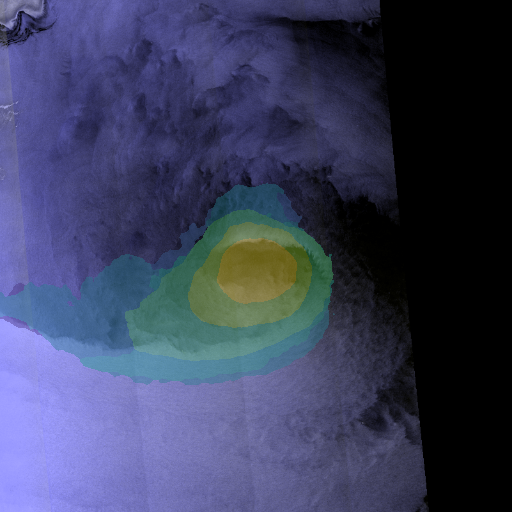}                                                      \\
    \includegraphics[width=.22\linewidth]{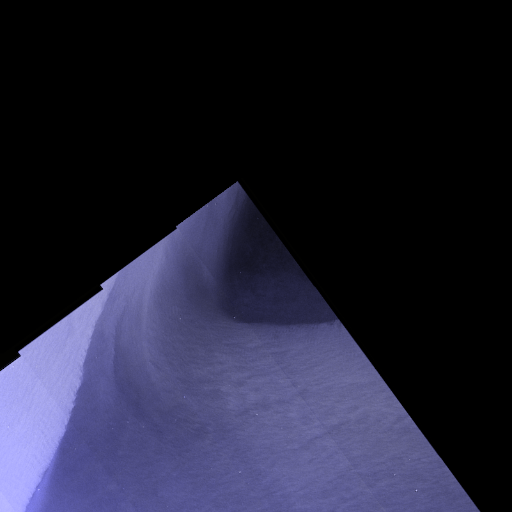}   &
    \includegraphics[width=.22\linewidth]{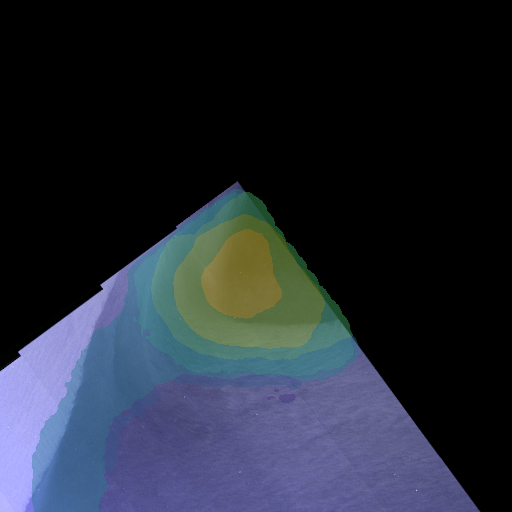} &
    \includegraphics[width=.22\linewidth]{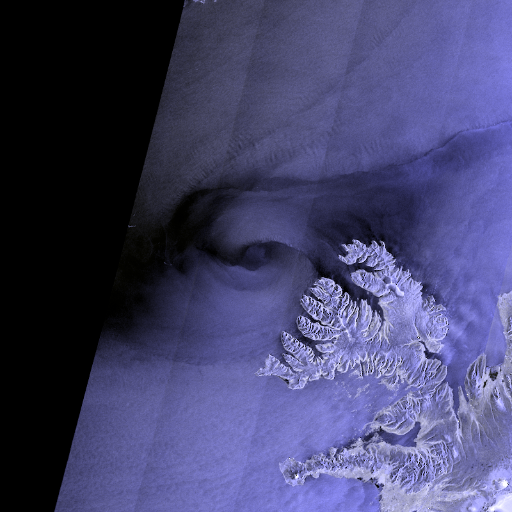}   &
    \includegraphics[width=.22\linewidth]{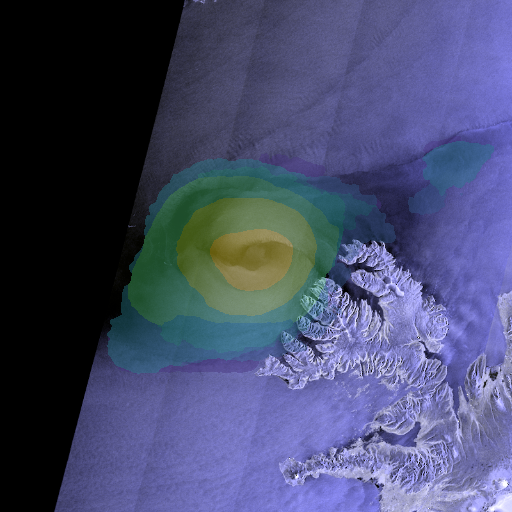}                                                      \\
    \includegraphics[width=.22\linewidth]{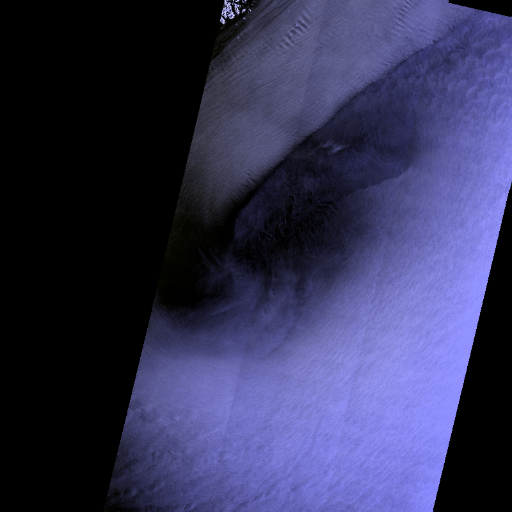}   &
    \includegraphics[width=.22\linewidth]{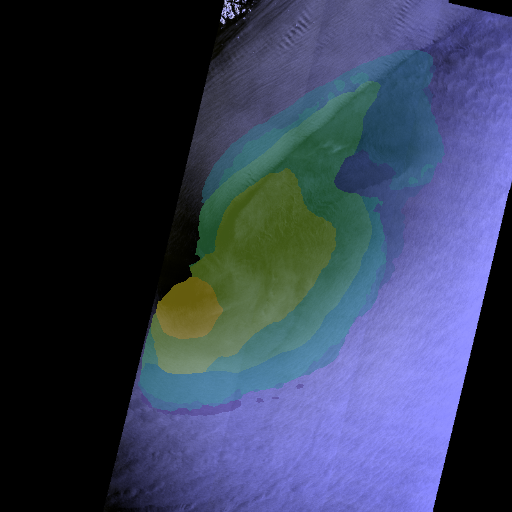} &
    \includegraphics[width=.22\linewidth]{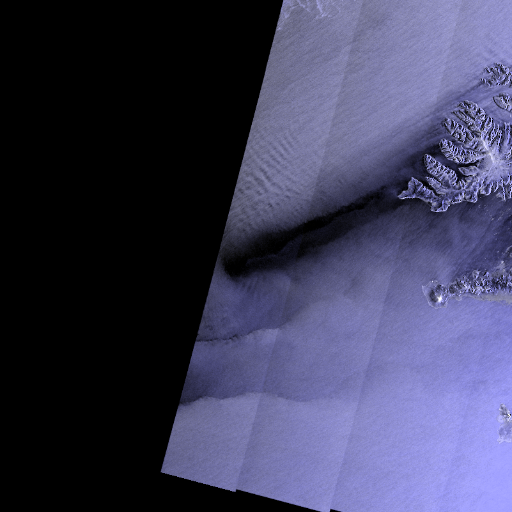}   &
    \includegraphics[width=.22\linewidth]{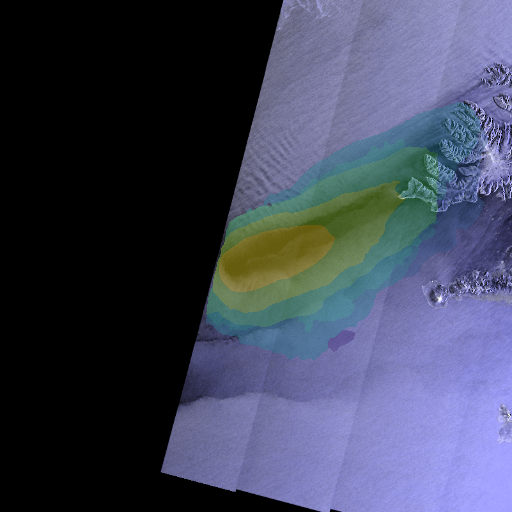}                                                      \\
  \end{tabular}
  \caption{Mosaic of inputs and predicted masks for test images. Warmer colors (red/yellow) denote earlier-mined regions; cooler colors (blue/cyan) denote later-mined regions.}
  \label{fig:pl_mosaic}
\end{figure}

To provide a broader qualitative view, we present a gallery of segmentation results in \Cref{fig:pl_mosaic}. These examples are drawn from the test set and cover varied cyclone morphologies and sea states.
In these examples, \gls{crest} identifies both compact regions (warm colors) and more diffuse structures (cool colors). 
The colors encode mining order and should not be interpreted as calibrated probabilities.

\section{Qualitative Consistency under Geometric Transformations}
\label{app:robustness}

\begin{figure}[h]
  \centering
  \begin{subfigure}[t]{0.98\linewidth}
    \centering
    \begin{minipage}[c]{0.36\linewidth}
      \centering
      \includegraphics[width=\linewidth]{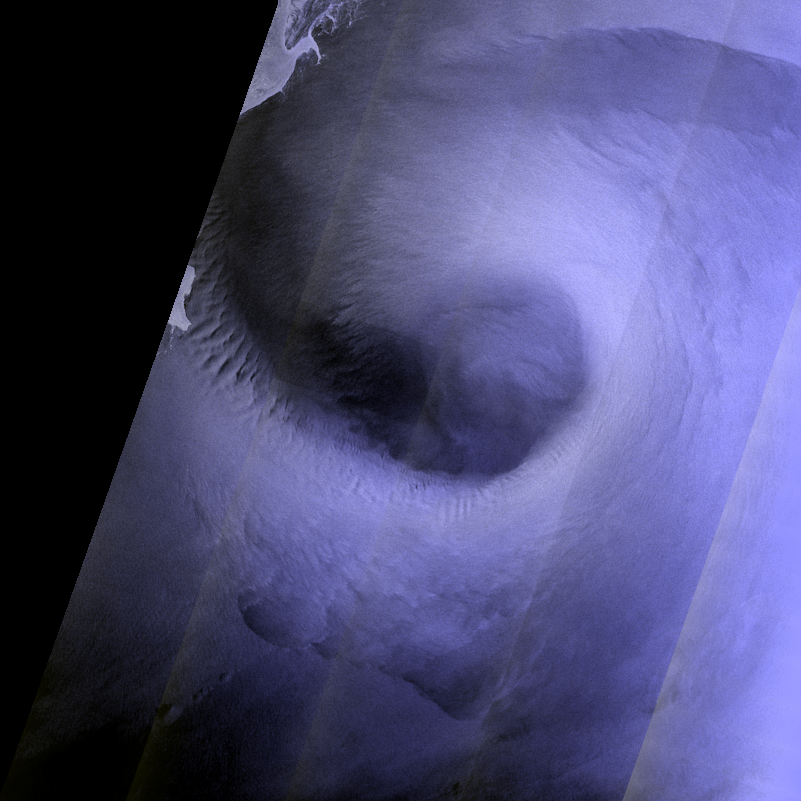}\\[-2pt]
      \textbf{Original image 1}
    \end{minipage}\hfill
    \begin{minipage}[c]{0.62\linewidth}
      \centering
      \includegraphics[width=0.495\linewidth]{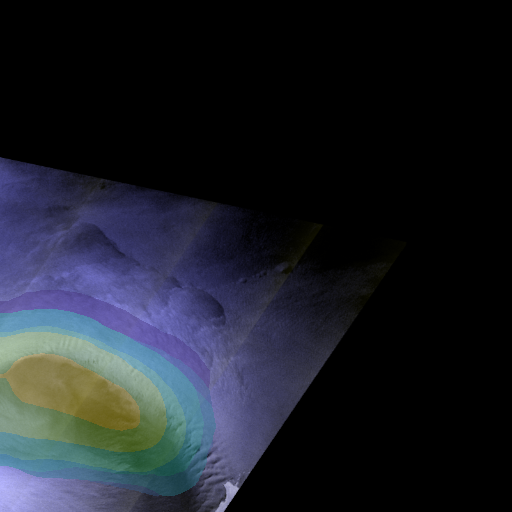}\hfill
      \includegraphics[width=0.495\linewidth]{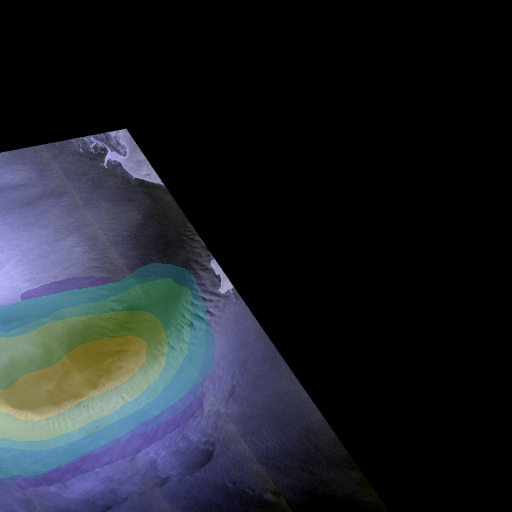}\\[4pt]
      \includegraphics[width=0.495\linewidth]{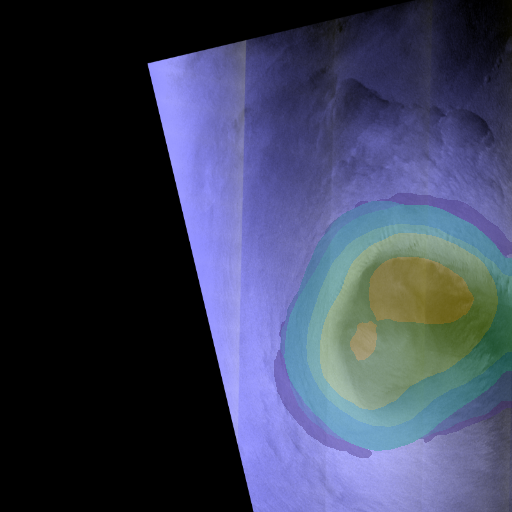}\hfill
      \includegraphics[width=0.495\linewidth]{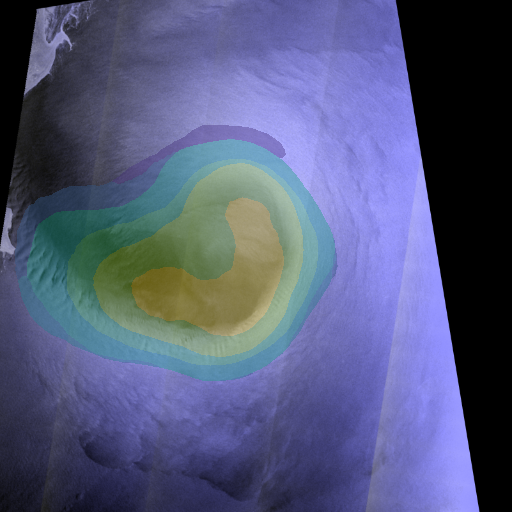}
    \end{minipage}
    \subcaption{Example 1: The predicted mask follows the displayed rotation.}
    \label{fig:augmentation_consistency_ex1}
  \end{subfigure}\\[10pt]
  \begin{subfigure}[t]{0.98\linewidth}
    \centering
    \begin{minipage}[c]{0.36\linewidth}
      \centering
      \includegraphics[width=\linewidth]{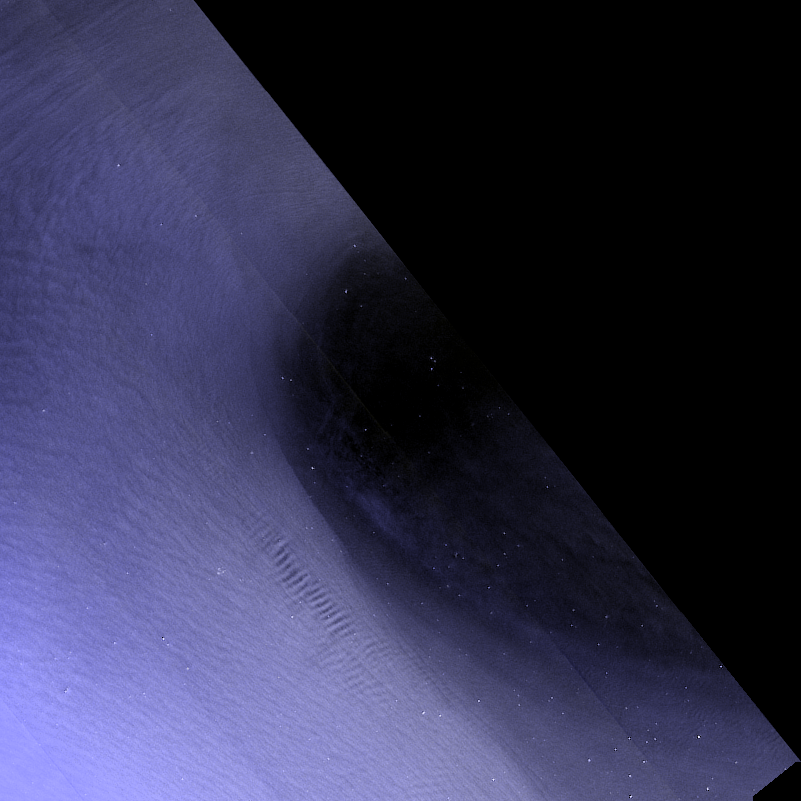}\\[-2pt]
      \textbf{Original image 2}
    \end{minipage}\hfill
    \begin{minipage}[c]{0.62\linewidth}
      \centering
      \includegraphics[width=0.495\linewidth]{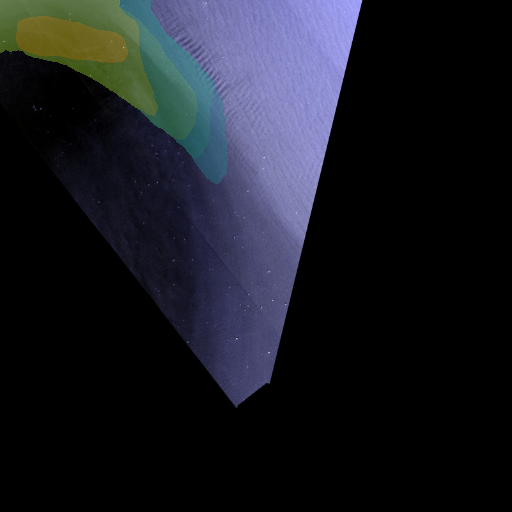}\hfill
      \includegraphics[width=0.495\linewidth]{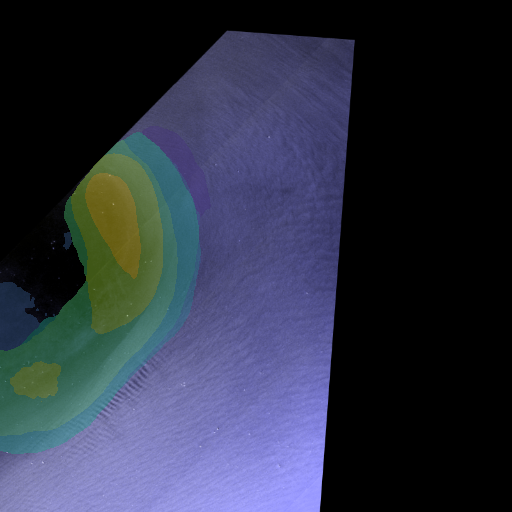}\\[4pt]
      \includegraphics[width=0.495\linewidth]{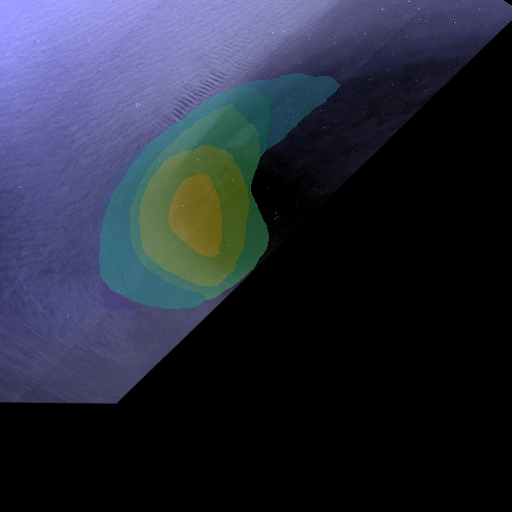}\hfill
      \includegraphics[width=0.495\linewidth]{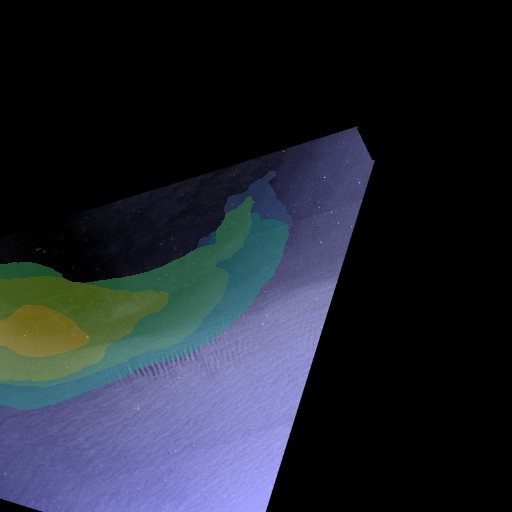}
    \end{minipage}
    \subcaption{Example 2: Similar structure is retained despite visible border effects.}
    \label{fig:augmentation_consistency_ex2}
  \end{subfigure}

  \caption{Qualitative consistency under geometric transformations. Left: original image. Right: predictions for the original and three transformed versions.}
  \label{fig:augmentation_consistency}
\end{figure}

An important property of a segmentation model is equivariance to geometric transformations. In \Cref{fig:augmentation_consistency}, we qualitatively inspect \gls{crest} predictions under rotations and translations.
Each row displays an original \gls{sar} image followed by its segmentation and the segmentations of three augmented versions. Ideally, the predicted mask should rotate and translate in lockstep with the input image, preserving the same shape and internal structure.
The examples show that the predicted structures rotate and translate consistently with the displayed inputs.

\section{Detailed Per-Iteration Results}
\label{app:detailed_results}

\begin{table*}[!ht]
  \centering
  \caption{Per-iteration results on the \gls{bus} test set. For the iterative methods, bold marks the best value per column across mining iterations.}
  \label{tab:bus_full_results}
  \small
  \begin{tabular}{l|cc|cccc}
    \toprule
    \textbf{Method}   & \textbf{Macro IoU $\uparrow$} & \textbf{Macro Dice $\uparrow$} & \textbf{IoU$_{c{=}1}$ $\uparrow$} & \textbf{Dice$_{c{=}1}$ $\uparrow$} & \textbf{Precision$_{c{=}1}$ $\uparrow$} & \textbf{Recall$_{c{=}1}$ $\uparrow$} \\
    \midrule
    \textit{Grad-CAM} & 57.2                          & 64.2                           & 17.7                              & 30.1                               & 62.4                                    & 19.8                                 \\
    \midrule
    \multicolumn{7}{l}{\textit{Standard \aer{}}}                                                                                                                                                                                                 \\
    Iteration 0       & 49.2                          & 51.0                           & 2.0                               & 3.9                                & \textbf{70.6}                           & 2.0                                  \\
    Iteration 1       & 57.6                          & 64.7                           & 18.3                              & 31.0                               & 67.0                                    & 20.1                                 \\
    Iteration 2       & \textbf{60.2}                 & \textbf{68.6}                  & \textbf{24.3}                     & \textbf{39.1}                      & 45.0                                    & 34.6                                 \\
    Iteration 3       & 58.0                          & 66.5                           & 22.0                              & 36.0                               & 29.6                                    & 46.1                                 \\
    Iteration 4       & 53.4                          & 61.7                           & 16.6                              & 28.5                               & 19.4                                    & \textbf{53.3}                        \\
    \midrule
    \multicolumn{7}{l}{\textit{\aercore{}}}                                                                                                                                                                                                      \\
    Iteration 0       & 48.9                          & 50.5                           & 1.4                               & 2.7                                & \textbf{74.8}                           & 1.4                                  \\
    Iteration 1       & 56.4                          & 63.0                           & 16.0                              & 27.6                               & 74.2                                    & 16.9                                 \\
    Iteration 2       & 59.0                          & 66.7                           & 21.2                              & 34.9                               & 70.0                                    & 23.3                                 \\
    Iteration 3       & 61.2                          & 69.5                           & 25.5                              & 40.7                               & 64.2                                    & 29.8                                 \\
    Iteration 4       & \textbf{61.4}                 & \textbf{70.0}                  & \textbf{26.4}                     & \textbf{41.8}                      & 49.9                                    & \textbf{35.9}                        \\
    \midrule
    \multicolumn{7}{l}{\textit{\textbf{\gls{crest} (\aercoredb{})}}}                                                                                                                                                                            \\
    Iteration 0       & 49.3                          & 51.2                           & 2.2                               & 4.2                                & \textbf{83.6}                           & 2.2                                  \\
    Iteration 1       & 58.4                          & 65.9                           & 20.1                              & 33.4                               & 64.5                                    & 22.6                                 \\
    Iteration 2       & 62.8                          & 71.5                           & 28.7                              & 44.5                               & 62.0                                    & 34.8                                 \\
    Iteration 3       & \textbf{64.8}                 & \textbf{73.9}                  & \textbf{32.7}                     & \textbf{49.3}                      & 58.7                                    & 42.5                                 \\
    Iteration 4       & 63.3                          & 72.4                           & 30.5                              & 46.8                               & 46.0                                    & \textbf{47.6}                        \\
    \bottomrule
  \end{tabular}
\end{table*}

\begin{table*}[!ht]
  \centering
  \caption{Per-iteration results on the \gls{voc} test set (person class). For the iterative methods, bold marks the best value per column across mining iterations.}
  \label{tab:voc_full_results}
  \small
  \begin{tabular}{l|cc|cccc}
    \toprule
    \textbf{Method}   & \textbf{Macro IoU $\uparrow$} & \textbf{Macro Dice $\uparrow$} & \textbf{IoU$_{c{=}1}$ $\uparrow$} & \textbf{Dice$_{c{=}1}$ $\uparrow$} & \textbf{Precision$_{c{=}1}$ $\uparrow$} & \textbf{Recall$_{c{=}1}$ $\uparrow$} \\
    \midrule
    \textit{Grad-CAM} & 51.8                          & 54.8                           & 5.7                               & 10.7                               & 39.8                                    & 6.2                                  \\
    \midrule
    \multicolumn{7}{l}{\textit{Standard \aer{}}}                                                                                                                                                                                                 \\
    Iteration 0       & 52.2                          & 55.5                           & 6.3                               & 11.9                               & \textbf{60.8}                           & 6.6                                  \\
    Iteration 1       & 60.3                          & 68.0                           & 22.8                              & 37.1                               & 44.3                                    & 31.9                                 \\
    Iteration 2       & 63.7                          & 72.4                           & 29.8                              & 45.9                               & 42.8                                    & 49.5                                 \\
    Iteration 3       & 65.1                          & 74.0                           & 32.8                              & 49.4                               & 40.5                                    & 63.3                                 \\
    Iteration 4       & \textbf{65.4}                 & \textbf{74.4}                  & \textbf{33.5}                     & \textbf{50.2}                      & 38.8                                    & 71.3                                 \\
    Iteration 5       & 64.0                          & 73.1                           & 31.5                              & 47.9                               & 34.1                                    & 80.3                                 \\
    Iteration 6       & 63.6                          & 72.7                           & 31.0                              & 47.3                               & 33.0                                    & \textbf{83.4}                        \\
    \midrule
    \multicolumn{7}{l}{\textit{\aercore{}}}                                                                                                                                                                                                      \\
    Iteration 0       & 52.1                          & 55.3                           & 6.1                               & 11.5                               & \textbf{57.9}                           & 6.4                                  \\
    Iteration 1       & 60.6                          & 68.4                           & 23.3                              & 37.8                               & 44.8                                    & 32.8                                 \\
    Iteration 2       & 64.9                          & 73.7                           & 32.1                              & 48.6                               & 44.9                                    & 52.9                                 \\
    Iteration 3       & 66.4                          & 75.4                           & 35.1                              & 52.0                               & 42.9                                    & 66.0                                 \\
    Iteration 4       & \textbf{66.5}                 & \textbf{75.6}                  & \textbf{35.6}                     & \textbf{52.5}                      & 41.4                                    & 71.8                                 \\
    Iteration 5       & 65.9                          & 75.1                           & 35.0                              & 51.8                               & 37.8                                    & 82.4                                 \\
    Iteration 6       & 65.5                          & 74.7                           & 34.2                              & 51.0                               & 36.4                                    & \textbf{84.9}                        \\
    \midrule
    \multicolumn{7}{l}{\textit{\textbf{\gls{crest} (\aercoredb{})}}}                                                                                                                                                                            \\
    Iteration 0       & 52.4                          & 55.8                           & 6.7                               & 12.5                               & \textbf{61.1}                           & 7.0                                  \\
    Iteration 1       & 59.6                          & 67.0                           & 21.3                              & 35.1                               & 45.3                                    & 28.7                                 \\
    Iteration 2       & 63.7                          & 72.3                           & 29.6                              & 45.6                               & 46.9                                    & 44.5                                 \\
    Iteration 3       & 66.9                          & 76.0                           & 36.2                              & 53.1                               & 44.4                                    & 66.0                                 \\
    Iteration 4       & 67.1                          & 76.2                           & 36.8                              & 53.8                               & 42.5                                    & 73.2                                 \\
    Iteration 5       & \textbf{67.4}                 & \textbf{76.6}                  & 37.4                              & 54.5                               & 41.1                                    & 80.8                                 \\
    Iteration 6       & \textbf{67.4}                 & \textbf{76.6}                  & \textbf{37.5}                     & \textbf{54.6}                      & 41.1                                    & \textbf{81.0}                        \\
    \bottomrule
  \end{tabular}
\end{table*}

Tables \ref{tab:bus_full_results} and \ref{tab:voc_full_results} provide a granular breakdown of segmentation performance at each mining iteration for the \gls{bus} and \gls{voc} datasets, respectively.
These tables detail the evolution of Macro IoU/Dice alongside class-specific metrics (mask IoU, precision, recall) for the positive class.
These results show how each method changes as more mining iterations are retained, and they document the sweep from which the values in the main tables are drawn.
Key observations from these tables include:
\begin{itemize}
  \item \textbf{Standard AER:} In both datasets, precision decreases as more iterations are retained (e.g., from 70.6\% at iteration 0 to 19.4\% at iteration 4 on \gls{bus}), a pattern consistent with progressively broader and noisier foreground predictions.
  \item \textbf{\gls{crest}:} The later-iteration trajectory retains more precision than standard \gls{aer}, reflecting the reduced influence of late-mined pixels.
\end{itemize}

\section{Dataset Details}
\label{app:dataset}

In the following, we describe the primary polar low dataset and the two benchmark datasets for quantitative evaluation.

\subsection{Sentinel-1 Dataset}

We use the public polar low dataset introduced by Grahn and Bianchi~\cite{Grahn_2022}\footnote{\url{https://doi.org/10.18710/FV5T9U}}.
The images are not native optical images, but standardized three-channel composites derived from Sentinel-1 \gls{sar} acquisitions. 
Each sample consists of a standardized \gls{sar}-derived RGB composite paired with an image-level binary label, indicating the presence or absence of a cyclone pattern.
The dataset provides image-level labels but no pixel-level masks, motivating the weakly supervised setting.

The dataset contains 1,982 geocoded samples, of which 318 belong to the positive class and 1,664 to the negative class. 
The images cover approximately $400 \times 400$ km at 500 m spacing, corresponding to roughly $800 \times 800$ pixels. 
The negative samples are chosen to remain visually challenging, so that background structures and acquisition conditions are comparable to those of the positive cases. 
Full details on dataset construction, \gls{sar} preprocessing, geocoding, and RGB-composite generation are deferred to the original dataset paper~\cite{Grahn_2022}.
Pixels outside the valid \gls{sar} swath are encoded as zero in the supplied composites and are treated as background during pseudo-label generation and segmentation training.

The auxiliary \gls{slp} and 10\,m wind fields used for the illustrative overlay in \Cref{fig:slp} are taken from \gls{era5}~\cite{hersbach2020era5}, retrieved as hourly data on single levels from the Copernicus Climate Data Store~\cite{C3S_CDS_ERA5_SingleLevels_2023, Hersbach_ERA5_SingleLevels_2023}.
The Sentinel-1 scene displayed in that figure is the same acquisition as the corresponding dataset sample, retrieved through the Copernicus Browser~\cite{Copernicus_Data_Space_Browser_2025} for visualization only.
These fields are used solely for post-hoc interpretation and are never provided to the model as inputs or supervision.


The data are split into 1,547 images used for training---254 positive ($c=1$) and 1,293 negative ($c=0$)---and 435 images for testing---64 positive ($c=1$) and 371 negative ($c=0$).
Since the dataset lacks an official validation split, we use a stratified 80/20 split of the training set for model selection.

\subsection{\gls{bus}} We use a dataset of breast ultrasound images containing pixel-level masks for three classes: benign tumors, malignant tumors, and normal tissue (no lesion).
We merge benign and malignant tumors into the positive class ($c=1$) and use normal tissue as the negative class ($c=0$). Totals: 419 negative, 264 positive.
Some images contain clinician overlays (bounding boxes, text) that could act as shortcuts for the classifier. We identify these artifacts and replace them with interpolated values to ensure a fair learning process.
We form the training, validation, and test subsets using stratified group splitting by patient, reserving approximately 20\% of the data for testing. All images from a patient remain in the same subset, so the splits are patient-disjoint.

\subsection{\gls{voc} (person-only, binarized)}
We use a custom subset derived from PASCAL VOC 2012, treating all \texttt{person} instances as foreground ($c=1$) and other labeled classes as background ($c=0$). The recorded split contains 1,595 positive and 1,595 negative development images, and 399 positive and 399 negative held-out images. A validation subset is obtained through a stratified 80/20 split of the development set.

\section{Experimental Settings}
\label{app:exp_settings}

In the following, we describe the experimental setting used for the two stages. 
All experiments are performed on an NVIDIA RTX A6000 (48\,GB).
All quantitative results are reported from a single run initialized with seed \(s=42\).

\paragraph{Stage 1---Pseudo-label generation} The classifier uses an Xception~\cite{xception2017} backbone pretrained on ImageNet-1K~\cite{deng2009imagenet, russakovsky2015imagenet}, followed by global average pooling and a multi-layer perceptron configured with $2048 \to 32 \to 2$ neurons. 
The classifier is optimized using Adam~\cite{kingma2017adammethodstochasticoptimization} (\(\mathrm{lr}=5\times10^{-4}\)) and \gls{ce}. The binarization thresholds for the Grad-CAM heatmaps are 0.7 (cyclone), 0.85 (\glsdisp{voc}{VOC}), and 0.9 (\glsdisp{bus}{BUS}).
In the zero-based indexing of Algorithm~\ref{alg:core_pipeline}, \core{} uses $t_{\mathrm{core}}=3$ and $\kappa=0.3$ for cyclone/\glsdisp{voc}{VOC}, and $t_{\mathrm{core}}=2$ and $\kappa=0.15$ for \glsdisp{bus}{BUS}. 
The number of iterations $N$ in the \gls{aer}+CORE loop is 7 for cyclone and \gls{voc}, and 5 for \glsdisp{bus}{BUS}.
We use early stopping and the \texttt{ReduceLROnPlateau} callback (factor 0.5, patience 7), monitoring the validation loss. Augmentations include translation/rotation/scale, flips, and random erasing~\cite{zhong2020random_erasing}. Color jitter is added for \glsdisp{bus}{BUS}. Inputs are resized to \(512\times512\) for cyclone/\glsdisp{bus}{BUS} and to \(480\times480\) for \glsdisp{voc}{VOC}.

\paragraph{Stage 2---Segmentation training} We initialize SegFormer~\cite{xie2021segformer} from a MiT-B3 checkpoint pretrained on ImageNet-1K~\cite{deng2009imagenet, russakovsky2015imagenet} and then densely supervised on ADE20K~\cite{zhou2017ade20k}.
We train SegFormer on pseudo-labels using AdamW~\cite{loshchilov2019adamw} (\(\mathrm{lr}=5\times10^{-4}\), weight decay 0.02), batch size 16. We use early stopping and \texttt{ReduceLROnPlateau} (factor 0.5, patience 7), monitoring validation F1 against the Stage 1 pseudo-labels, so that no dense annotation enters Stage~2 training on any dataset. Dense masks exist only for \gls{bus} and \gls{voc}, where they are used after training to compute the reported metrics.
As the loss, we use Dice+\gls{ce} with \gls{db} (equal \gls{ce}/Dice weighting, \(\alpha=0.5\); tier reliabilities $\rho_s$: [0.8, 0.2, 0.35, 0.5, 0.65, 0.75, 0.9, 1.0] for cyclone/\gls{voc}, [0.8, 0.2, 0.5, 0.7, 0.9, 1.0] for \glsdisp{bus}{BUS}). Augmentations and input resizing are the same as in Stage 1.
One full run of a given configuration therefore comprises the $N$ Stage~1 classifier cycles described above and a single Stage~2 training run.
No \gls{crf} or other post-processing is applied at either stage, for any method, on any of the three datasets.

\end{document}